%% file: main.tex
\documentclass[11pt]{article}

\usepackage[preprint]{acl}

\usepackage{times}
\usepackage{latexsym}
\usepackage[T1]{fontenc}
\usepackage[utf8]{inputenc}
\usepackage{microtype}
\usepackage{inconsolata}
\usepackage{graphicx}

\usepackage{subcaption}
\usepackage{booktabs}
\usepackage{amsmath}
\usepackage{amssymb}
\usepackage{mathtools}
\usepackage{amsthm}
\usepackage[capitalize,noabbrev]{cleveref}

\usepackage{algorithm}
\usepackage{algpseudocode}

\usepackage{xcolor}

\newcommand{\rev}[1]{#1}

\newcommand{\revcolor}{}   % used inside appendix tables, which \rev cannot reach
\newenvironment{revblock}{}{}

\theoremstyle{plain}

\theoremstyle{definition}

\theoremstyle{remark}

\title{Reasoning or Pattern Exploitation? \\
  Mechanistic Insights from RL-Trained Language Models}

\author{
  \textbf{Aishik Nagar}\textsuperscript{1} \quad
  \textbf{Cheston Tan}\textsuperscript{2,3} \\
  \textsuperscript{1}Carnegie Mellon University \\
  \textsuperscript{2}Institute of High Performance Computing (IHPC), Agency for Science, Technology and Research (A*STAR) \\
  \textsuperscript{3}Centre for Frontier AI Research (CFAR), Agency for Science, Technology and Research (A*STAR)
}

\begin{document}
\maketitle

\begin{abstract}
  Large language models (LLMs) are increasingly described as acquiring ``reasoning'' skills after reinforcement learning with verifiable rewards. 
  % Benchmark improvements are widely celebrated as progress toward higher-order reasoning.
% \cite{li2025system,liu2025logical,xu2025towards,sun2025game,topsakal2024evaluating,xie2026play}
However, whether these gains reflect genuine reasoning or more superficial adaptations remains underexplored.
% \cite{hua2024disentangling,xie2024memorization,wu2025knowledge}
In this work, we probe LLMs trained in a finite and exhaustively analyzable logical domain, namely \textbf{Tic-Tac-Toe}, and trace how internal representations evolve across reinforcement learning (RL) with \textbf{Group Relative Policy Optimization (GRPO}).
% \cite{tang2024generalized,guo2025deepseek}
Quantitatively, RL yields high accuracy and robustness across prompt variations, suggesting emergence of game reasoning. 
% \cite{dang2025reinforcement,srivastava2025towards}
Mechanistic interpretability, however, paints a different picture: features extracted with sparse autoencoders (SAEs) reveal that models primarily adapt to rely more heavily on information already explicit in the prompt, such as whose turn it is, game progression and board occupancy. 
% \cite{huben2024sparse,templeton2024scaling,marks2025sparse}
In comparison, high-level concepts like board symmetries and strategic forks remain weakly represented, raising concerns that reasoning benchmarks overestimate their abilities.
% \cite{stechly2025self,Hazra2025HaveLL,toh2025jumping,cosentino2024reasoning,MITNewsLLMReasoning2024}
This suggests that RL-driven ``reasoning'' may be conflating task-specific adaptations based on input information with reasoning ability.
We provide, for the first time:
\textbf{(i)} A systematic interpretability pipeline \textbf{tracing representation dynamics} across RL training in small-scale LLMs.
\textbf{(ii)} An \textbf{extension of SAE-based feature discovery to hypothesis-driven testing} in a finite logical domain.
% \cite{engels2025not}
\textbf{(iii)} \rev{An interpretability-based demonstration that \textbf{RL post-training} neither adds nor removes concept information: it \textbf{reorganises which features dominate the representation}, with causal interventions locating the decision downstream of it, in the late-layer state.}
% These results highlight the need to redefine “reasoning” in terms of representational change, not quantitative results.
\end{abstract}

\section{Introduction}

Large language models (LLMs) are frequently described as acquiring reasoning abilities through reinforcement learning with chain-of-thought  \citep{guo2025deepseek,tang2024generalized,wang2025emergent}. 
Gains on reasoning benchmarks are widely celebrated as evidence of higher-order cognitive skills \citep{li2025system,xu2025towards,sun2025game}. 
At the same time, small reasoning models,often with just a few billion or fewer parameters, are increasingly being embedded into agentic systems and reasoning tasks, offering efficient, modular alternatives for structured decision-making and autonomous workflows \cite{belcak2025small, wang2025short, mitra2023orca}.
% However, whether these improvements correspond to genuine reasoning or instead reflect superficial adaptations remains an open question.
This raises an important question: whether RL training on small models truly improves reasoning abilities or instead reflects superficial adaptations to the reasoning training tasks. 
A growing body of work warns that LLM reasoning may be overstated: models can exploit shallow patterns, perform reward hacking, overfit to training data, or depend on distributional cues \citep{hua2024disentangling,xie2024memorization,wu2025knowledge}. 
Counterfactual and symbolic-perturbation studies repeatedly find that accuracy degrades substantially when presentation changes while the underlying logical content does not \citep{mirzadeh2024gsmsymbolic,wu2024counterfactual,dziri2023faith}.

For systematically evaluating reasoning capabilities, games can provide a structured domain. 
From Othello and Checkers to Go, grid-worlds and board games have long served as testbeds for systematic evaluations of logical reasoning \citep{joshi2024checkersgpt,spies2024transformers,dao2025alphamaze}. 
Studies of LLMs in these domains reveal that while models can track state and generate legal moves or actions, they often fail to capture higher-order concepts such as symmetries, forks or long-horizon threats \citep{yang2024reasoning,wu2024enhance,zhang2024llm}. 
This raises whether RL agents in games are learning to reason, or just extracting surface-level features.

Mechanistic interpretability offers a way to answer this question.
Sparse autoencoders (SAEs) and related methods make it possible to identify feature circuits, track representation dynamics and gauge abstraction in models \citep{templeton2024scaling,galichin2025have,karvonen2024measuring}.
Empirical studies of model world representations suggest that LLMs do encode structured features \citep{hendel2023context,gurnee2024language,engels2025not}, but whether these support genuine reasoning remains debated \citep{venhoff2025understanding,ma2026general}.
What is missing is a direct mechanistic readout of what RL training itself changes inside the model. Prior interpretability work has focused on the final trained model artifacts, while prior RL-reasoning work has focused on behaviour rather than internals. We close that gap on a fully solvable domain, tracing representation change \emph{across} the GRPO training, not just at its endpoints.

% In this work, we \textbf{combine reinforcement learning with interpretability to investigate reasoning in small LLMs on the closed logical domain of Tic-Tac-Toe.}
% This simplicity of the game allows us to fully enumerate game states, rigorously control for symmetry, and distinguish between shallow features (turn order, board occupancy) and higher-order abstractions (strategic threats, forks, symmetries).
% In this controlled setting, we train small yet widely used instruction-tuned models (0.5--1.5B parameters) for reasoning over the logical domain of tic-tac-toe. 
% We train models using GRPO, and probe their internal representations using SAEs.
% By analyzing activations during training, we \textbf{trace how reinforcement learning changes the internal representations.}
% % They do not directly speak to frontier-scale RLHF, to other RL objectives such as PPO or DPO, or to open-ended natural-language reasoning. 
% The exploration-vs-exploitation of RL post-training has been studied at the policy level. Our contribution is at the representation level. We trace, with SAEs and a controlled hypothesis set, how popular reasoning post-training recipes like GRPO actually change model representation dynamics, and whether quantitative performance improvements truly entail the emergence of higher order reasoning in models.

In this work, we \textbf{combine reinforcement learning with interpretability to \rev{trace which internal representations emerge and change during RL training} in small LLMs on the closed logical domain of Tic-Tac-Toe}\rev{, and ask whether those representations support reasoning or reflect adaptation to prompt-level features}. The game's simplicity lets us fully enumerate states, rigorously control for symmetry, and cleanly distinguish shallow features (turn order, board occupancy) from higher-order abstractions (strategic threats, forks, symmetries). We train widely-used instruction-tuned models (0.5--1.5B parameters) with GRPO and probe their internal representations with SAEs at each checkpoint, \textbf{tracing how reinforcement learning changes the representations across training}. The exploration-vs-exploitation question for RL post-training has been studied at the policy level. Our contribution is at the representation level: using SAEs and a controlled hypothesis set, we measure how popular post-training recipes like GRPO actually reshape representation dynamics, and whether quantitative performance gains truly entail the emergence of higher-order reasoning in small reasoning models.

Our findings are that \textbf{reinforcement learning dramatically improves quantitative performance} compared to supervised fine-tuning, consistent with prior reports \citep{dang2025reinforcement,srivastava2025towards}. 
However, \textbf{mechanistic analysis shows that the improvements arise from improved usage of prompt-level features, not the development of strategic higher level reasoning.} 
In other words, models become better at exploiting what is already present in the input, rather than reasoning beyond it.
% Our study highlights the need for interpretability-first approaches to evaluating reasoning in language models.

\section{Related Work}

\begin{figure*}[!t]
    \centering
    \includegraphics[height=0.2\textheight]{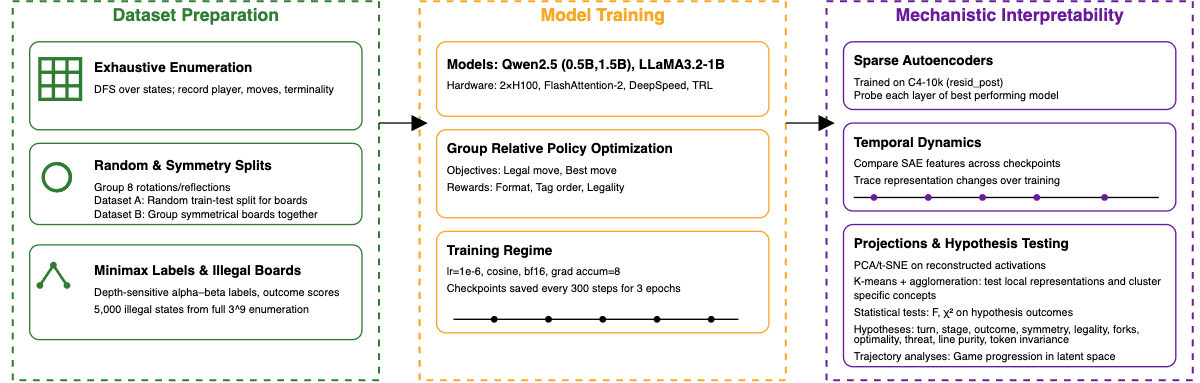}
    \caption{Overview of our experimental setup}
    \label{fig:methodology}
\end{figure*}

\paragraph{Reasoning in language models.}
Surveys consolidate a fast-growing literature on LLM reasoning spanning symbolic logic, mathematics and games \citep{li2025system,xu2025towards,gallotta2024large}, and reinforcement learning is widely credited with unlocking reasoning beyond supervised fine-tuning \citep{guo2025deepseek,tang2024generalized,dang2025reinforcement}.
Critics counter that these benchmarks reward task-specific shortcuts rather than structural reasoning \citep{hua2024disentangling,zhao2026chain,wu2025knowledge}, and that failures in self-verification and abstraction expose how fragile the claims are \citep{stechly2025self,Hazra2025HaveLL,toh2025jumping}. We ask what this training changes inside the model.

\paragraph{Games as reasoning benchmarks.}
Games give structured testbeds where reasoning can be precisely defined. OthelloGPT showed that transformers track state and legal moves while failing to generalize abstractions such as board symmetries \citep{he2024dictionary}, with similar results in checkers, Gomoku and maze solving \citep{joshi2024checkersgpt,todd2024gavel,dao2025alphamaze}. Models track local features but stay brittle on forks, forced wins and symmetry invariance \citep{chen2024llmarena,zhang2024llm}, which motivates our mechanistic approach.

\paragraph{Mechanistic interpretability.}
Sparse autoencoders (SAEs) have become a central method for opening the black box of LLMs, allowing the discovery of monosemantic features \citep{huben2024sparse,templeton2024scaling,marks2025sparse} and making it possible to trace how features evolve with training, building on dictionary-learning approaches \citep{faruqui2015sparse,karvonen2024measuring}. Applied to reasoning domains, SAEs reveal that LLMs encode shallow patterns more readily than abstract structures \citep{he2024dictionary,spies2024transformers}, raising the possibility that reinforcement learning amplifies prompt-level features without inducing genuine reasoning.

\begin{revblock}
\paragraph{Relation to the closest prior work.}
Our study differs from the nearest interpretability work in what is under the microscope. \citet{karvonen2024measuring} benchmark dictionary-learning quality on from-scratch board-game models, so their object of study is the SAE method itself, whereas ours is RL post-training of instruction-tuned LLMs with SAEs as the instrument. \citet{he2024dictionary} analyse a from-scratch OthelloGPT at a single endpoint, whereas we compare checkpoints along the training trajectory. \citet{galichin2025have} and \citet{venhoff2025understanding} find reasoning-associated features in already-trained models, and our results refine theirs: those features are discourse-level, while the ones RL strengthens here are prompt-level.
\end{revblock}

% TODO: Move this to appendix
% \paragraph{Model world representations.}  
% Beyond games, studies show that LLMs can learn structured internal models of text and environment. Context vectors, task vectors, and emergent representations highlight the ability of models to organize knowledge in ways resembling world models \citep{li2023emergent,hendel2023context,todd2023function,gurnee2024language,,engels2025not,burns2022discovering,kadavath2022language,liu2022transformers,AIGuide2023LLMsWorldModels}. Yet whether these representations enable reasoning or simply encode correlations is actively debated \citep{venhoff2025understanding,wang2025resa,ma2026general,wu2025knowledge}. Our work contributes to this debate by tracing how reinforcement learning reshapes world representations in a controlled logical setting.

\section{Methodology}
\label{sec:method}

\begin{figure*}[!t]
    \centering
    \includegraphics[width=1.0\textwidth]{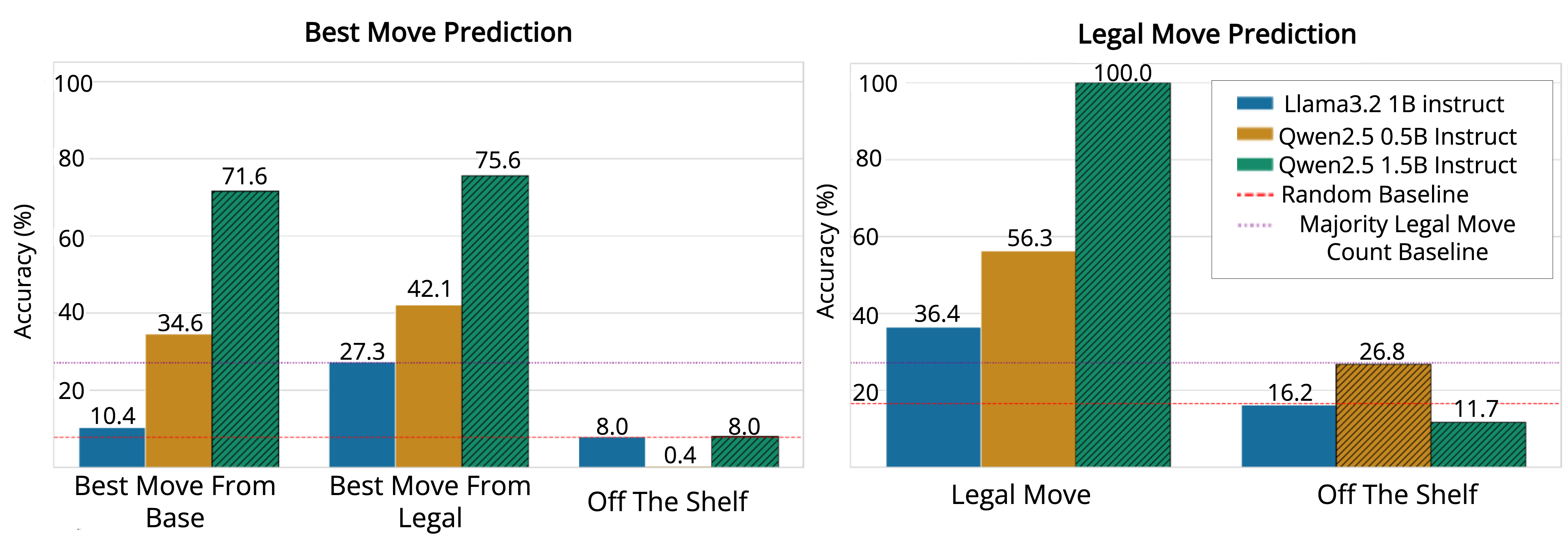}
    \caption{Best performance of each model and training setting for the legal move and best move prediction tasks on the random-split dataset. Complete performance plots for all settings (different board and move token representations) and datasets can be found in Appendix Figure \ref{fig:appendix-overall}.}
    \label{fig:quant-overall}
\end{figure*}

\begin{figure*}[!t]
    \centering
    \includegraphics[width=1.0\textwidth]{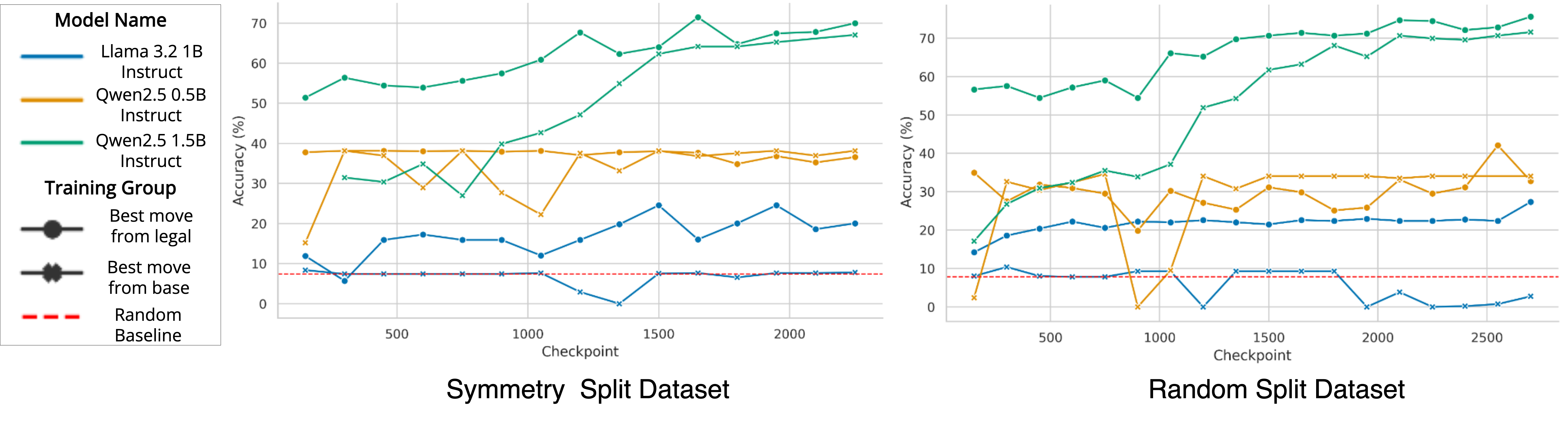}
    \caption{Progression of model performance across model checkpoints for the two datasets. Complete progression plots for all settings (different board and move token representations) can be found in Appendix Figure \ref{fig:appendix-progression}.}
    \label{fig:quant-progression}
\end{figure*}

We study reasoning development under reinforcement learning in small LLMs using a controlled, fully interpretable environment: Tic-Tac-Toe. Figure \ref{fig:methodology} describes our experimental setup, which was broadly split into 3 parts: dataset preparation, model training and mechanistic analysis.

\subsection{Models and Training}

Experiments used three models: Qwen2.5 0.5B Instruct, Qwen2.5 1.5B Instruct \cite{qwen2.5}, and Llama3.2 1B Instruct \citep{grattafiori2024llama}. Models were trained with Group Relative Policy Optimization (GRPO) \citep{tang2024generalized,guo2025deepseek}. Supervised fine-tuning (SFT) was attempted as a baseline but failed, with models repeating the input prompt rather than learning valid continuations \citep{dang2025reinforcement,srivastava2025towards}.  

Training was performed on 2$\times$H100 GPUs with \texttt{Transformers reinforcement learning} library \citep{vonwerra2022trl}, \texttt{flashattention-2} \citep{dao2023flashattention} and DeepSpeed \citep{rasley2020deepspeed}. Hyperparameters: learning rate $1\times10^{-6}$, cosine scheduler, bf16 precision, gradient accumulation 8. Checkpoints were saved every 300 steps up to 2240 steps (3 epochs).

\subsection{Datasets}

Two datasets were constructed. \textbf{Random split} with 80-10-10 train/validation/test partition over all legal Tic-Tac-Toe states and \textbf{Canonical symmetry split} where all eight symmetry variants of a board (rotations and reflections) were grouped by canonical ID. Each symmetry class was placed entirely in train or test, preventing symmetry leakage.
\rev{We treat the two splits as separate designs answering different questions. The random split measures performance when symmetric variants of a test board are available in training, while the canonical split measures generalisation to unseen symmetry groups. 
% We therefore interpret each curve (Fig.~\ref{fig:quant-progression}) on its own terms rather than as a controlled effect estimate. 
Full dataset statistics are provided in Table~\ref{tab:dataset-stats}.}

% \begin{itemize}
%     \item \textbf{Random split:} 80-10-10 train/validation/test partition over all legal Tic-Tac-Toe states.
%     \item \textbf{Canonical symmetry split:} all eight symmetry variants of a board (rotations and reflections) were grouped by canonical ID. Each symmetry class was placed entirely in train or test, preventing symmetry leakage.
% \end{itemize}

\paragraph{Generation.} States were produced via exhaustive depth-first traversal from the empty board. Each state records: current player, legal moves, terminality, minimax-computed best moves, and canonical symmetry ID. Player moves were encoded as integers: 1–9 for X, 10–18 for O. Illegal boards  (5,000) for hypothesis testing were added by enumerating all $3^9$ possible states and discarding unreachable ones. (Details of algorithm described in App. \ref{alg:illegal_legal})

\paragraph{Dataset fields.} Each entry includes: board (numeric and ASCII), natural language descriptions, move sequences, outcome labels, legal moves, best moves, and symmetry identifiers.

\subsection{Objectives and Rewards}

We trained models under two objectives. Given a board state, models had to either: Predict a legal move (legal move experiments) or, predict the most optimal move (best move experiments). 
We also included standard format rewards: (i) legality, (ii) format compliance (\texttt{<think>}...\texttt{</think><answer>}...\texttt{</answer>}), (iii) single-tag correctness. 
Rewards were equally weighted. 
Best-move training was initialized either from scratch or from the best legal-move checkpoint of the model for the given dataset (random or symmetry) to compare the difference in performance and convergence speeds.
\rev{Although each step predicts a single move, the objective spans the full game. The labels come from full-depth minimax (Algo.~\ref{alg:minimax}) over an exhaustive state space, so every decision a complete game could visit is present with its optimal label. We also verify multi-turn play explicitly in App.~\ref{sec:arena-appendix}.}

\subsection{Evaluation}

At each checkpoint, we evaluated on both natural language and ASCII board formats, with robustness tested by \rev{\emph{random symbol substitution}, replacing the player symbols X and O with random characters}.
% Metrics included Accuracy, Outcome score (minimax evaluation from current board state), Game phase accuracy (early, mid, late) ,Branching factor (legal move count), Best-move multiplicity. 
Baseline models used structured generation to generate a chain-of-thought reasoning string followed by a move prediction.

\subsection{Mechanistic Interpretability}

To analyze internal representations we used sparse autoencoders (SAEs) \citep{huben2024sparse,templeton2024scaling,marks2025sparse}, trained on a generic corpus (\texttt{NeelNanda/c4-10k}) to extract general-purpose features and then applied to hypothesis-driven testing in our controlled domain (Algo.~\ref{alg:sae_discovery}).
We train on a generic distribution rather than on Tic-Tac-Toe activations, since \citet{engels2025not} show that SAEs trained on narrow corpora produce features that look clean but mirror the corpus rather than the model's internal geometry, so in-domain training could imprint task-specific structure onto the dictionary. Generic-corpus SAEs instead read out which concepts the model already represents, and the domain lets hypotheses about symmetry, strategy and game dynamics be defined exactly.

For each layer of each model, we trained SAEs on activations (\texttt{resid\_post}) using the configuration in \citet{engels2025not}. 
Reconstructed activations were projected using PCA and t-SNE for visualization (Algo.~\ref{alg:tsne}). Board states in the projections were labeled along multiple axes: player turn, game stage, outcome, strategic situation (Algo.~\ref{alg:strategic_situation}), symmetry group, legality, and correctness of model predictions. 
To test the models' capabilities on general board and game understanding, the representations of the illegal boards were compared with those of the legal boards used for training and testing. 
Similarly to quantitative evaluation, model representations were also tested for robustness using token invariance. \rev{The player symbols X and O were replaced with the fixed pair P and Q (a second variant uses A and B)} to check if the models can generalize their developed representations to variances in prompt input without changes in the task setting. The complete set of algorithms for hypothesis testing can be found in the Appendix section \ref{sec:algorithms-hypothesis}.
% Then, hypothesis-based evaluations were conducted. This involved comparing the same projections across multiple hypotheses to identify potential concepts emerging in the model which were consistent with the updates of board representations across training. 
Clustering with k-means and hierarchical agglomeration was performed to automatically identify the feature groups, and statistical tests ($F$-tests, $\chi^2$) evaluated the dependence between clusters and the previously defined game hypotheses. Manual analysis of the concepts and boards was done to identify higher level patterns. 
\rev{\textbf{The resulting corpus spans eight checkpoints, eleven layers each and eight hypothesis-family tests per configuration.} We find the decodability results replicate across all three models (App.~\ref{sec:probe-appendix}), and we report the dominant-axis reorganisation for Qwen2.5-1.5B, the best performing model, where it is clearest.}

% \FloatBarrier

\begin{figure*}[!t]
    \centering
    \includegraphics[width=1.0\textwidth]{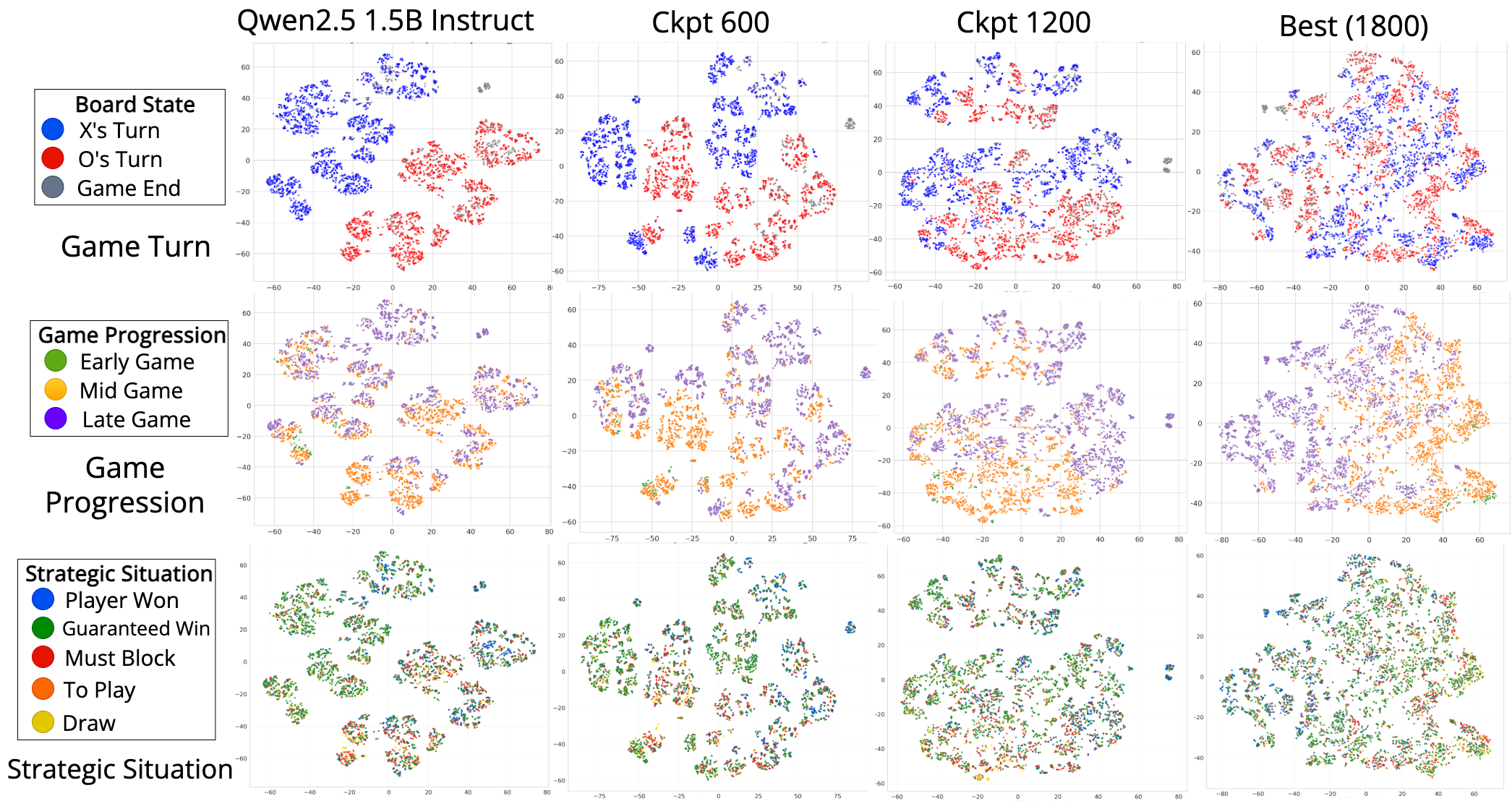}
    \caption{Model representations dynamics of Qwen2.5 1.5B (layer 12) through reasoning (GRPO) training. The figure represents the same projections colored based on three hypotheses: \textbf{(H1)} whose turn it is (Game Turn), \textbf{(H2)} which stage of the game it is (Game Progression) and \textbf{(H3)} what is the logical play for the given board (Strategic Situation). It is evident that while the base model only clearly represents the game turn, during the course of training, the model evolves to represents the boards based on game progression in addition to game turn, leading to improvement in performance. However, logic for strategic situations does not play a big role in model representations and no clear patterns emerge over the course of training. Full training dynamics across all layers and over all hypotheses tested are provided in Appendix Section~\ref{sec:mech-interp-appendix}.}
    \label{fig:interp-comparison}
\end{figure*}

\section{Results}

\begin{table*}[!htbp]
\centering\footnotesize
\setlength{\tabcolsep}{3pt}
\input{revisions/tables/sae/cluster_stats_combined.tex}
\caption{$F$-statistic on the principal t-SNE axis (Qwen2.5-1.5B, NL). Larger $F$ means hypothesis-labelled boards separate more cleanly on the dominant axis. \emph{Trajectory} (left) is the GRPO-from-base run at layer 12: H1 mid-layer F drops $17\times$ and re-emerges in late layers and H2 progression grows $8.5\times$, while H3/H4 stay weak. \emph{Per-layer} (right) is the converged checkpoint: H2 peaks at \textbf{layer 12}, H4 stays at noise at every layer, and late layers (L17--L24) recover H1 for output decoding. The mid-layer locus is a property of the reorganisation, not of layer choice. 
% Cram\'er's $V$ and AMI details are in App.~Table~\ref{tab:cluster-main-snapshot}.
}
\label{tab:cluster-stats-combined}
\end{table*}
% We begin with the quantitative results: performance curves across objectives, scales, and modalities, which sets the stage for how much reinforcement learning improves over supervised fine-tuning, and where it falls short. We then dive deeper with mechanistic analysis, showing how internal representations change under GRPO, which features are strengthened, and which never emerge. Together, these perspectives form a coherent picture: reinforcement learning amplifies prompt-level cues but leaves higher-order abstractions largely untouched.
We present quantitative results first, then the mechanistic analysis. Together they give a coherent picture: \textbf{reinforcement learning amplifies prompt-level cues but leaves higher-order abstractions largely untouched}.

\subsection{Quantitative Results}

\paragraph{Summary.}
Across three models, two objectives, two splits and four input modalities, GRPO improves performance over off-the-shelf and SFT baselines, most on natural-language prompts, while symmetry-controlled curves closely track random-split curves (Fig.~\ref{fig:quant-progression}), \textbf{indicating limited use of symmetry-aware structure.} Further ablations are in App.~\ref{sec:appendix-quant} and~\ref{sec:appendix-additional-behaviour}.

\paragraph{Input condition ablations reveal modality dependence.}
Performance is \emph{not} consistent across input modalities. \textbf{ASCII board representations} induce the \textbf{largest accuracy drop} for both objectives across all models, with the best performing Qwen2.5 1.5B model losing a relative 40\% (73\% to 44\%) against natural language (full reports in App.~Fig.~\ref{fig:appendix-overall}). 
\textbf{Random symbol substitution} (symbol swap in prompts) causes a \textbf{smaller, consistent} decrease relative to the standard NL condition.
The \textbf{combined} (ASCII + random symbol substitution) setting yields the \textbf{most severe degradation}, often approaching baseline.
% \end{itemize}
These asymmetries indicate that GRPO improved robustness to superficial \emph{token} perturbations but did \emph{not} produce modality-invariant board understanding: the models rely on the natural-language scaffold to parse state, and accuracy drops steeply when forced to build a spatial map from ASCII. This adds evidence to reports that LLM “reasoning’’ is entangled with input format and context cues rather than abstract structure \citep{hua2024disentangling,wu2025knowledge,cosentino2024reasoning}.

\paragraph{Symmetry split has minimal impact on learning curves.}
Learning curves under symmetry-controlled splitting remain close to random-split curves (Fig.~\ref{fig:quant-progression}). 
If models learned symmetry-aware abstractions, symmetry-based training should change performance.
\rev{Instead, we measure this directly and find the models pick corresponding moves on symmetry-related boards barely half the time, $56.7\%$ on fully trained symmetry groups against $57.7\%$ on never-seen ones (App.~\ref{sec:equivariance-appendix}).}

\begin{revblock}
\paragraph{Terminal-state detection, not easy positions, is learned first.}
Raw accuracy across difficulty bins is misleading, because the chance level itself varies with the number of legal moves. We therefore score each non-terminal state against a uniform-random legal-move policy. For Qwen2.5-1.5B at the best-move checkpoint, low-branching positions ($\le 2$ legal moves) reach $83.1\%$ against a $76.0\%$ chance level, high-branching positions ($\ge 5$) reach $63.6\%$ against $39.2\%$. Normalized for chance, skill is $29.7\%$ on low-branching versus $40.2\%$ on high-branching positions, so the apparent easy-position advantage is a chance artifact. What the models do acquire immediately is terminal-state detection ($97.8\%$ already at the first checkpoint), while we see above-chance move selection emerge only gradually and at a similar rate across bins. The instantly acquired competence is recognition of a prompt-visible property (a completed line), consistent with GRPO strengthening reliance on prompt-level cues rather than instilling search. Per-position breakdowns are in App.~\ref{sec:per-position-app}.
\end{revblock}

\paragraph{Format alone does not explain GRPO gains.}
Forcing off-the-shelf models to emit the exact required \texttt{<think>}/\texttt{<answer>} template through structured generation isolates the format component of the reward. Under that identical template they reach $0.4$--$8.0\%$ best-move accuracy against $27$--$76\%$ for GRPO models, which rules out format compliance as the primary driver (App.~\ref{sec:reward-decomp}).
% Per-architecture and per-input-modality numbers are in Appendix~\ref{sec:reward-decomp}, Table~\ref{tab:reward-full}.

\begin{revblock}
\paragraph{Representation and behaviour come apart out of distribution.}
We run an out-of-distribution test using the unreachable (illegal) boards, which no model was trained on. Probes separate them from legal boards at AUC $1.00$ at every layer, so the representation registers that these states are unlike training data, yet the policy emits ordinary legal continuations on them exactly as it does on legal boards. Information present in the representation is therefore not information the behaviour acts on, the same gap the mechanistic analysis below documents for strategic structure (App.~\ref{sec:illegal-appendix}).
\end{revblock}

\begin{figure*}[!t]
    \centering
    \includegraphics[width=0.7\textwidth]{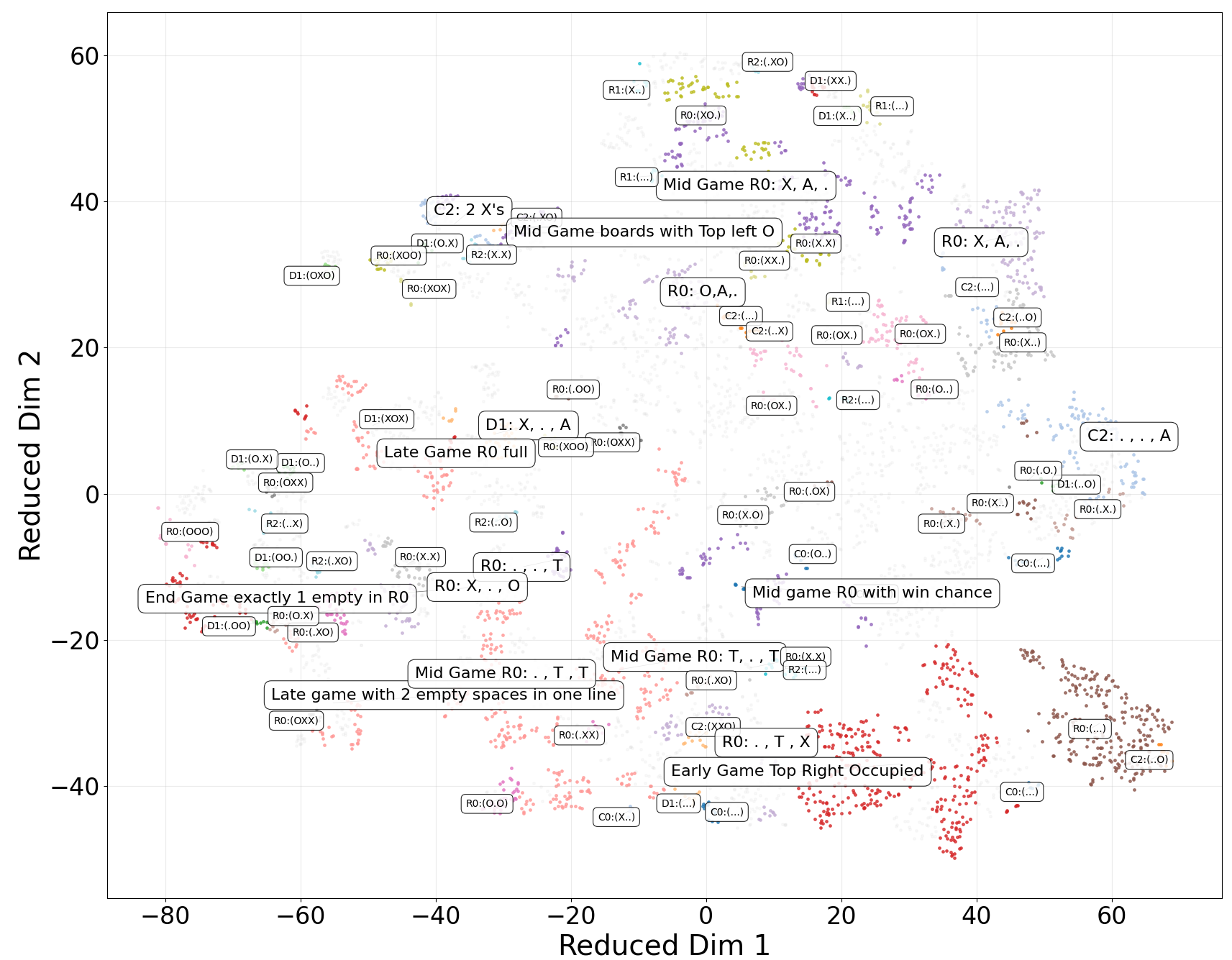}
    % \caption{Clustering of board state representations based on structural line purity for Qwen2.5 1.5B (layer 12) on text representations of the boards. The labels represent the dominant line purity sub-clusters or the common line pattern across the board states. The smaller labels represent the automatically mined line subclusters within k-means clusters found through agglomerative clustering. For clarity, we include manually annotated clusters which combine automatically mined subclusters across different k-means clusters. This shows how common structural patterns are represented in local subclusters while the general representation of the boards follows the game progression as found in Figure~\ref{fig:interp-comparison}. For example, late game board states which contain exactly one empty cell in the top row are all clustered together.  In the labels, R, C and D represent rows, columns and diagonals respectively. O and X represent player tokens, ``.'' represents empty, T represents taken (by either X or O) and A represents any of X, O or empty. The complete map of automatically mined line purity clusters for all layers and models across all representation modes and training checkpoints can be found in Appendix Figure~\ref{fig:appendix-interp-dominant-line}.}
    \caption{Clustering of board state representations based on structural line purity for Qwen2.5 1.5B (layer 12) on text-form boards. Labels mark dominant line-purity clusters (common row, column, or diagonal patterns), with smaller labels denoting automatically mined subclusters from agglomerative k-means. Manually annotated clusters merge related subclusters across groups for clarity. For instance, late-game states with a single empty cell in the top row cluster together. In labels, R/C/D denote rows, columns, and diagonals. O and X are player tokens. ``.'' is empty, T is taken (either player), A is any of X/O/empty. Full cluster maps for all layers, models, and modalities appear in Appendix Figure \ref{fig:appendix-interp-dominant-line}. \rev{The key finding is that these local structures are already present in the base model and are retained through GRPO, while the macroscopic arrangement is what changes, acquiring the game-progression axis. Symmetry-class membership never organises the clusters.}}
    \label{fig:interp-clusters}
\end{figure*}

\subsection{Mechanistic Analysis}
We group hypotheses into two classes: 
\emph{Prompt-level} features explicitly available from the input (\textbf{H1} turn identity; \textbf{H2} game progression), and \emph{high-level} game abstractions (\textbf{H3} strategic situation: must-block, guaranteed win, etc.; \textbf{H4} symmetry class; \textbf{H5} legality; \textbf{H6} best-move correctness; \textbf{H7} line-purity). 
Game progression bins use move counts: early $=0\!-\!3$, mid $=4\!-\!6$, late $=7\!-\!9$. 
The controlled domain allows hypothesis-driven tests that are hard to do in open-ended corpora.
GRPO increases separability for H1 turn identity and H2 progression, but does not produce clean structure for H3 strategic situation, H4 symmetry, or H6 optimality (Fig.~\ref{fig:interp-comparison}). Table~\ref{tab:cluster-stats-combined} reports the one-way ANOVA $F$ on the dominant t-SNE axis. \rev{We compute these over all 5,478 states, so the degrees of freedom follow from the class counts, $(1,5476)$ for H1 and $(2,5475)$ for H2, and every $F$ we report is significant at $p<10^{-68}$ (App.~\ref{sec:cluster-stats}). Since that holds of every value in the table, we rest the argument on relative magnitudes, on the effect sizes in Table~\ref{tab:cluster-main-snapshot} and on the assumption-free permutation tests in App.~\ref{sec:probe-appendix}.} We see that \textbf{H1 turn representation decreases $17\times$}
% ($F\!=\!5285\!\to\!316$)
and \textbf{H2 progression representation increases $8.5\times$}
% ($274\!\to\!2330$)
over GRPO training, while H3 and H4 stay at noise level. \rev{H5 and H6 are quantified in the same form in App.~\ref{sec:probe-appendix}. The asymmetry also appears behaviourally: terminal-state detection is near ceiling from the first checkpoint, whereas chance-corrected differences across branching and tie structure are small.}
% Prompt-level cues are amplified by an order of magnitude. Strategic abstraction is not.
Prompt-level structure (H2) is thus amplified by an order of magnitude and concentrated in mid layers, while strategic and symmetry abstractions (H3, H4) are not, at any layer or point in training.

\begin{revblock}
\paragraph{Full-space probes reproduce the result without the SAE.}
The SAE pipeline interprets concepts through a dictionary, a projection and clustering. As an independent test, we repeat every hypothesis test with classifiers trained on the raw residual stream, with none of those steps in the path (Table~\ref{tab:probe-body}, with the protocol and cross-model replication provided in App.~\ref{sec:probe-appendix}). We find the prompt-level properties already at ceiling in the base model, so what GRPO amplifies is their prominence in the representation rather than their presence.
\end{revblock}

\begin{table}[!ht]
\centering\small
\rev{
\setlength{\tabcolsep}{4pt}
\begin{tabular}{lcc}
\toprule
Hypothesis & Base & Final ckpt \\
\midrule
H1 turn identity & $1.000$ & $1.000$ \\
H2 game progression & $1.000$ & $1.000$ \\
H5 legality & $1.000$ & $1.000$ \\
H7 line purity & $0.861$ & $0.856$ \\
H3 strategic situation & $0.686$ & $0.684$ \\
H4 symmetry-group similarity & $0.70$ & $0.70$ \\
\bottomrule
\end{tabular}}
\caption{\rev{Probe decodability (cross-validated AUC, middle layer of Qwen2.5-1.5B) read from raw activations. Prompt-level properties (H1, H2) and legality (H5) are already perfectly decodable in the base model, so GRPO cannot be adding missing information, and the properties with headroom (H3, H4) do not move across training. A non-linear MLP probe raises H3 by at most $0.06$ and likewise does not move it across training, so GRPO changes which features dominate the representation, not which concepts are recoverable.}}
\label{tab:probe-body}
\end{table}

\subsubsection{Macroscopic Representation Changes under GRPO}
\label{subsec:geom}

% Fig.~\ref{fig:interp-comparison} shows how the internal representation changes over the course of training the best-performing model (Qwen2.5 1.5B), specifically for layer 12. 
% Each row shows different colorings of all the possible legal board states, with each coloring representing a hypothesis. 
% For the top row, the boards are colored by \textbf{H1}: red and blue depending on whose turn it is, and grey when the game has ended. The middle rows colors boards by the \textbf{H2} (game progression: current turn number, out of the max possible 9 turns in a game). The bottom row colors boards by \textbf{H3} (strategic situation, e.g. whether a player has won already, is guaranteed a win no matter what, must block in order not to lose, etc.)

\paragraph{Turn and progression are the first-class axes.}
At initialization, clusters align most with \textbf{H1} (whose turn), a strong axis already present in base models. With GRPO a second macroscopic axis for \textbf{H2} (game progression) emerges and stretches across the manifold (Fig.~\ref{fig:interp-comparison}, middle row), consistent with reports that small LLMs first acquire representations tied to immediately accessible features \citep{spies2024transformers,gurnee2024language}. 

\paragraph{Strategic abstractions do not emerge.}
Labelling by \textbf{H3} (strategic situation, e.g. must block, guaranteed win) yields scattered speckles with no clean clusters even at peak performance late in training (Fig.~\ref{fig:interp-comparison}, bottom row). Occasional micro-islands appear for \textit{immediate} tactics such as one-move wins, but these may track structural organization of the boards rather than logical abstraction.
% There are no larger coherent sheets that would indicate a compact basis for long-horizon strategy or symmetry invariance.
% This mirrors Othello-style observations that legal tracking emerges before abstract invariances \citep{he2024dictionary} and relates to geometric descriptions of reasoning that distinguish shallow separators from global structure \citep{cosentino2024reasoning}.

\paragraph{Input structure and local clusters.}
\rev{We see the ASCII drop mechanistically as well: crisp SAE detectors aligned to lexical markers support H1/H2 under natural-language boards, but the base-model turn pattern is absent under ASCII. Beneath the macroscopic axes we recover local pockets of high line purity (Fig.~\ref{fig:interp-clusters}), already present in the base model and retained after training, while symmetry-class membership never organises the clusters (App.~\ref{sec:local-structure-appendix}).}

% Move this after the figure to bring the table up
\subsubsection{Layerwise representational change}
\label{subsec:layer-dynamics}

We locate the most substantial reorganization in the \emph{middle layers}, around layer 12 in Qwen2.5-1.5B. Early and late layers stay comparatively stable across GRPO training, as the layerwise dynamics map in App.~\ref{sec:mech-interp-appendix} shows.
\rev{Middle layers gain a well-defined game-progression axis while turn structure decreases mid-network, the layer-12 turn $F$ falling from 5285 to 316 (Table~\ref{tab:cluster-stats-combined}). Turn structure re-emerges in the late layers of the converged checkpoint ($F$ of 2208, 5379 and 5484 at layers 17, 18 and 24), where it is decoded for output.}

% This aligns with prior work that early layers often specialize in lexical or surface encoding, while middle layers form reusable abstractions, and later layers map abstractions to task-specific outputs \citep{, , elhage2022toy}. In reinforcement learning fine-tuning, middle layers are also where preference-aligned features tend to emerge, with output layers primarily adjusting stylistic or formatting control \citep{tang2024generalized,stechly2025self}. Sparse autoencoder studies likewise find that mid-layer dictionaries yield the most interpretable, monosemantic features \citep{huben2024sparse,templeton2024scaling,marks2025sparse}, whereas later layers contain highly entangled, task-specific mixtures that are harder to disentangle \citep{Marks2024ReconstructionErrors,Foote2024FeatureSuppression}.

GRPO updates therefore reshape mid-layer manifolds to better capture prompt-level structure while leaving early and late layers comparatively unchanged across training. This explains why models gain robustness to prompt variation without forming new higher-order abstractions: the learning signal improves already-accessible mid-layer features rather than rewriting the pipeline. Two explanations (reward shape, architectural compute) are in App.~\ref{sec:extended-limitations}.

\begin{revblock}
\subsubsection{Causal interventions locate the decision in late layers}
\label{subsec:causal}
Feature presence does not establish feature use, so we perform causal intervention through activation patching (Algo.~\ref{alg:intervention}) on 100 minimal board pairs whose optimal moves differ (App.~\ref{sec:causal-appendix}). Running the model on one board while overwriting the residual stream at the answer position with the paired board's activations flips the prediction to the paired optimum in 66 of 100 pairs at layer 25 of 28, against an unpatched rate of 10 ($p<10^{-14}$). Two controls establish that the donor's board content is what transfers: a random unrelated donor moves the prediction in only 12 of 100 cases, and the same patch at layers 13, 17 and 21 leaves behaviour unchanged, so the effect is not generic disruption. Steering along a mid-layer turn direction likewise leaves the output intact where norm-matched random directions at the same magnitude disrupt it, so the reorganised mid-layer axis is not what the move is computed from. Both interventions converge on the same picture: the late-layer state in which turn structure re-emerges is where the board is decoded into a move, which is what the output-decoding interpretation of Table~\ref{tab:cluster-stats-combined} predicts.
\end{revblock}
%%% --------------------------- END ADJUSTABLE BLOCK --------------------------

\section{Conclusion}

We provide the first mechanistic trace of RL based reasoning training dynamics on small LLMs in a fully analyzable logical domain.
\rev{RL post-training neither adds nor removes concept information: it reorganises which features dominate the representation mid-network, while the decision is carried downstream in the late-layer state. Gains thus stem from heavier use of prompt-level cues, not from new higher-order abstractions.} We thus provide an interpretability-first blueprint for studying reasoning in controlled settings.

\clearpage

\section*{Limitations}

We highlight limitations directly relevant to interpreting our claims. Further discussion is in Appendix~\ref{sec:extended-limitations}.

\paragraph{Scope: small models, GRPO, one solvable domain.}
Our findings are specific to 0.5--1.5B-parameter instruction-tuned models post-trained with GRPO on Tic-Tac-Toe. Mid-layer reorganisation and prompt-level adaptation mirror trends reported in larger models. We do not claim to characterise frontier-scale RLHF, other RL objectives such as PPO or DPO, or open-ended reasoning. \rev{Within the scope we do claim, the pattern holds across two model families and a threefold range of scale, varying data composition and initialisation rather than the training seed.} Extension to richer logical domains, such as Connect Four, simple algebra, and grid-world planning, and to alternative RL recipes is a natural next step.

\begin{revblock}
\paragraph{Where representations are measured.}
We read activations at the last prompt token and at eleven of the twenty-eight layers, and we sample the training trajectory every 300 steps. Reasoning circuits that are built transiently inside the generated reasoning trace, at an unsampled layer, or appearing and reversing within a checkpoint interval would not enter this measurement. The sampled layers do span the full depth, covering both the mid-network locus of the reorganisation and the late layers where patching locates the decision.

\paragraph{A closed hypothesis set.}
Clustering is automatic, but H1--H7 are specified by us from the structure of the game. An abstraction the model does represent that falls outside this set is not tested for, so our negative results are statements about these seven hypotheses rather than about every abstraction the task allows. The set is nonetheless exhaustive for the reasoning that our chosen formulation defines, and we did not specify line purity (H7) in advance: it came out of automatic subcluster mining, so the procedure can surface structure we did not anticipate.
\end{revblock}

\clearpage


% TODO (bib hygiene, pre-existing): four cited keys are absent from verified_refs.bib and will
% render as "?" -- li2023emergent (lines ~319, ~388), Marks2024ReconstructionErrors (~430),
% wang2026tina (~775, bib has wang2026tina), nie2025diffusion (~1500, bib has nie2026large).
% TODO (bib hygiene, new): the causal-intervention appendix names Vig et al. 2020, Geiger et al.
% 2021, Meng et al. 2022, Wang et al. 2023, Zhang and Nanda 2024, Turner et al. 2023 and
% Panickssery et al. 2024 inline rather than citing them, pending bib entries.
\bibliography{verified_refs}

%%%%%%%%%%%%%%%%%%%%%%%%%%%%%%%%%%%%%%%%%%%%%%%%%%%%%%%%%%%%%%%%%%%%%%%%%%%%%%%
%%%%%%%%%%%%%%%%%%%%%%%%%%%%%%%%%%%%%%%%%%%%%%%%%%%%%%%%%%%%%%%%%%%%%%%%%%%%%%%
% APPENDIX
%%%%%%%%%%%%%%%%%%%%%%%%%%%%%%%%%%%%%%%%%%%%%%%%%%%%%%%%%%%%%%%%%%%%%%%%%%%%%%%
%%%%%%%%%%%%%%%%%%%%%%%%%%%%%%%%%%%%%%%%%%%%%%%%%%%%%%%%%%%%%%%%%%%%%%%%%%%%%%%
\newpage
\appendix
\onecolumn

\section{Algorithms for Hypothesis Testing}
\label{sec:algorithms-hypothesis}
This section details the algorithms used for Sparse Autoencoder (SAE) feature discovery, board state analysis, and hypothesis testing.

\begin{algorithm}
\caption{SAE Feature Discovery and Clustering}
\label{alg:sae_discovery}
\begin{algorithmic}[1]
\Require Language Model $\mathcal{M}$, Target Layer $L$
\Ensure Trained SAE $\mathcal{S}$, Feature Clusters $\mathcal{C}$
\State $\mathcal{S}_{path} \gets \text{Path to cached SAE for } \mathcal{M}, L$
\If{$\mathcal{S}_{path}$ exists}
    \State $\mathcal{S} \gets \text{LoadSAE}(\mathcal{S}_{path})$
\Else
    \State $\mathcal{D}_{train} \gets \text{Load text corpus (e.g., C4-10k)}$
    \State $\mathcal{A}_{train} \gets \text{Extract activations from } \mathcal{M} \text{ at layer } L \text{ on } \mathcal{D}_{train}$
    \State $\mathcal{S} \gets \text{TrainSAE}(\mathcal{A}_{train})$
    \State $\text{SaveSAE}(\mathcal{S}, \mathcal{S}_{path})$
\EndIf
\State $W_{dec} \gets \mathcal{S}.\text{decoder\_weights}$
\State $W_{norm} \gets W_{dec} / \Vert W_{dec} \Vert_2$ \Comment{Normalize feature vectors}
\State $Labels \gets \text{SpectralClustering}(W_{norm})$
\State $\mathcal{C} \gets \text{Group feature indices by } Labels$
\State \Return $\mathcal{S}, \mathcal{C}$
\end{algorithmic}
\end{algorithm}

\begin{algorithm}
\caption{t-SNE Dimensionality Reduction and Caching}
\label{alg:tsne}
\begin{algorithmic}[1]
\Require Model $\mathcal{M}$, SAE $\mathcal{S}$, Tic-Tac-Toe Dataset $\mathcal{D}$, Target Cluster Indices $\mathcal{I}_{cluster}$
\Ensure 2D Reduced Activations $R$, Associated Board Metadata $B_{meta}$
\State $Cache_{path} \gets \text{Path to cached t-SNE results for } \mathcal{M}, \text{layer}, \text{style}$
\If{$Cache_{path}$ exists}
    \State $R, B_{meta} \gets \text{LoadFromCache}(Cache_{path})$
    \State \Return $R, B_{meta}$
\EndIf
\State $B_{meta} \gets \text{Get unique boards from } \mathcal{D}$
\State $Prompts \gets \{\text{GeneratePrompt}(b) \text{ for } b \in B_{meta}\}$
\State $\mathcal{A}_{orig} \gets \text{Get activations from } \mathcal{M} \text{ for } Prompts$
\State $f_{acts} \gets \mathcal{S}.\text{encode}(\mathcal{A}_{orig})$
\State $\text{mask} \gets \text{Zeros like } f_{acts}$
\State $\text{mask}[:, \mathcal{I}_{cluster}] \gets 1$
\State $\mathcal{A}_{recon} \gets \mathcal{S}.\text{decode}(f_{acts} \odot \text{mask})$ \Comment{Filter with cluster features}
\State $R \gets \text{t-SNE}(\mathcal{A}_{recon}, \text{n\_components}=2)$
\State $\text{SaveToCache}(Cache_{path}, R, B_{meta})$
\State \Return $R, B_{meta}$
\end{algorithmic}
\end{algorithm}

\begin{algorithm}
\caption{Game-Theoretic Strategic Situation Analysis}
\label{alg:strategic_situation}
\begin{algorithmic}[1]
\Require Reduced Activations $R$, Board Metadata $B_{meta}$
\Ensure Saved plot colored by game-theoretic state
\Function{Evaluate}{$board, player$} \Comment{Memoized function}
    \State $winner, terminal \gets \text{CheckTerminal}(board)$
    \If{$terminal$}
        \State \Return $1$ if $winner=player$, $0$ if $draw$, $-1$ if $loss$
    \EndIf
    \State $best\_outcome \gets -2$ \Comment{Losing is the default}
    \For{move in LegalMoves($board$)}
        \State $next\_board \gets \text{ApplyMove}(board, move, player)$
        \State $outcome \gets \text{Evaluate}(next\_board, \text{opponent}(player))$
        \If{$outcome = -1$}
            \State \Return $1$ \Comment{Opponent loss is a win for me}
        \EndIf
        \State $best\_outcome \gets \max(best\_outcome, -outcome)$
    \EndFor
    \State \Return $best\_outcome$
\EndFunction
\State $Categories \gets []$
\For{board $b$ in $B_{meta}$}
    \State $winner, terminal \gets \text{CheckTerminal}(b)$
    \If{$terminal$}
        \State Append "Player Won" or "Draw" to $Categories$
    \Else
        \State $gt\_eval \gets \text{Evaluate}(b, \text{CurrentPlayer}(b))$
        \State $has\_threat \gets \text{OpponentHasImmediateWin}(b, \text{CurrentPlayer}(b))$
        \If{$gt\_eval = 1$}
            \State Append "Guaranteed Win"
        \ElsIf{$has\_threat$}
            \State Append "Must Block"
        \ElsIf{$gt\_eval = 0$}
            \State Append "Draw"
        \Else
            \State Append "To Play" \Comment{Forced loss, no immediate threat}
        \EndIf
    \EndIf
\EndFor
\State Plot $R$, coloring points by $Categories$. Save figure.
\end{algorithmic}
\end{algorithm}

\begin{algorithm}
\caption{Cluster-Based Statistical Analysis}
\label{alg:stat_analysis}
\begin{algorithmic}[1]
\Require Clustered Boards $\mathcal{C}_{boards}$ (map from cluster ID to board lists)
\Ensure Printed statistical test results
\State $ClusterFeatures \gets \{\}$
\For{cluster $cid$ in $\mathcal{C}_{boards}$}
    \State $Features_{cid} \gets []$
    \For{board $b$ in $\mathcal{C}_{boards}[cid]$}
        \State Append $\{\text{'piece\_count'}: \text{CountPieces}(b), \text{'center'}: \text{GetCenter}(b), \dots\}$ to $Features_{cid}$
    \EndFor
    \State $ClusterFeatures[cid] \gets Features_{cid}$
\EndFor
\State \Comment{Example for a continuous feature}
\State $Data_{anova} \gets [[\text{f['piece\_count']} \text{ for f in } ClusterFeatures[cid]] \text{ for cid in } \mathcal{C}_{boards}]$
\State $F, p \gets \text{ANOVA}(Data_{anova})$
\State Print("Piece Count", $F, p$)
\State \Comment{Example for a categorical feature}
\State $Table_{chi2} \gets \text{BuildContingencyTable}(\text{'center'}, ClusterFeatures)$
\State $\chi^2, p \gets \text{ChiSquaredTest}(Table_{chi2})$
\State Print("Center Control", $\chi^2, p$)
\end{algorithmic}
\end{algorithm}

\begin{algorithm}
\caption{Hybrid Hierarchical-Agglomerative Analysis}
\label{alg:hybrid_analysis}
\begin{algorithmic}[1]
\Require Reduced Activations $R$, Board Metadata $B_{meta}$
\Ensure Saved hybrid plot visualization
\State $L0_{labels} \gets \text{KMeans}(R, n\_clusters=18)$
\State $L0_{analysis} \gets \text{AnalyzeClusters}(L0_{labels})$ \Comment{For L0 properties}
\State $Map_{L0\_to\_Sub} \gets \{\}$
\For{cluster $cid$ from $0$ to $17$}
    \State $L0_{indices} \gets \text{Indices where } L0_{labels} = cid$
    \State $R_{sub} \gets R[L0_{indices}]$
    \State $k_{micro} \gets \max(5, \lfloor |L0_{indices}| / 10 \rfloor)$
    \State $Micro_{labels} \gets \text{KMeans}(R_{sub}, n\_clusters=k_{micro})$
    \State $PatternMap \gets \{\}$ \Comment{Map pattern key to list of global indices}
    \For{micro-cluster $mcid$ in $k_{micro}$}
        \State $MC_{indices} \gets \text{Global indices for micro-cluster } mcid$
        \State purity, pattern\_key $\gets \text{CalculateDominantLinePattern}(B_{meta}[MC_{indices}])$
        \If{purity $\geq$ THRESHOLD}
            \State Append $MC_{indices}$ to $PatternMap[\text{pattern\_key}]$
        \EndIf
    \EndFor
    \State $Map_{L0\_to\_Sub}[cid] \gets \{\text{MergeIndicesByPattern}(PatternMap)\}$
\EndFor
\State Visualize: For each L0 cluster, draw its convex hull. Inside, color and annotate each discovered pure sub-cluster from $Map_{L0\_to\_Sub}$ based on its defining line pattern.
\end{algorithmic}
\end{algorithm}

\begin{algorithm}
\caption{Illegal vs. Legal Board Comparison Analysis}
\label{alg:illegal_legal}
\begin{algorithmic}[1]
\Require Model $\mathcal{M}$, SAE $\mathcal{S}$, Legal Dataset $\mathcal{D}_L$, Illegal Dataset $\mathcal{D}_I$
\Ensure Saved comparison visualizations
\State $B_L \gets \text{Sample}(\mathcal{D}_L, N_{legal})$
\State $B_I \gets \text{Sample}(\mathcal{D}_I, N_{illegal})$
\State $B_{combined} \gets []$
\For{$b$ in $B_L$}
    \State Append $\{'board': b, 'type': 'legal'\}$ to $B_{combined}$
\EndFor
\For{$b$ in $B_I$}
    \State Append $\{'board': b, 'type': 'illegal', 'reasons': b.reasons\}$ to $B_{combined}$
\EndFor
\State $R \gets \text{Run t-SNE on SAE-reconstructed activations for } B_{combined}$ \Comment{As in Alg. \ref{alg:tsne}}
\State \textbf{Plot 1: Legality View}
\State Plot $R$, coloring points blue if 'legal' and red if 'illegal'. Save figure.
\State \textbf{Plot 2: Pattern Agglomeration View}
\State $Labels_{kmeans} \gets \text{KMeans}(R, n\_clusters=k)$
\For{cluster $cid$ from $0$ to $k-1$}
    \State $Cluster_{indices} \gets \text{Indices where } Labels_{kmeans} = cid$
    \State $PatternStats \gets \text{FindDominantLinePatterns}(B_{combined}[Cluster_{indices}])$
    \State Annotate cluster centroid with top patterns, showing their legal/illegal counts.
\EndFor
\State Plot $R$ colored by legality. Overlay cluster annotations and convex hulls. Save figure.
\end{algorithmic}
\end{algorithm}

\begin{algorithm}
\caption{Causal Intervention via Activation Patching}
\label{alg:intervention}
\begin{algorithmic}[1]
\Require Model $\mathcal{M}$, SAE $\mathcal{S}$, Dataset $\mathcal{D}$, Target Cluster Indices $\mathcal{I}_{cluster}$
\Ensure Printed intervention success/failure reports
\For{square $i$ from $0$ to $8$}
    \State $b_{dirty}, b_{clean} \gets \text{FindBoardPair}(i, \mathcal{D})$ \Comment{Boards differ only at square $i$}
    \If{no pair found}
        \State \textbf{continue}
    \EndIf
    \State $p_{dirty}, p_{clean} \gets \text{Prompt}(b_{dirty}), \text{Prompt}(b_{clean})$
    \State $a_{clean} \gets \text{GetActivation}(\mathcal{M}, p_{clean})$
    \State $f_{clean} \gets \mathcal{S}.\text{encode}(a_{clean})$
    \Function{PatchHook}{$a_{dirty\_runtime}$}
        \State $f_{dirty} \gets \mathcal{S}.\text{encode}(a_{dirty\_runtime})$
        \State $f_{dirty}[\mathcal{I}_{cluster}] \gets f_{clean}[\mathcal{I}_{cluster}]$ \Comment{The patch}
        \State \Return $\mathcal{S}.\text{decode}(f_{dirty})$
    \EndFunction
    \State $logits_{orig} \gets \mathcal{M}(p_{dirty})$
    \State $logits_{patched} \gets \mathcal{M}.\text{run\_with\_hooks}(p_{dirty}, \text{hook}=\text{PatchHook})$
    \State $move_{orig} \gets \text{GetMove}(logits_{orig})$
    \State $move_{patched} \gets \text{GetMove}(logits_{patched})$
    \State $move_{expected} \gets \text{GetBestMove}(b_{clean})$
    \State Print results comparing $move_{orig}, move_{patched}, move_{expected}$.
\EndFor
\end{algorithmic}
\end{algorithm}

\begin{algorithm}
\caption{Depth-Sensitive Minimax Evaluation}
\label{alg:minimax}
\begin{algorithmic}[1]
\Require Board state $B$, current player $P$, current depth $d$
\Ensure Game-theoretic score $s$, best move index $m$
\Function{MinimaxGetScore}{$B, P, d$}
    \State $winner, is\_terminal \gets \text{CheckWinner}(B)$
    \If{$is\_terminal$}
        \If{$winner=1$}
            \State \Return $10 - d, \text{None}$ \Comment{Faster wins are better}
        \ElsIf{$winner=2$}
            \State \Return $-10 + d, \text{None}$ \Comment{Slower losses are better}
        \Else
            \State \Return $0, \text{None}$ \Comment{Draw}
        \EndIf
    \EndIf
    \State $EmptyCells \gets \text{FindEmptyCells}(B)$
    \If{$P=1$} \Comment{Maximizing}
        \State $max\_eval \gets -\infty$, $best\_move \gets \text{None}$
        \For{move in $EmptyCells$}
            \State $B_{new} \gets \text{ApplyMove}(B, move, P)$
            \State $evaluation, \_ \gets \text{MinimaxGetScore}(B_{new}, \text{Player } 2, d+1)$
            \If{$evaluation > max\_eval$}
                \State $max\_eval \gets evaluation$, $best\_move \gets move$
            \EndIf
        \EndFor
        \State \Return $max\_eval, best\_move$
    \Else \Comment{$P=2$, Minimizing}
        \State $min\_eval \gets \infty$, $best\_move \gets \text{None}$
        \For{move in $EmptyCells$}
            \State $B_{new} \gets \text{ApplyMove}(B, move, P)$
            \State $evaluation, \_ \gets \text{MinimaxGetScore}(B_{new}, \text{Player } 1, d+1)$
            \If{$evaluation < min\_eval$}
                \State $min\_eval \gets evaluation$, $best\_move \gets move$
            \EndIf
        \EndFor
        \State \Return $min\_eval, best\_move$
    \EndIf
\EndFunction
\end{algorithmic}
\end{algorithm}

\begin{algorithm}
\caption{Strategic Threat Detection and Fork Analysis}
\label{alg:threat_detection}
\begin{algorithmic}[1]
\Require Board state $B$, player $P$
\Ensure Count of immediate threats for player $P$
\Function{CountLineThreats}{$B, P$}
    \State $threats \gets 0$
    \State $\mathcal{L} \gets \text{All 8 winning lines of the board}$
    \For{line in $\mathcal{L}$}
        \State $pieces \gets \text{GetPiecesOnLine}(B, \text{line})$
        \If{$\text{count}(pieces, P) = 2$ and $\text{count}(pieces, \text{empty}) = 1$}
            \State $threats \gets threats + 1$
        \EndIf
    \EndFor
    \State \Return $threats$
\EndFunction
\end{algorithmic}
\vspace{1em}
\begin{algorithmic}[1]
\Require Board state $B$, player $P$
\Ensure Boolean indicating if a fork opportunity exists for player $P$
\Function{HasFork}{$B, P$}
    \State $OpenSquares \gets \text{FindEmptyCells}(B)$
    \For{square in $OpenSquares$}
        \State $B_{temp} \gets \text{ApplyMove}(B, \text{square}, P)$
        \If{$\text{CountLineThreats}(B_{temp}, P) \ge 2$}
            \State \Return \textbf{true}
        \EndIf
    \EndFor
    \State \Return \textbf{false}
\EndFunction
\end{algorithmic}
\end{algorithm}

\begin{algorithm}
\caption{Board Canonical Form Generation}
\label{alg:canonical_form}
\begin{algorithmic}[1]
\Require A $3 \times 3$ board matrix $B$
\Ensure The lexicographically smallest (canonical) representation of the board
\Function{GetCanonicalForm}{$B$}
    \State $Symmetries \gets []$
    \State $B_{current} \gets B$
    \For{$i=1$ to $4$}
        \State Append $\text{Flatten}(B_{current})$ to $Symmetries$
        \State $B_{flipped} \gets \text{FlipLeftRight}(B_{current})$
        \State Append $\text{Flatten}(B_{flipped})$ to $Symmetries$
        \State $B_{current} \gets \text{Rotate90Degrees}(B_{current})$
    \EndFor
    \State \Return $\min(Symmetries)$
\EndFunction
\end{algorithmic}
\end{algorithm}

\clearpage
\section{Quantitative analysis}
\label{sec:appendix-quant}
This section provides the complete set of quantitative results for all settings. For both random and symmetry datasets, we conducted evaluations for both text based natural language board representations as well as ascii based board representations. These were done to evaluate both legal move and best move objectives across all trained model checkpoints. Model's robustness to the prompt variations was tested by random symbol substitution, choosing random alphanumeric characters to replace the player symbols X and O for the same board.

\subsection{Additional behavioural results}
\label{sec:appendix-additional-behaviour}
\paragraph{Scaling helps within limits.}
% Model capacity matters: Qwen2.5 1.5B reliably outperforms Llama3.2 1B and Qwen2.5 0.5B on both objectives at their best checkpoints (Fig.~\ref{fig:quant-overall}). The weaker models hover near smart baselines on best-move prediction, consistent with evidence that small LMs struggle to form robust task circuits without targeted training signals \citep{hillier2024super,eldan2023tinystories,wang2026tina}. Still, even the strongest model plateaus well below perfect best-move play, foreshadowing the representational limits we uncover mechanistically. 
% Interestingly, scale is not the only factor in play, as seen by the relatively better performance of Qwen2.5 0.5B model compared to the Llama3.2 1B, indicating pre-training may play a factor in downstream performance on reasoning tasks \citep{wang2025larger}.
\textbf{Qwen2.5 1.5B achieves the strongest results}, surpassing Llama3.2 1B and Qwen2.5 0.5B on both objectives at their best checkpoints (Fig.~\ref{fig:quant-overall}). 
The weaker models hover near smart baselines on best-move prediction, consistent with observations that small LMs often require targeted signals to form task circuits \citep{hillier2024super,eldan2023tinystories,wang2026tina}. 
Even the strongest model plateaus short of perfect best-move play (legal: up to $100\%$, best move: $71\%$ under symmetry split and $74\%$ under random split), foreshadowing the representational limits seen mechanistically. 
Differences between the $0.5$B Qwen and $1$B Llama also suggest pretraining quality matters for downstream “reasoning’’ \citep{wang2025larger}.

\paragraph{Objective difficulty: accuracy on legal $\gg$ best.}
Legal-move prediction is substantially easier than best-move selection (Fig.~\ref{fig:quant-overall}), matching reports that models master surface constraints before long-horizon structure \citep{topsakal2024evaluating,chen2024llmarena,zhang2024llm}. 
Outcome-aware analysis (full reports provided in appendix Fig.~\ref{fig:appendix-heatmaps}) show peaks on tactically imminent positions (large $|$score$|$) and dips on low-magnitude, ambiguous states where lookahead or symmetry reasoning should matter \citep{spies2024transformers,dao2025alphamaze}.

%%%%%%%%%%%%%%%%%%%%%%%%%%%%%%%%%%%%%%%%%%%%%%%%%%%%%%
% 1. Progression
%%%%%%%%%%%%%%%%%%%%%%%%%%%%%%%%%%%%%%%%%%%%%%%%%%%%%%
\begin{figure*}[ht]
    \centering
    % ---- Best Move ----
    \begin{subfigure}[t]{0.45\textwidth}
        \includegraphics[width=\linewidth]{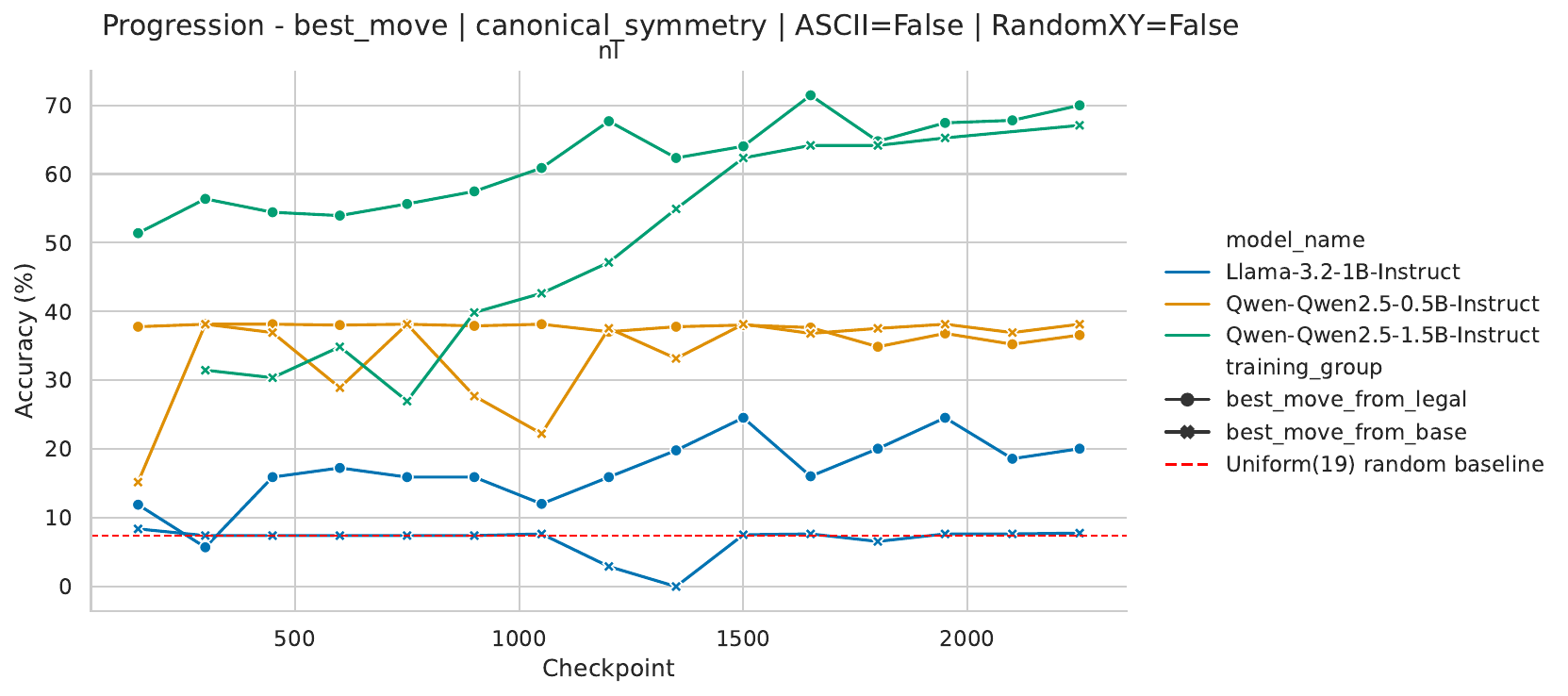}
        \caption{Progression, best move, Natural language board representation, random-False}
    \end{subfigure}%
    \hfill
    \begin{subfigure}[t]{0.45\textwidth}
        \includegraphics[width=\linewidth]{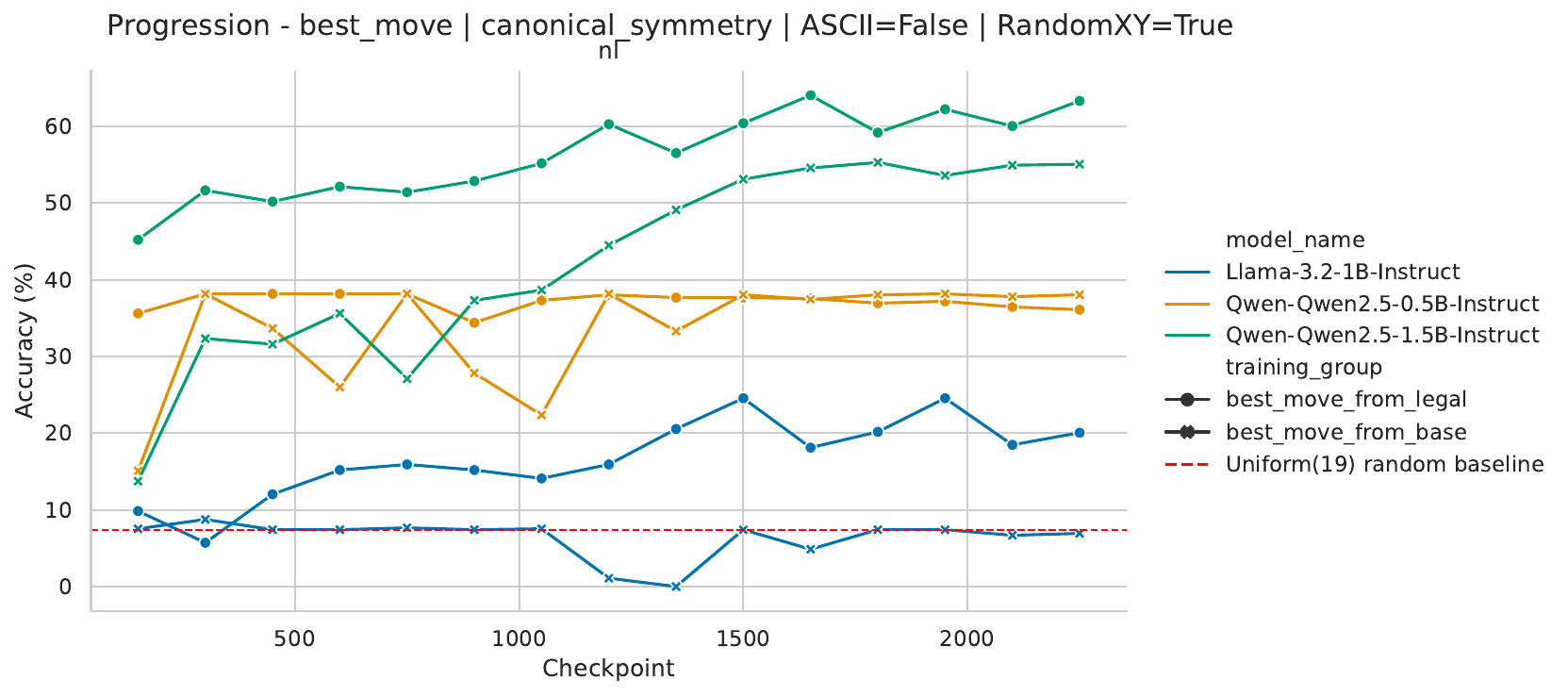}
        \caption{Progression, best move, Natural language board representation, random-True}
    \end{subfigure}

    \begin{subfigure}[t]{0.45\textwidth}
        \includegraphics[width=\linewidth]{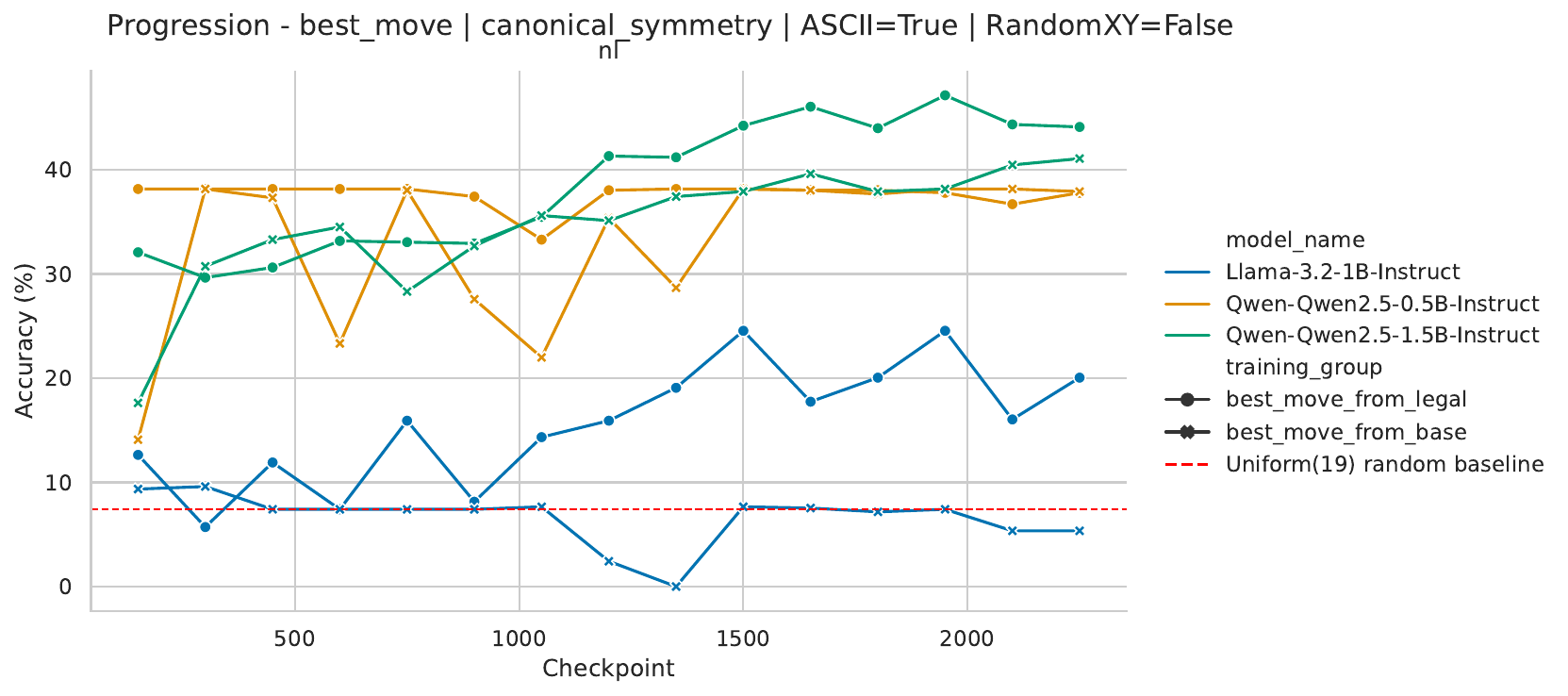}
        \caption{Progression, best move, ASCII board representation, random-False}
    \end{subfigure}%
    \hfill
    \begin{subfigure}[t]{0.45\textwidth}
        \includegraphics[width=\linewidth]{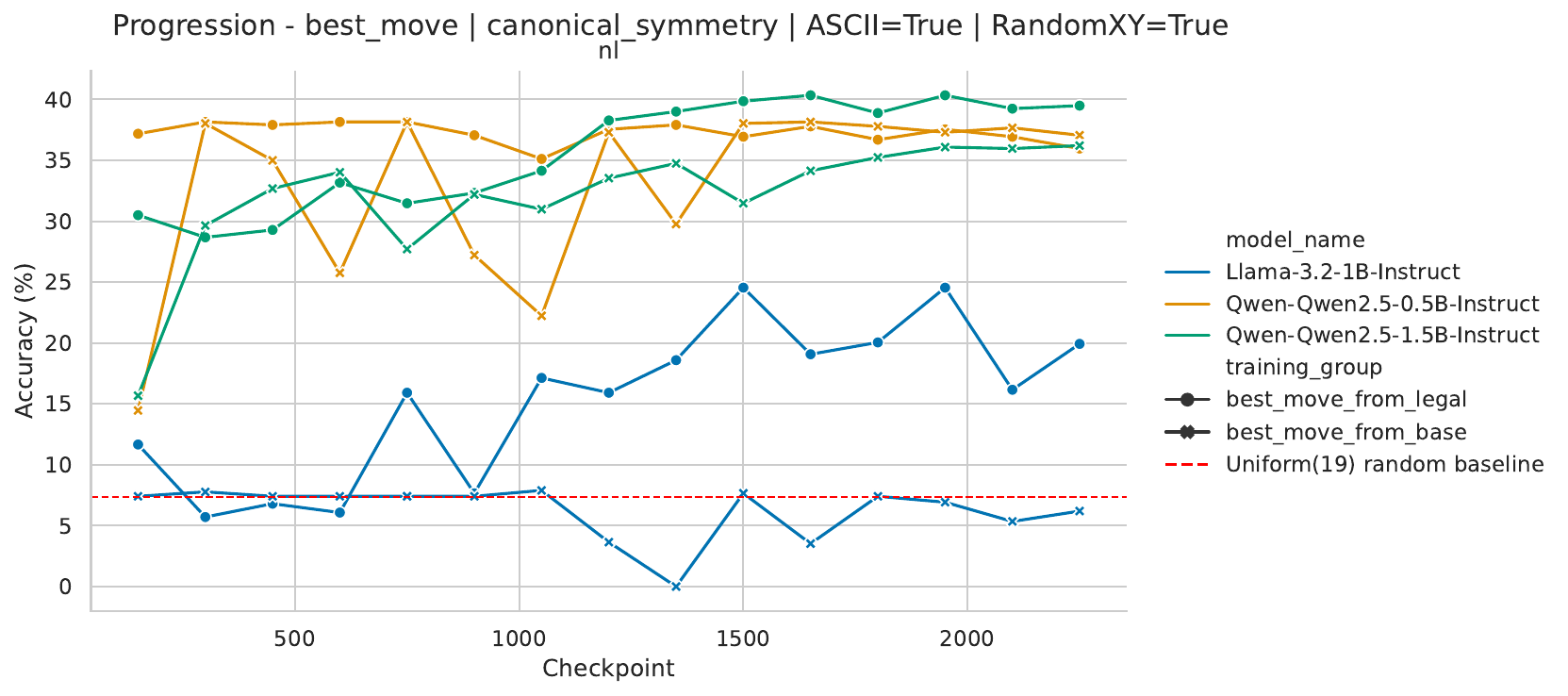}
        \caption{Progression, best move, ASCII board representation, random-True}
    \end{subfigure}

    % ---- Legal Move ----
    \begin{subfigure}[t]{0.45\textwidth}
        \includegraphics[width=\linewidth]{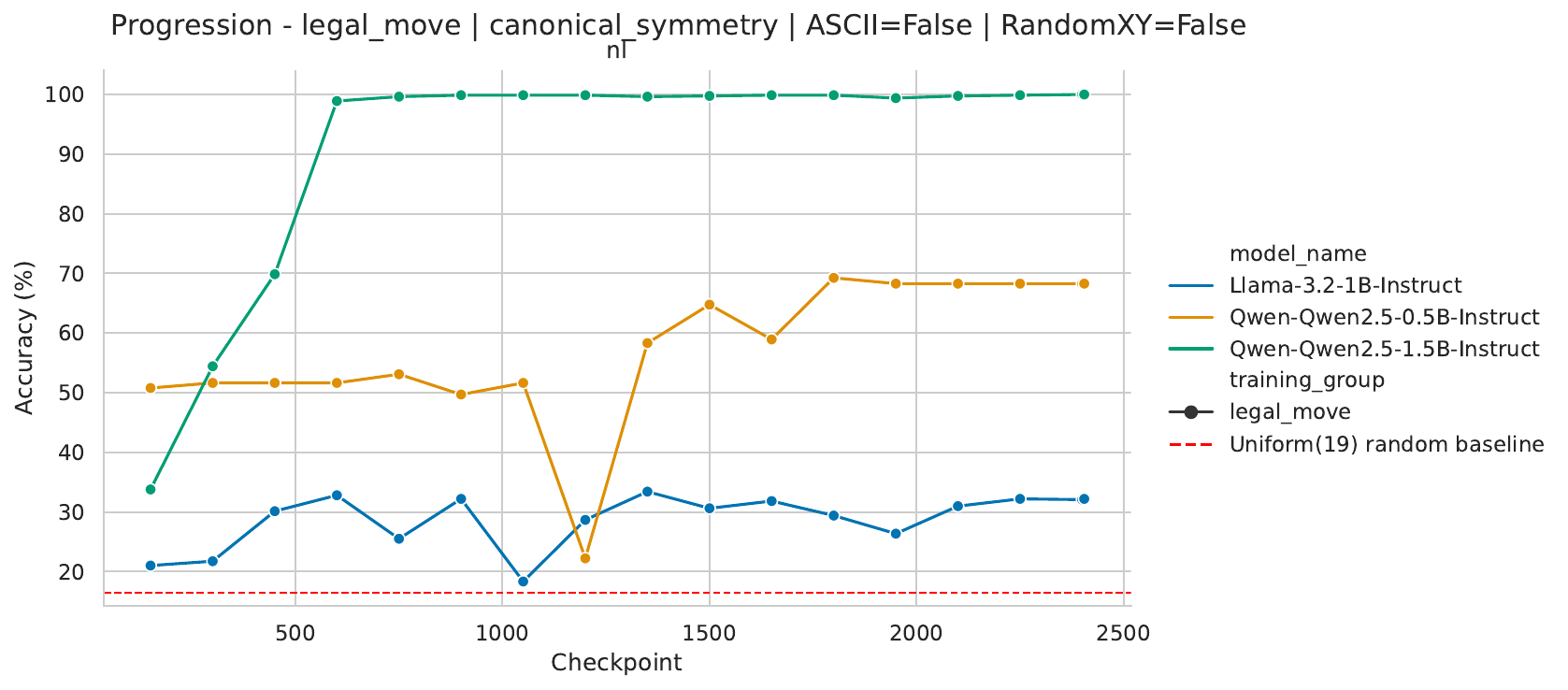}
        \caption{Progression, legal move, Natural language board representation, random-False}
    \end{subfigure}%
    \hfill
    \begin{subfigure}[t]{0.45\textwidth}
        \includegraphics[width=\linewidth]{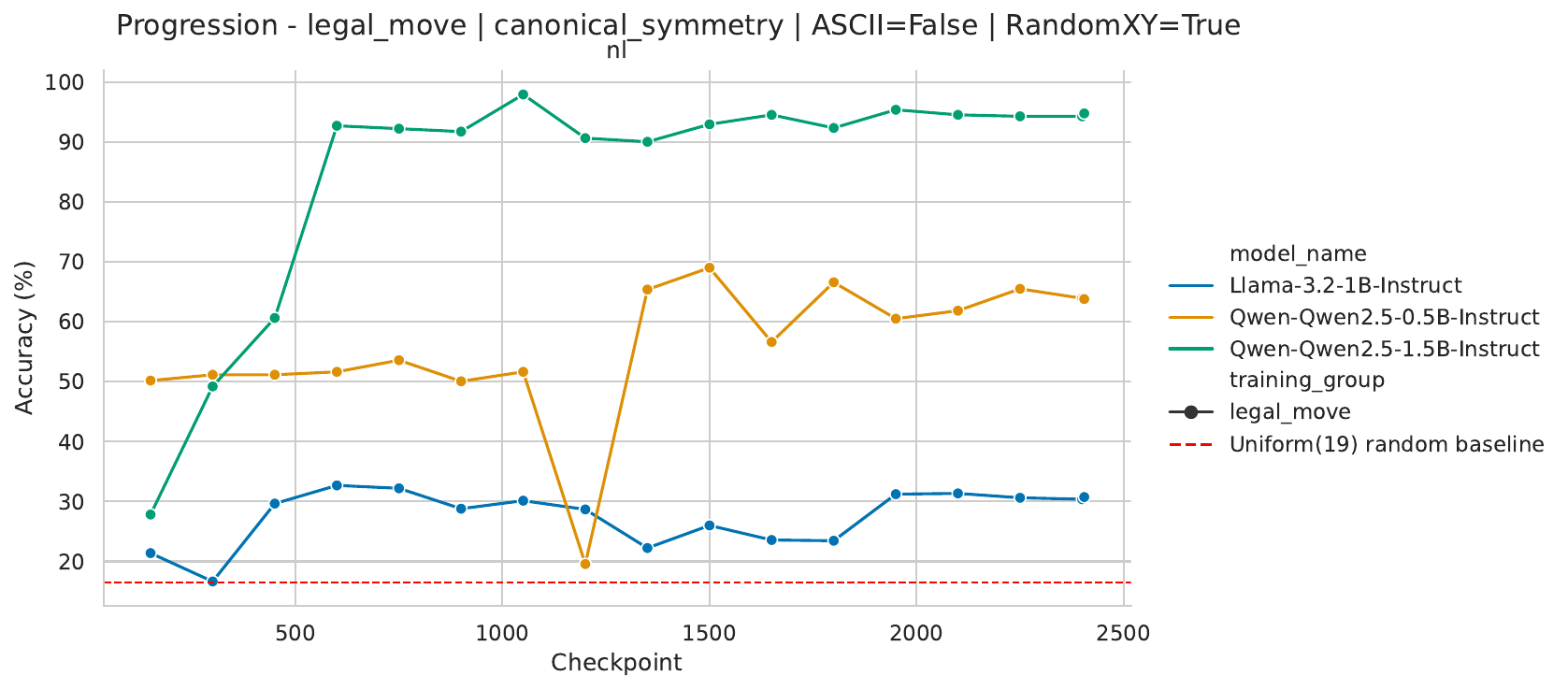}
        \caption{Progression, legal move, Natural language board representation, random-True}
    \end{subfigure}

    \begin{subfigure}[t]{0.45\textwidth}
        \includegraphics[width=\linewidth]{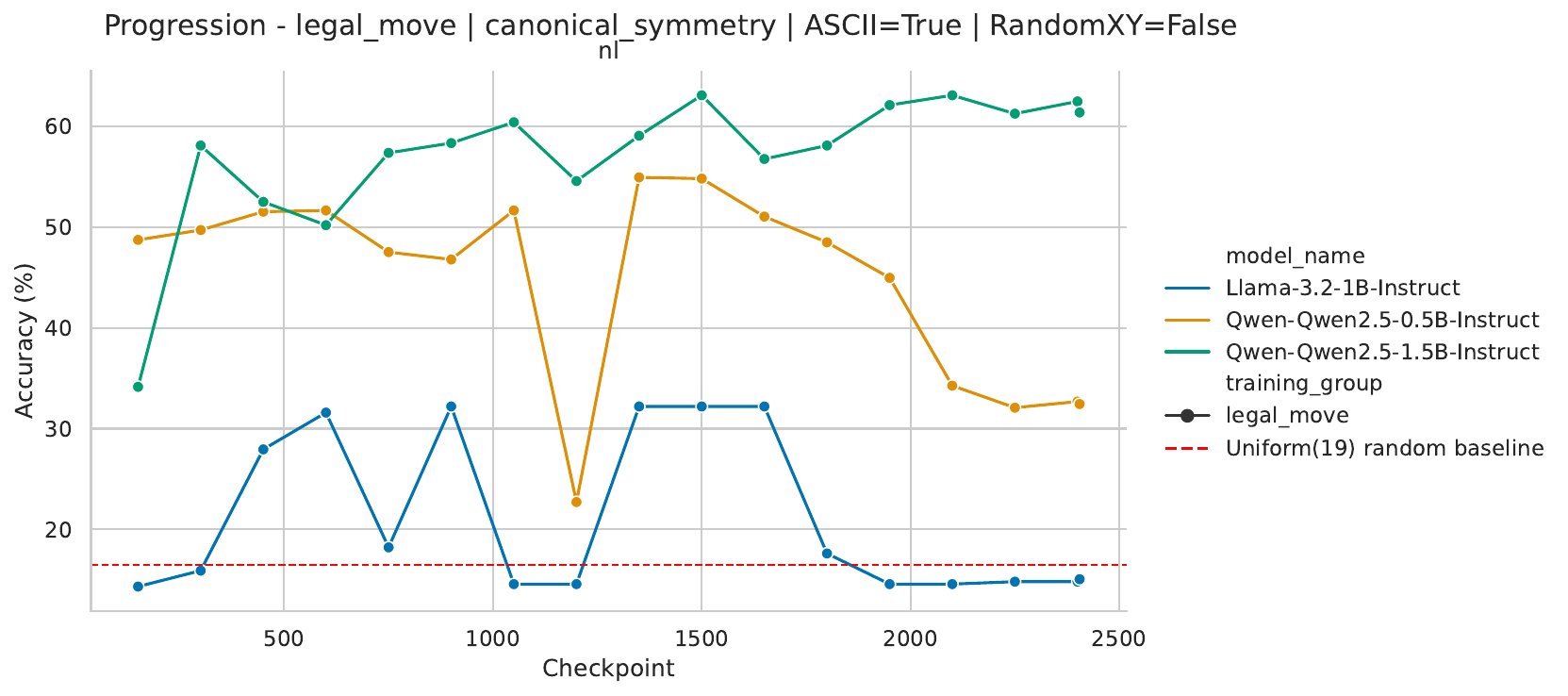}
        \caption{Progression, legal move, ASCII board representation, random-False}
    \end{subfigure}%
    \hfill
    \begin{subfigure}[t]{0.45\textwidth}
        \includegraphics[width=\linewidth]{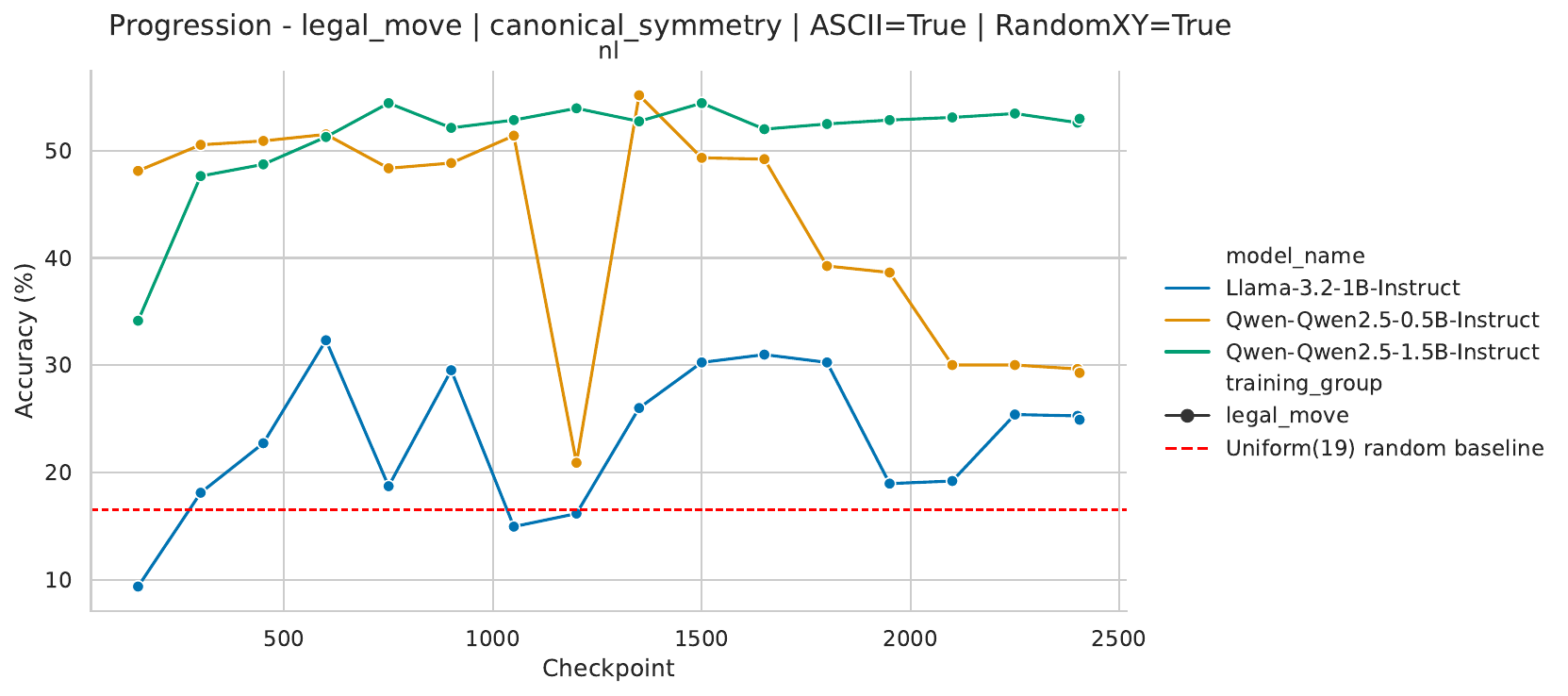}
        \caption{Progression, legal move, ASCII board representation, random-True}
    \end{subfigure}

    \caption{Progression analysis across board representations and randomization.}
    \label{fig:appendix-progression}
\end{figure*}

%%%%%%%%%%%%%%%%%%%%%%%%%%%%%%%%%%%%%%%%%%%%%%%%%%%%%%
% 2. Overall
%%%%%%%%%%%%%%%%%%%%%%%%%%%%%%%%%%%%%%%%%%%%%%%%%%%%%%
\begin{figure*}[ht]
    \centering
    % ---- Best Move ----
    \begin{subfigure}[t]{0.45\textwidth}
        \includegraphics[width=\linewidth]{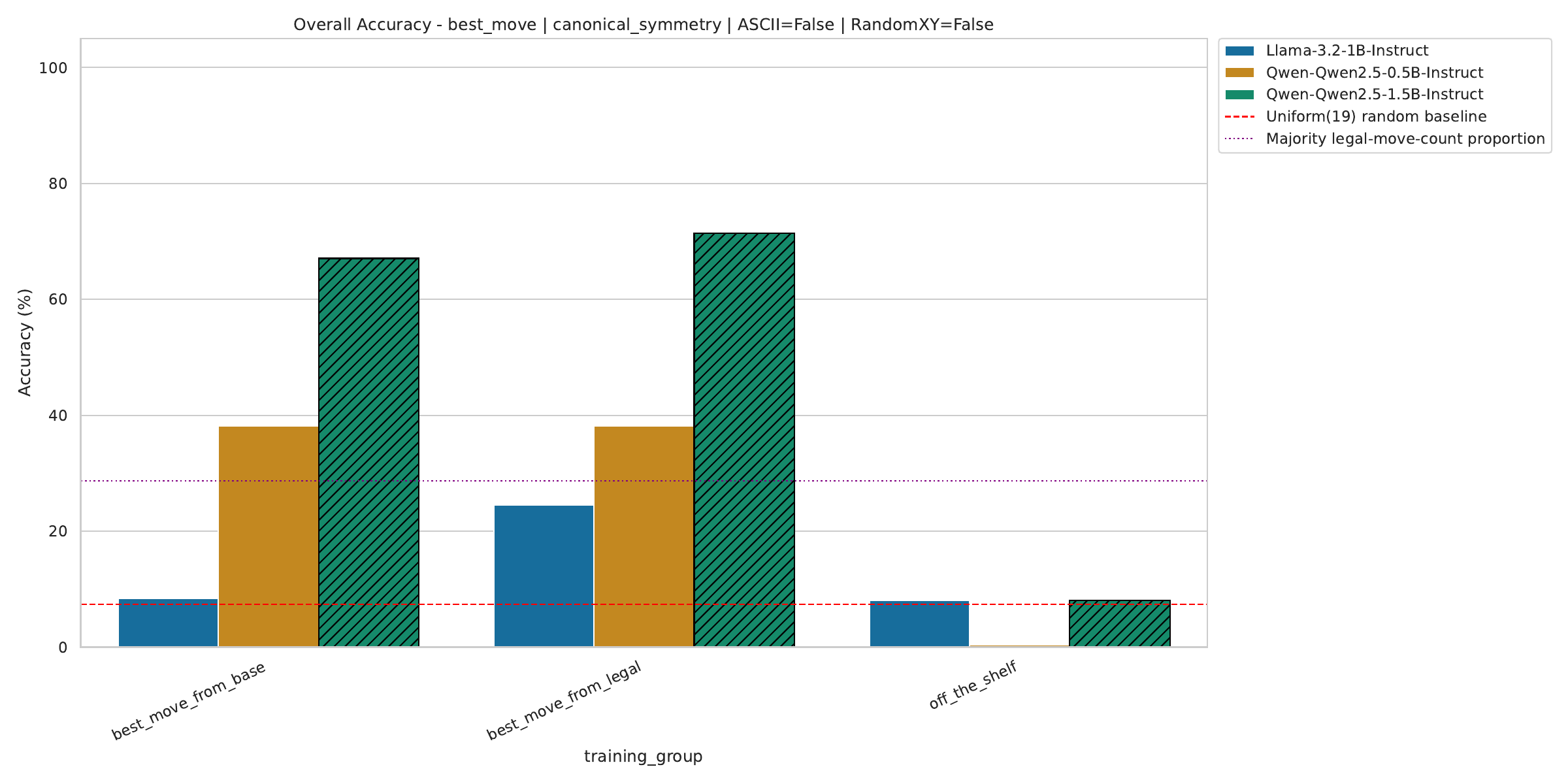}
        \caption{Overall, best move, Natural language board representation, random-False}
    \end{subfigure}%
    \hfill
    \begin{subfigure}[t]{0.45\textwidth}
        \includegraphics[width=\linewidth]{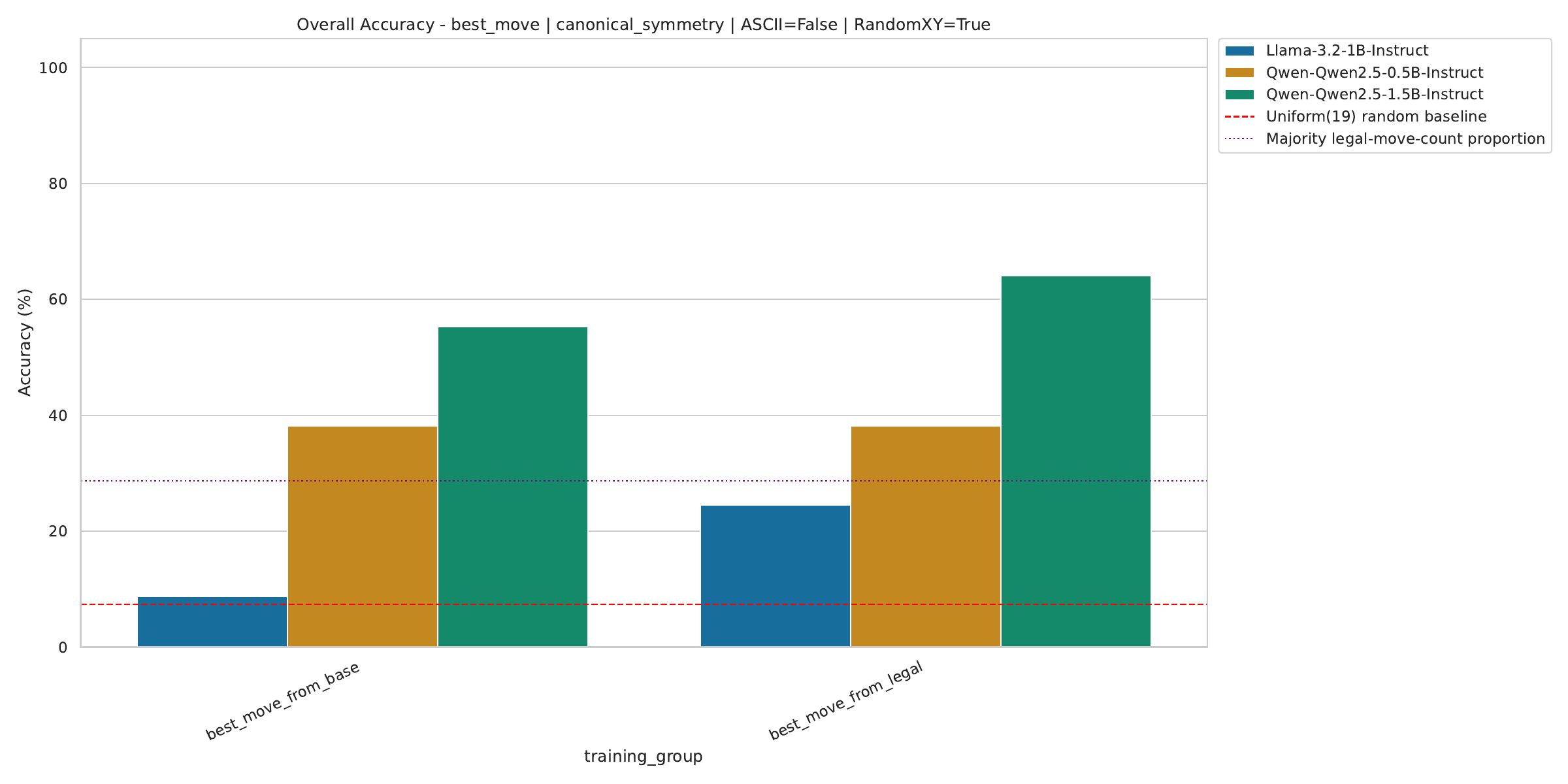}
        \caption{Overall, best move, Natural language board representation, random-True}
    \end{subfigure}

    \begin{subfigure}[t]{0.45\textwidth}
        \includegraphics[width=\linewidth]{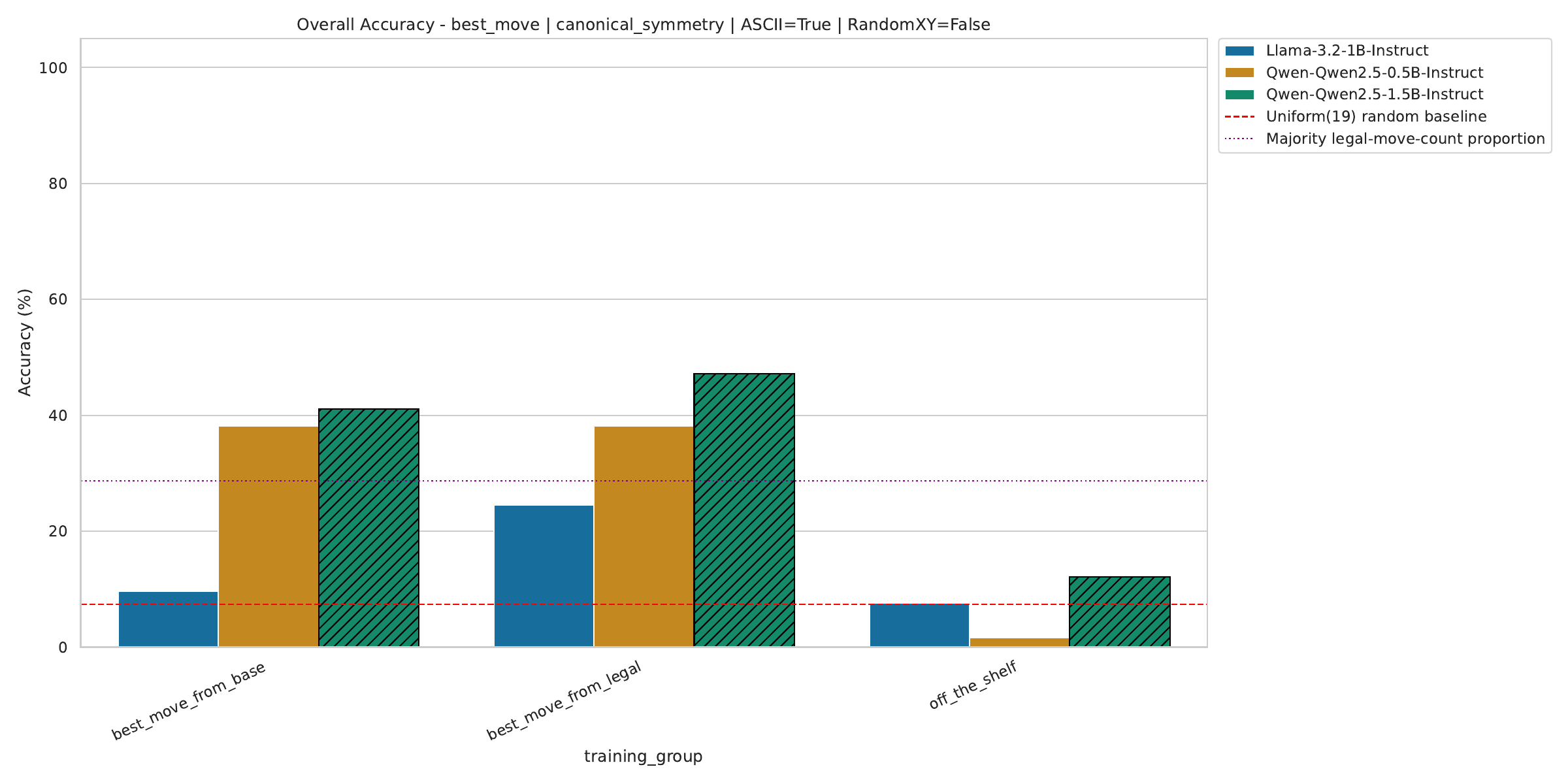}
        \caption{Overall, best move, ASCII board representation, random-False}
    \end{subfigure}%
    \hfill
    \begin{subfigure}[t]{0.45\textwidth}
        \includegraphics[width=\linewidth]{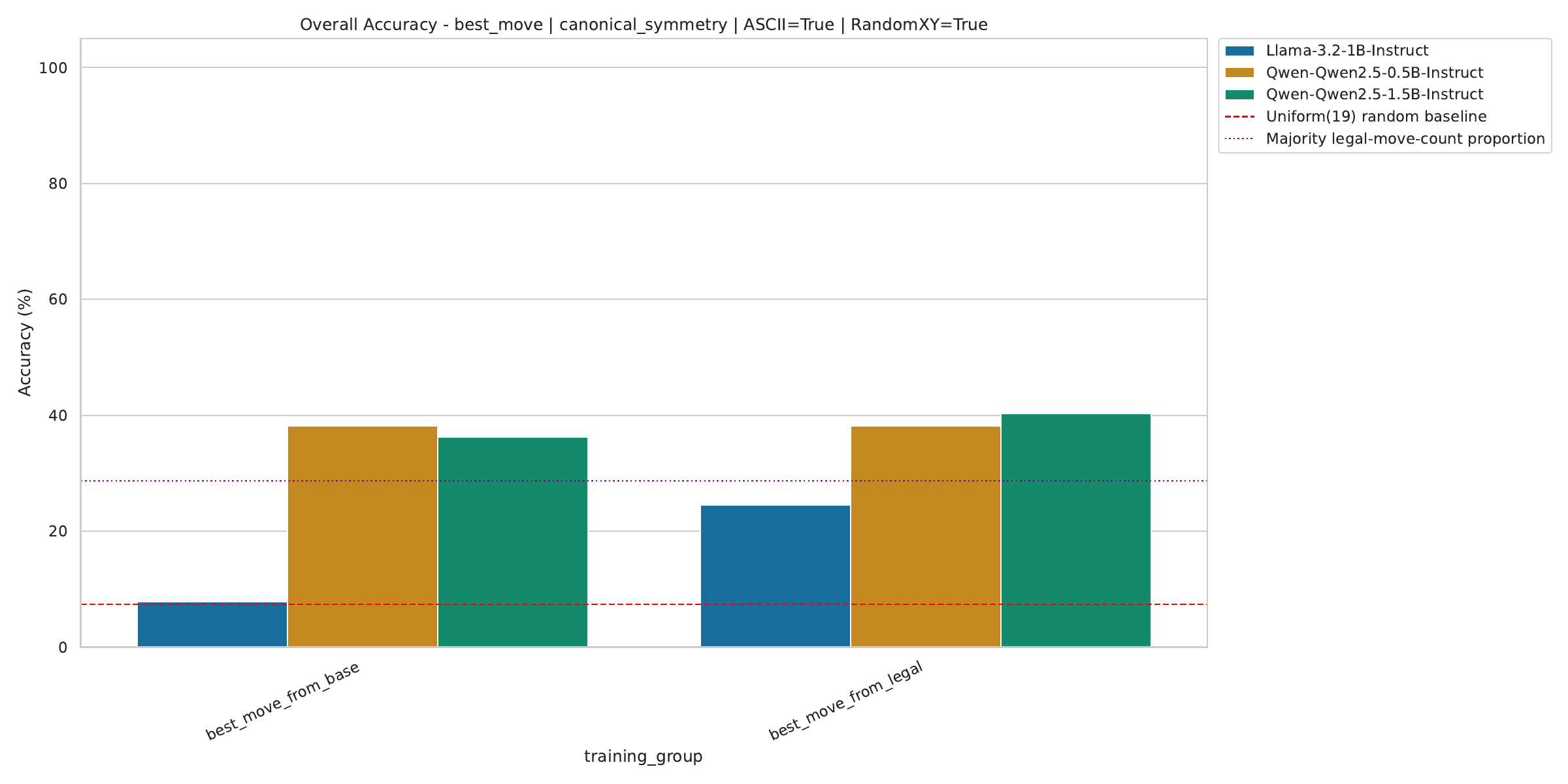}
        \caption{Overall, best move, ASCII board representation, random-True}
    \end{subfigure}

    % ---- Legal Move ----
    \begin{subfigure}[t]{0.45\textwidth}
        \includegraphics[width=\linewidth]{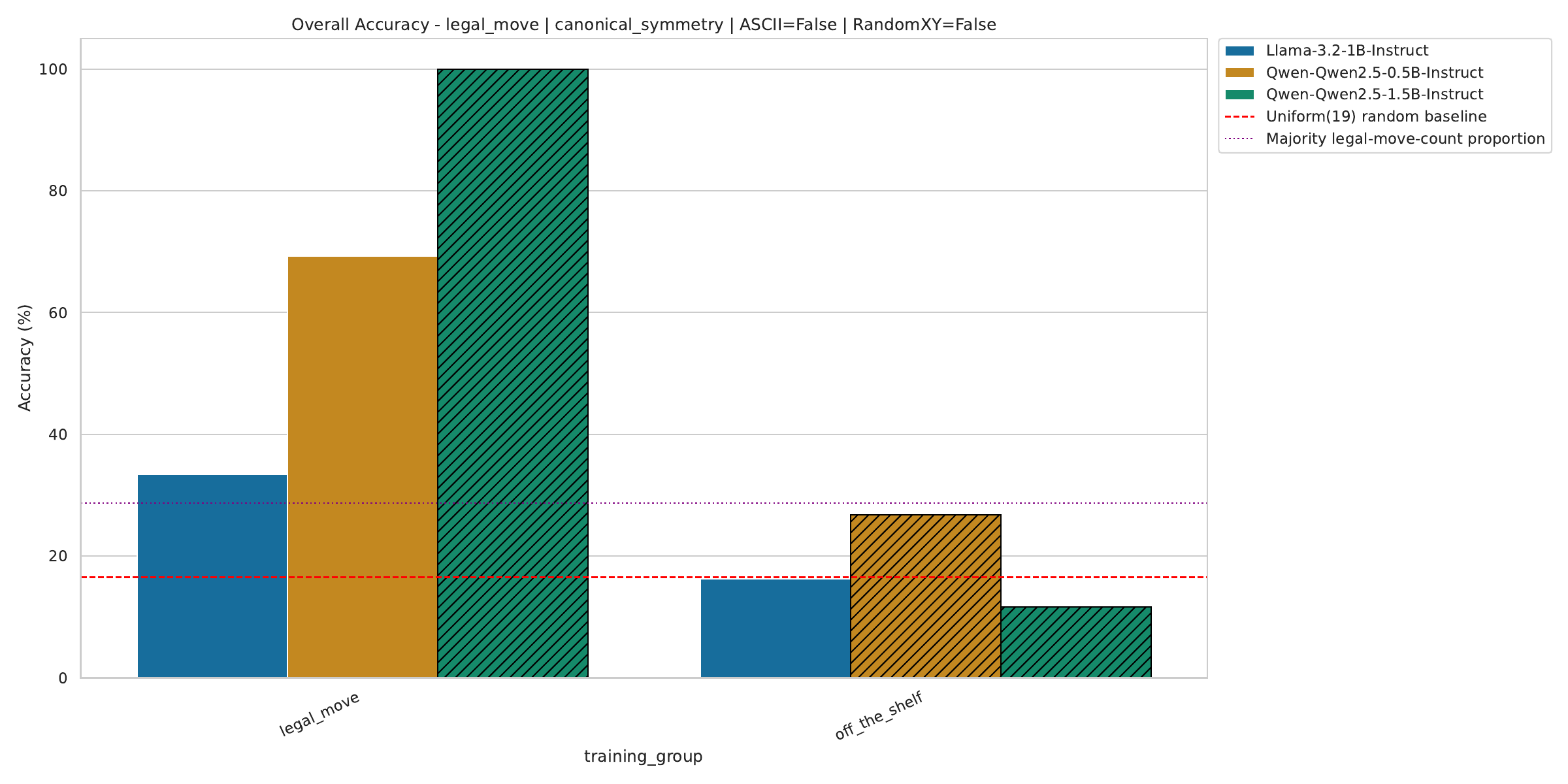}
        \caption{Overall, legal move, Natural language board representation, random-False}
    \end{subfigure}%
    \hfill
    \begin{subfigure}[t]{0.45\textwidth}
        \includegraphics[width=\linewidth]{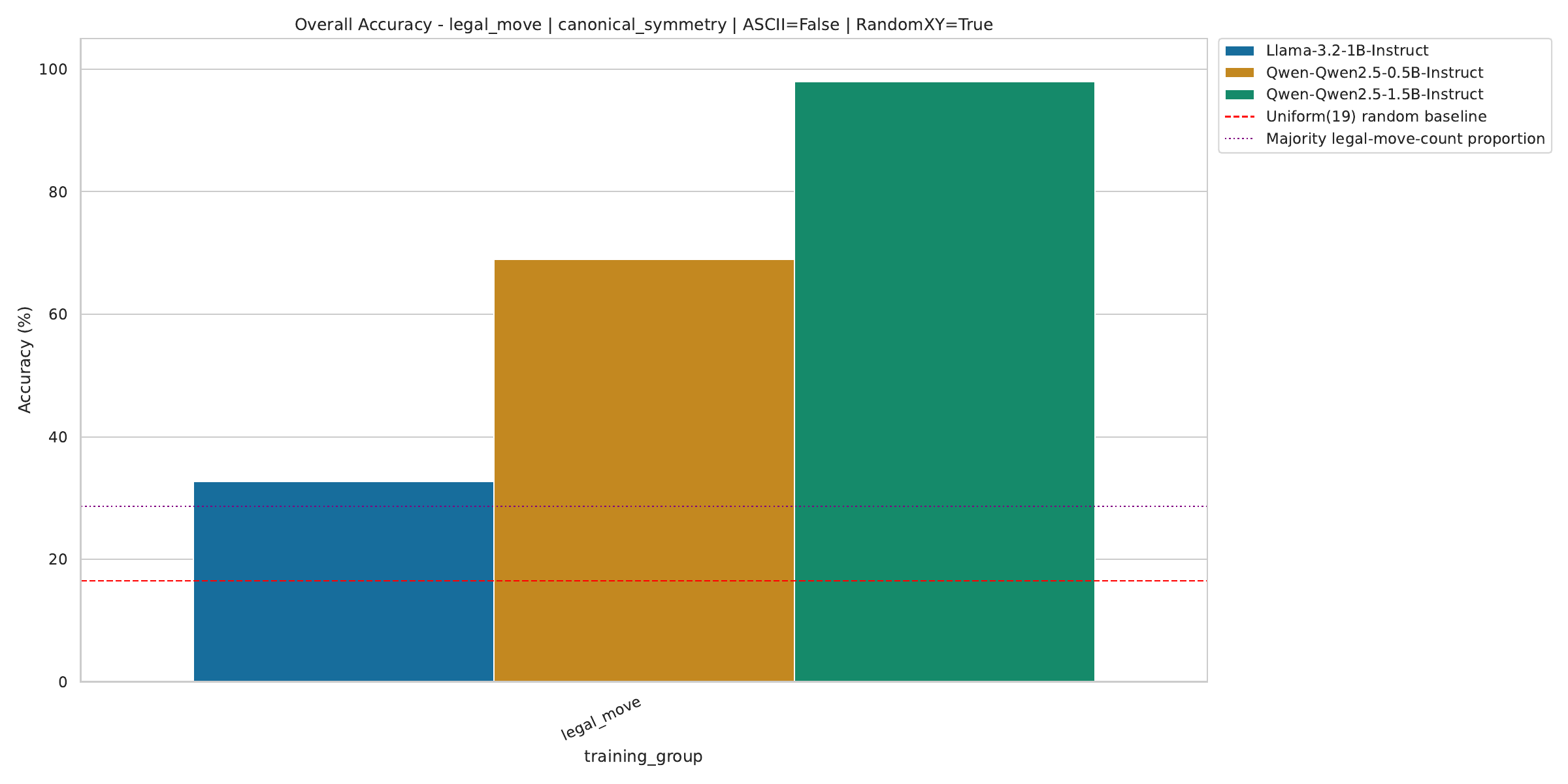}
        \caption{Overall, legal move, Natural language board representation, random-True}
    \end{subfigure}

    \begin{subfigure}[t]{0.45\textwidth}
        \includegraphics[width=\linewidth]{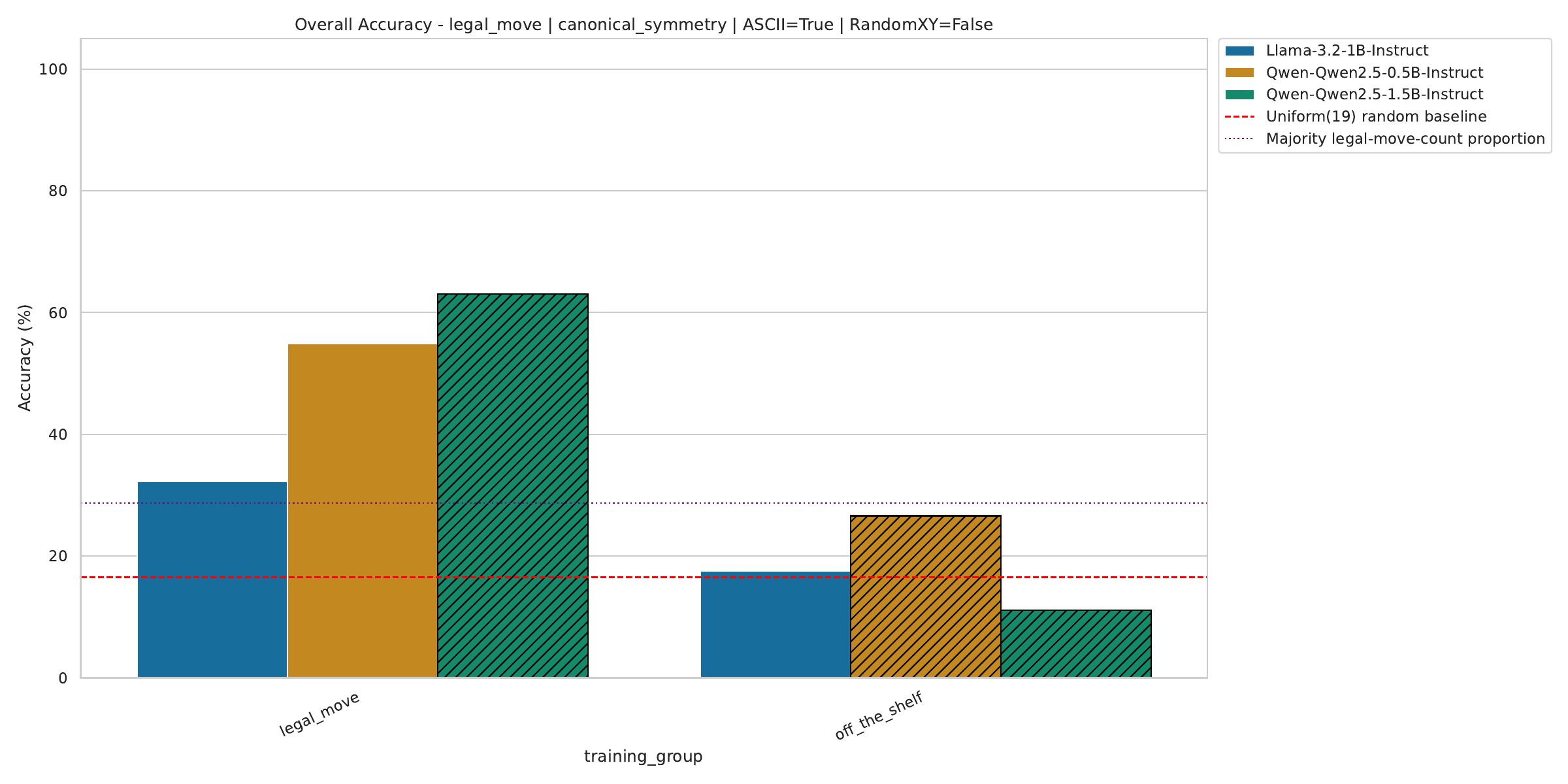}
        \caption{Overall, legal move, ASCII board representation, random-False}
    \end{subfigure}%
    \hfill
    \begin{subfigure}[t]{0.45\textwidth}
        \includegraphics[width=\linewidth]{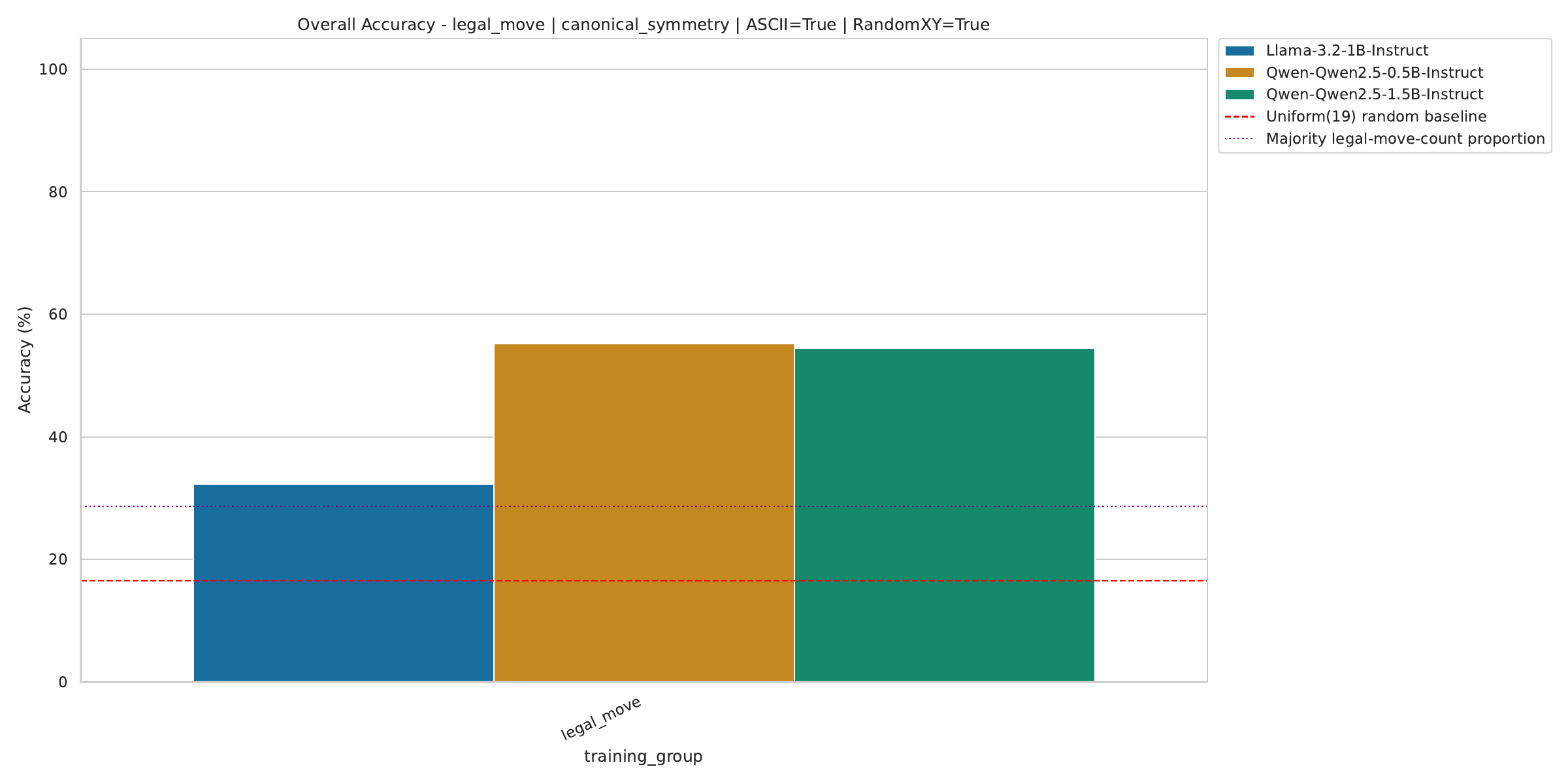}
        \caption{Overall, legal move, ASCII board representation, random-True}
    \end{subfigure}

    \caption{Overall performance across board representations and randomization.}
    \label{fig:appendix-overall}
\end{figure*}

%%%%%%%%%%%%%%%%%%%%%%%%%%%%%%%%%%%%%%%%%%%%%%%%%%%%%%
% 3. Heatmaps
%%%%%%%%%%%%%%%%%%%%%%%%%%%%%%%%%%%%%%%%%%%%%%%%%%%%%%
\begin{figure*}[ht]
    \centering
    % ---- Best Move ----
    \begin{subfigure}[t]{0.45\textwidth}
        \includegraphics[width=\linewidth]{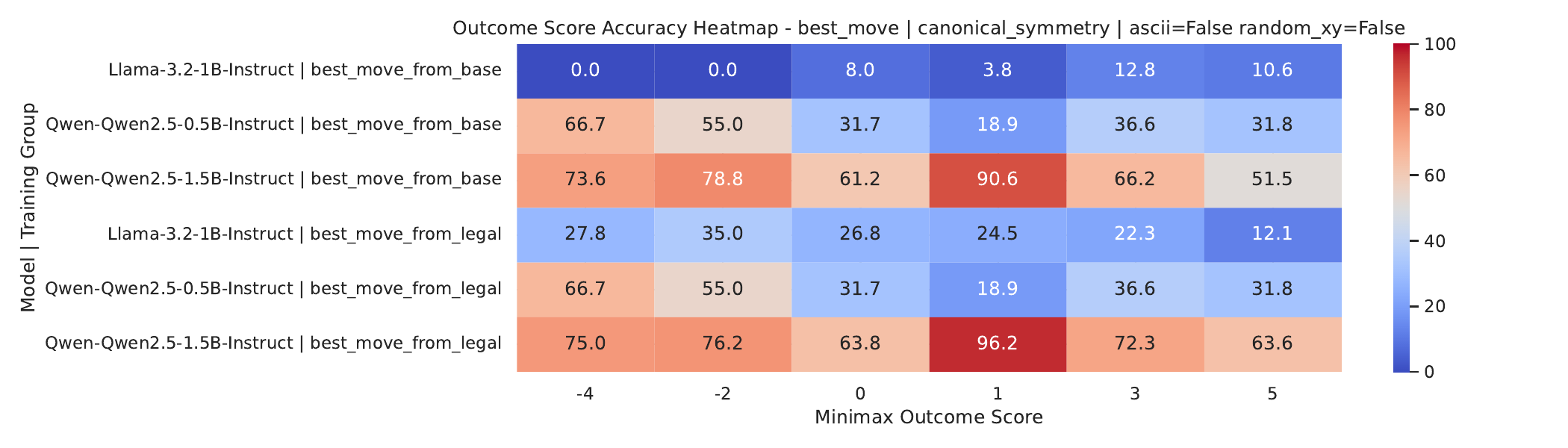}
        \caption{Heatmap, best move, Natural language board representation, random-False}
    \end{subfigure}%
    \hfill
    \begin{subfigure}[t]{0.45\textwidth}
        \includegraphics[width=\linewidth]{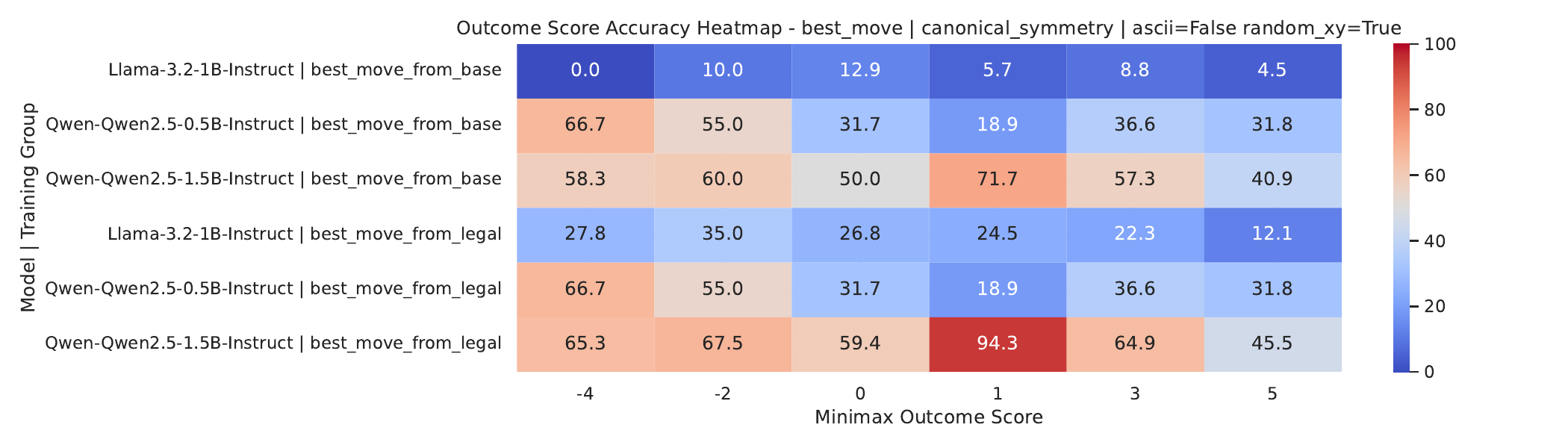}
        \caption{Heatmap, best move, Natural language board representation, random-True}
    \end{subfigure}

    \begin{subfigure}[t]{0.45\textwidth}
        \includegraphics[width=\linewidth]{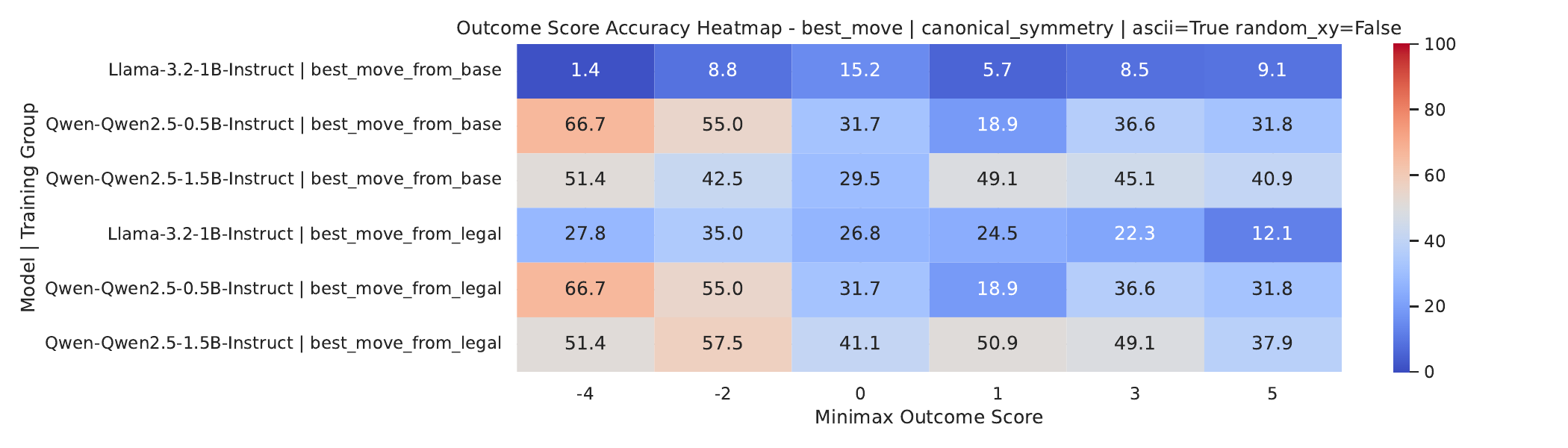}
        \caption{Heatmap, best move, ASCII board representation, random-False}
    \end{subfigure}%
    \hfill
    \begin{subfigure}[t]{0.45\textwidth}
        \includegraphics[width=\linewidth]{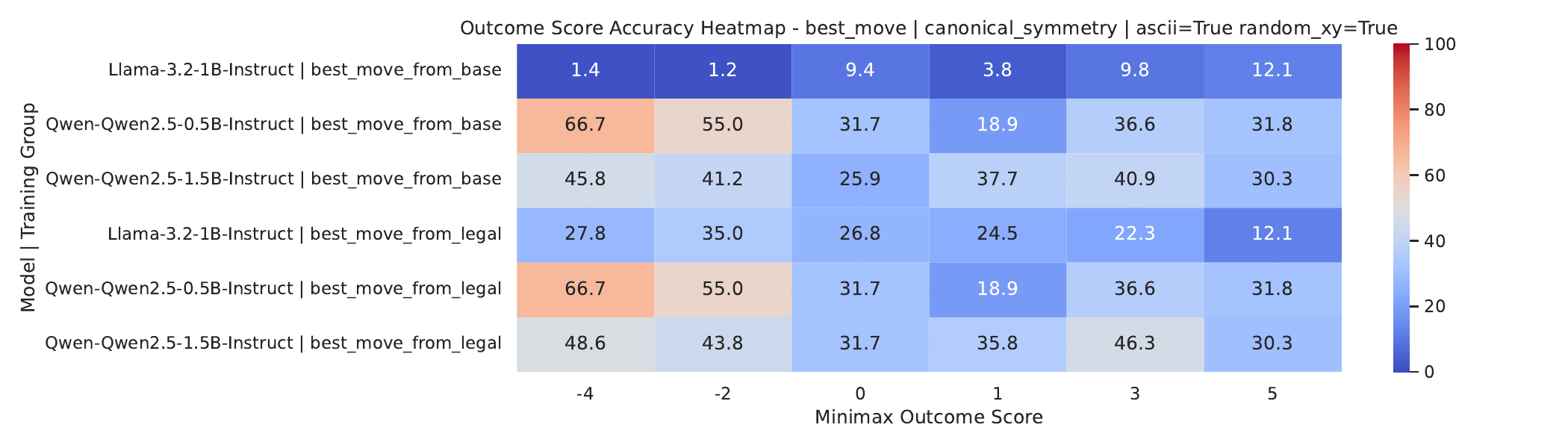}
        \caption{Heatmap, best move, ASCII board representation, random-True}
    \end{subfigure}

    % ---- Legal Move ----
    \begin{subfigure}[t]{0.45\textwidth}
        \includegraphics[width=\linewidth]{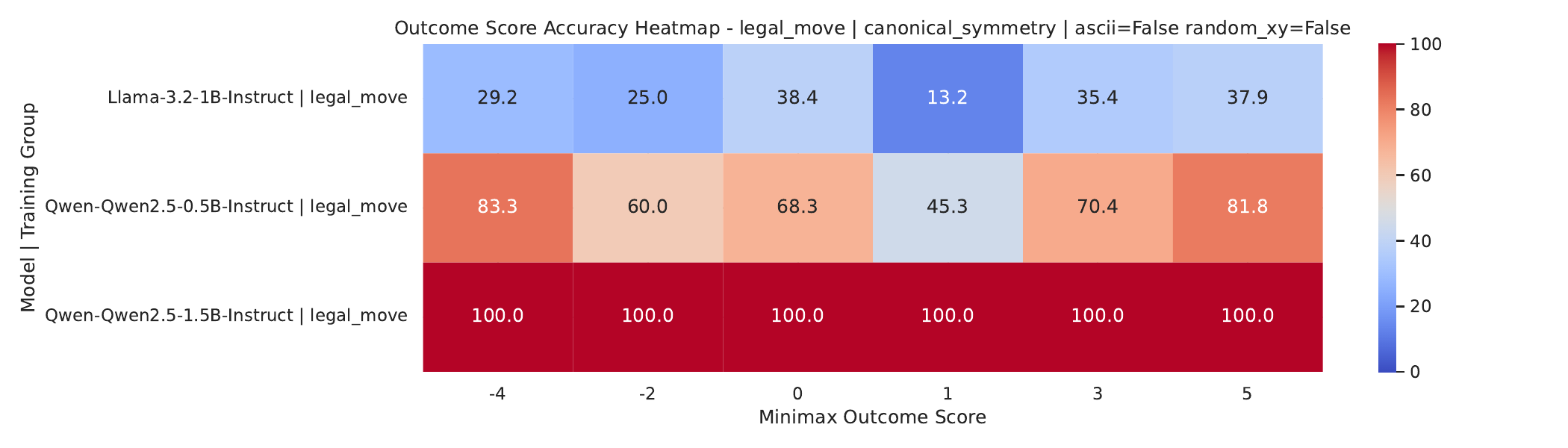}
        \caption{Heatmap, legal move, Natural language board representation, random-False}
    \end{subfigure}%
    \hfill
    \begin{subfigure}[t]{0.45\textwidth}
        \includegraphics[width=\linewidth]{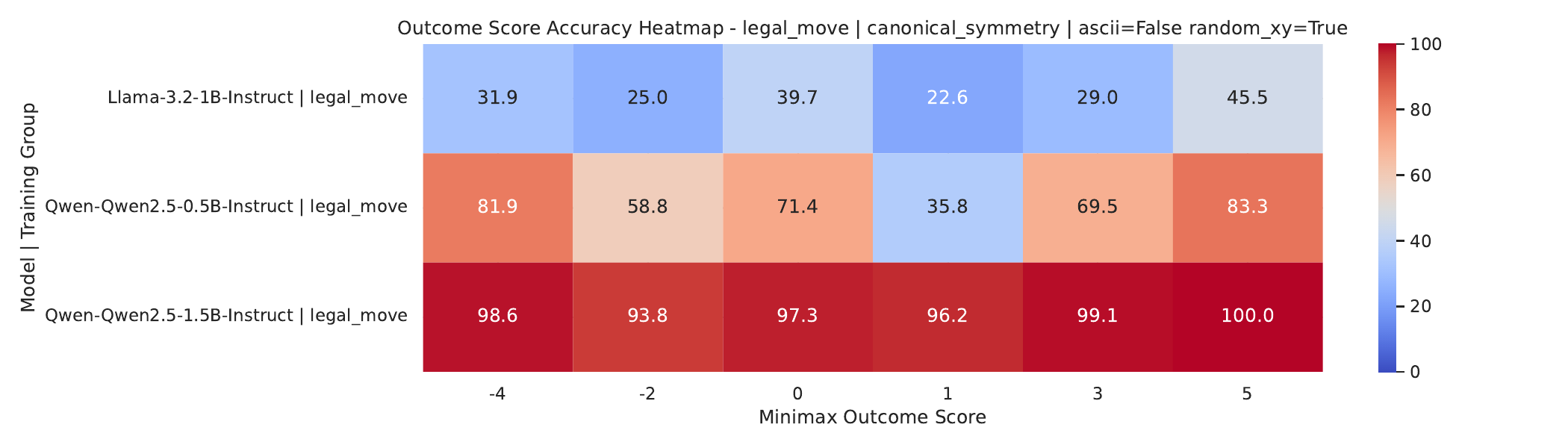}
        \caption{Heatmap, legal move, Natural language board representation, random-True}
    \end{subfigure}

    \begin{subfigure}[t]{0.45\textwidth}
        \includegraphics[width=\linewidth]{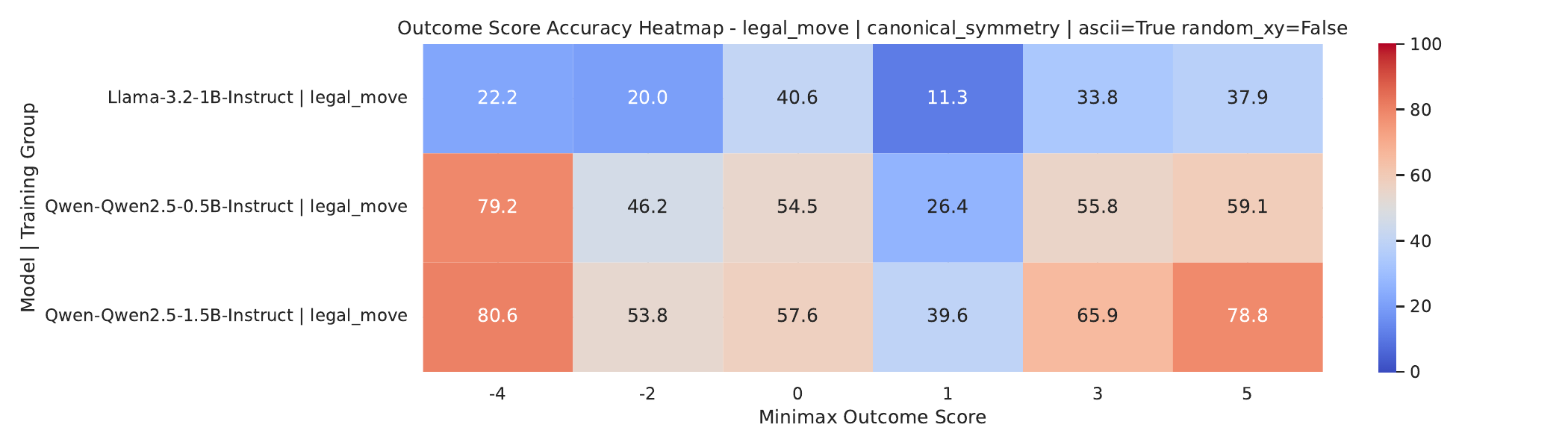}
        \caption{Heatmap, legal move, ASCII board representation, random-False}
    \end{subfigure}%
    \hfill
    \begin{subfigure}[t]{0.45\textwidth}
        \includegraphics[width=\linewidth]{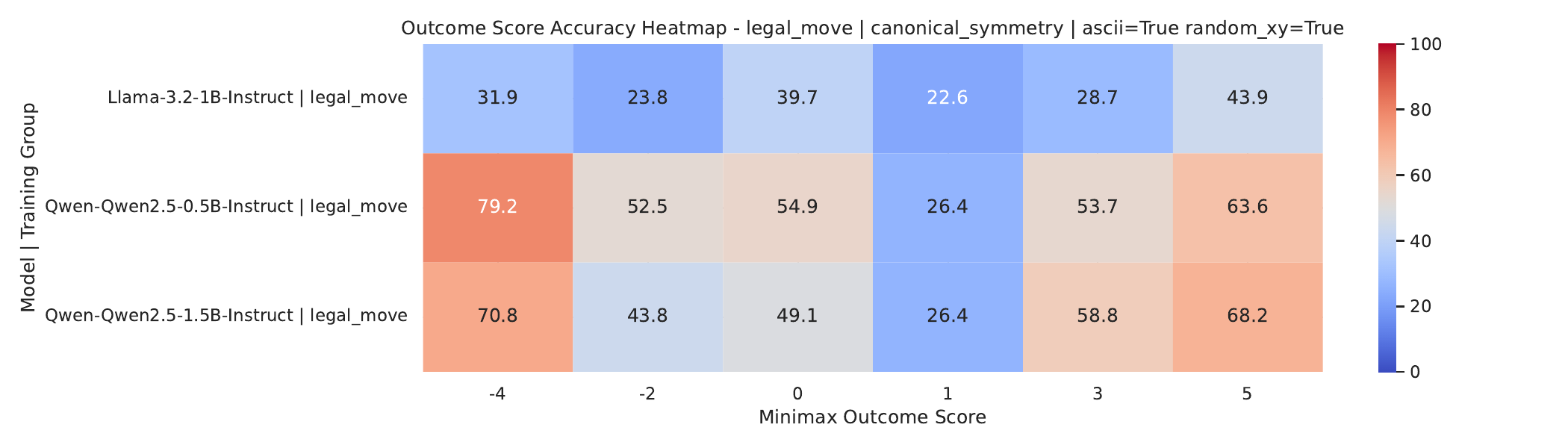}
        \caption{Heatmap, legal move, ASCII board representation, random-True}
    \end{subfigure}

    \caption{Outcome score heatmaps across board representations and randomization.}
    \label{fig:appendix-heatmaps}
\end{figure*}

%%%%%%%%%%%%%%%%%%%%%%%%%%%%%%%%%%%%%%%%%%%%%%%%%%%%%%
% 4. Open Spaces
%%%%%%%%%%%%%%%%%%%%%%%%%%%%%%%%%%%%%%%%%%%%%%%%%%%%%%
\begin{figure*}[ht]
    \centering
    % ---- Best Move ----
    \begin{subfigure}[t]{0.45\textwidth}
        \includegraphics[width=\linewidth]{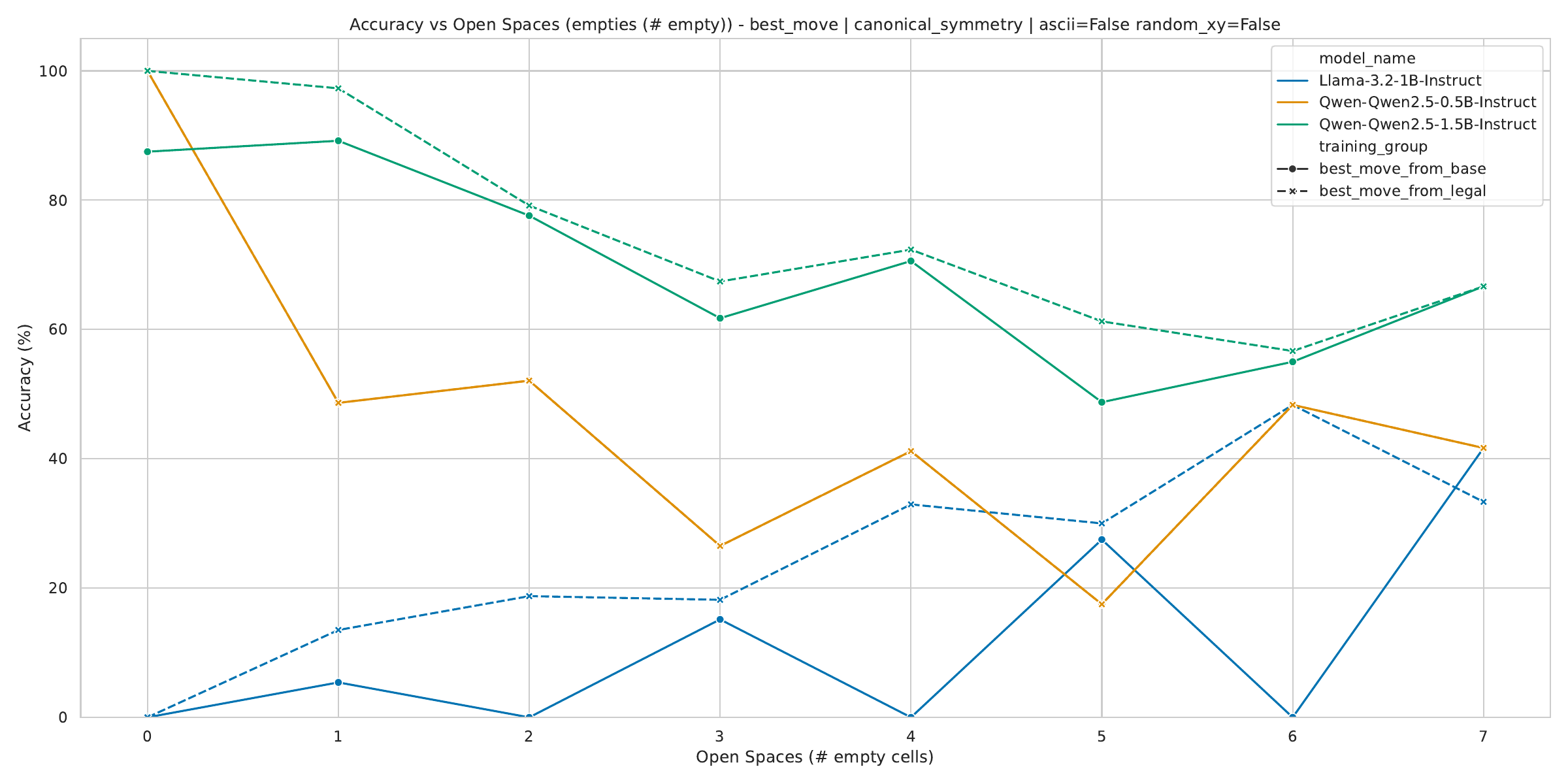}
        \caption{Open spaces, best move, Natural language board representation, random-False}
    \end{subfigure}%
    \hfill
    \begin{subfigure}[t]{0.45\textwidth}
        \includegraphics[width=\linewidth]{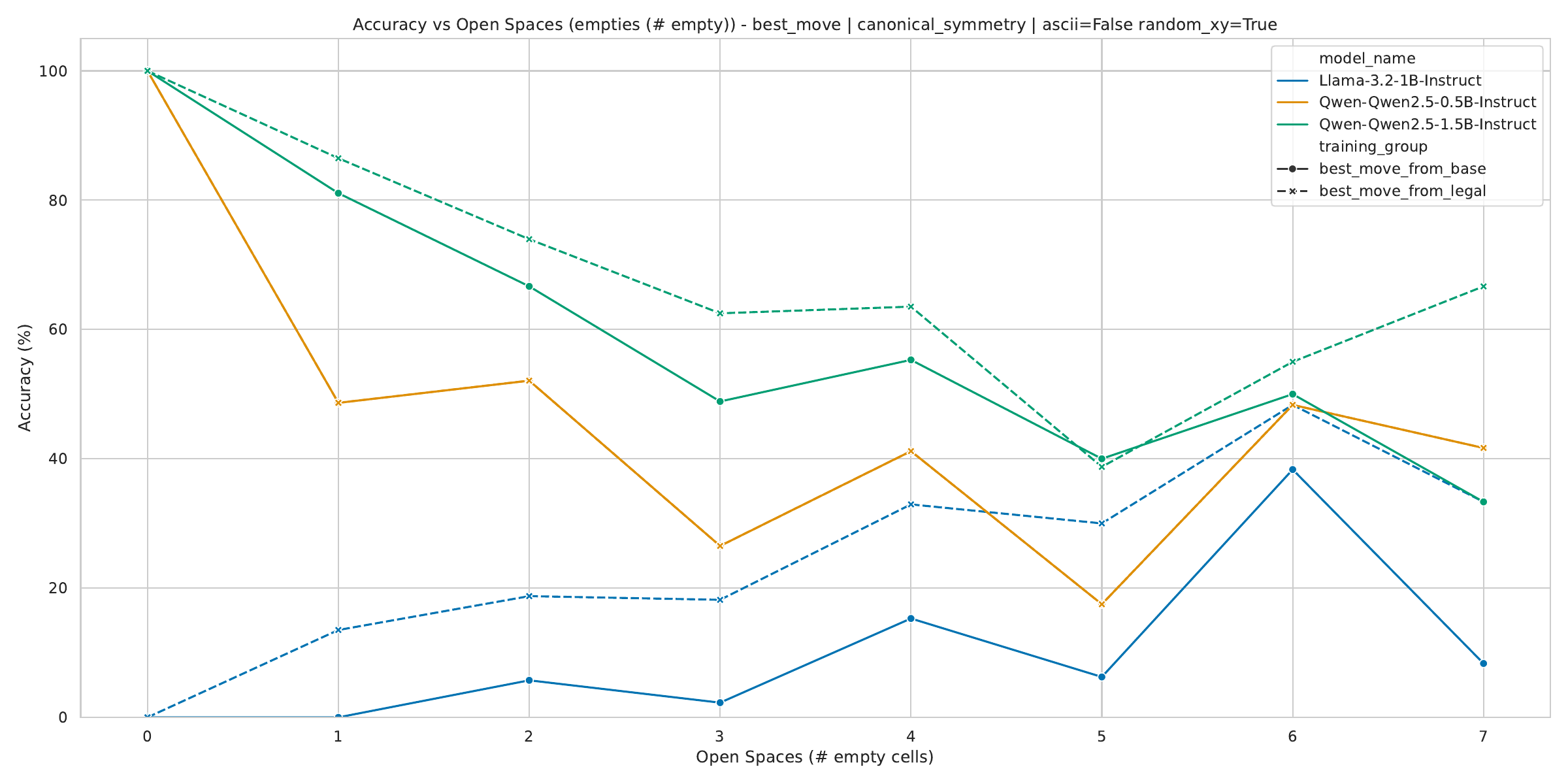}
        \caption{Open spaces, best move, Natural language board representation, random-True}
    \end{subfigure}

    \begin{subfigure}[t]{0.45\textwidth}
        \includegraphics[width=\linewidth]{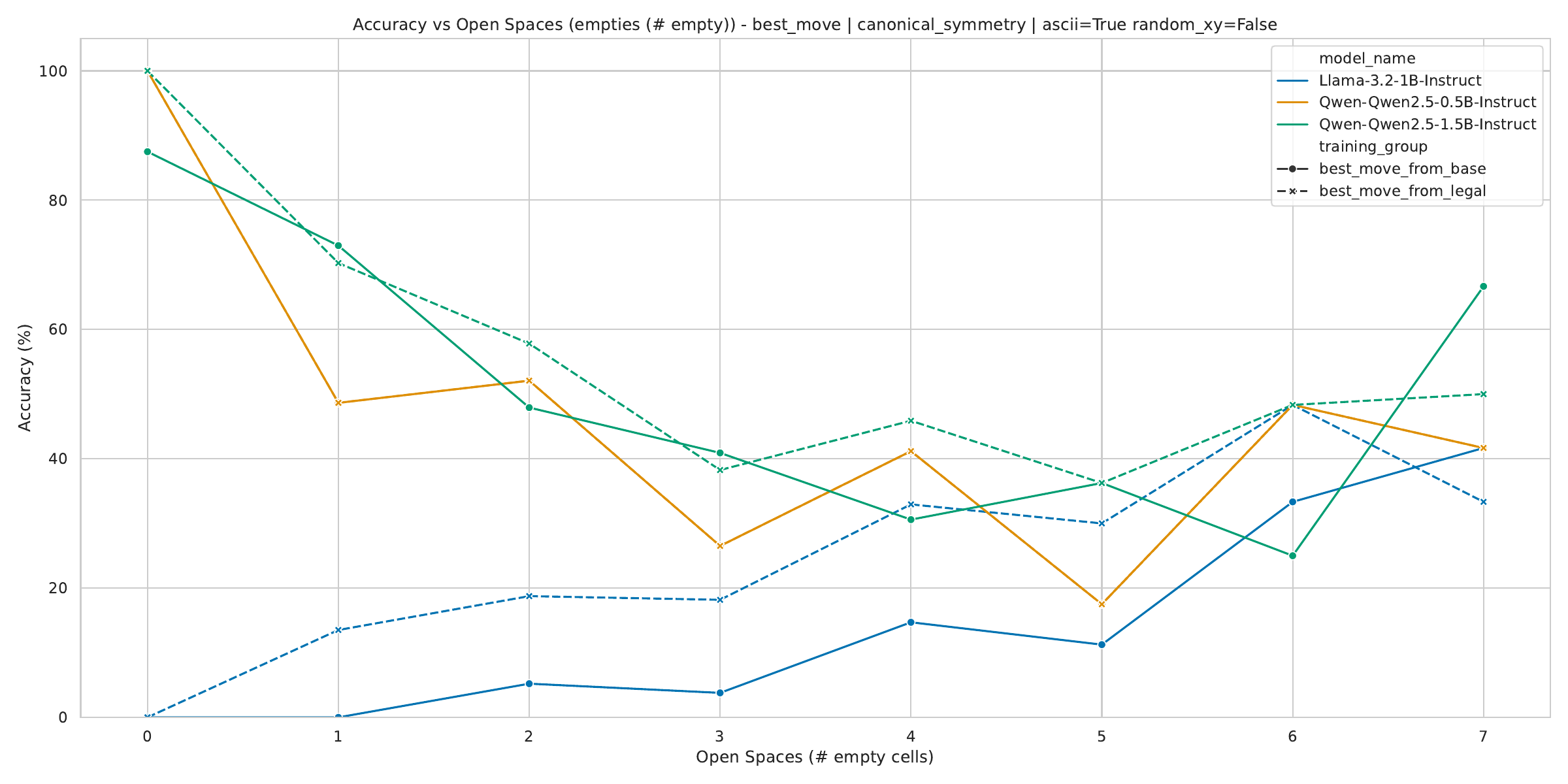}
        \caption{Open spaces, best move, ASCII board representation, random-False}
    \end{subfigure}%
    \hfill
    \begin{subfigure}[t]{0.45\textwidth}
        \includegraphics[width=\linewidth]{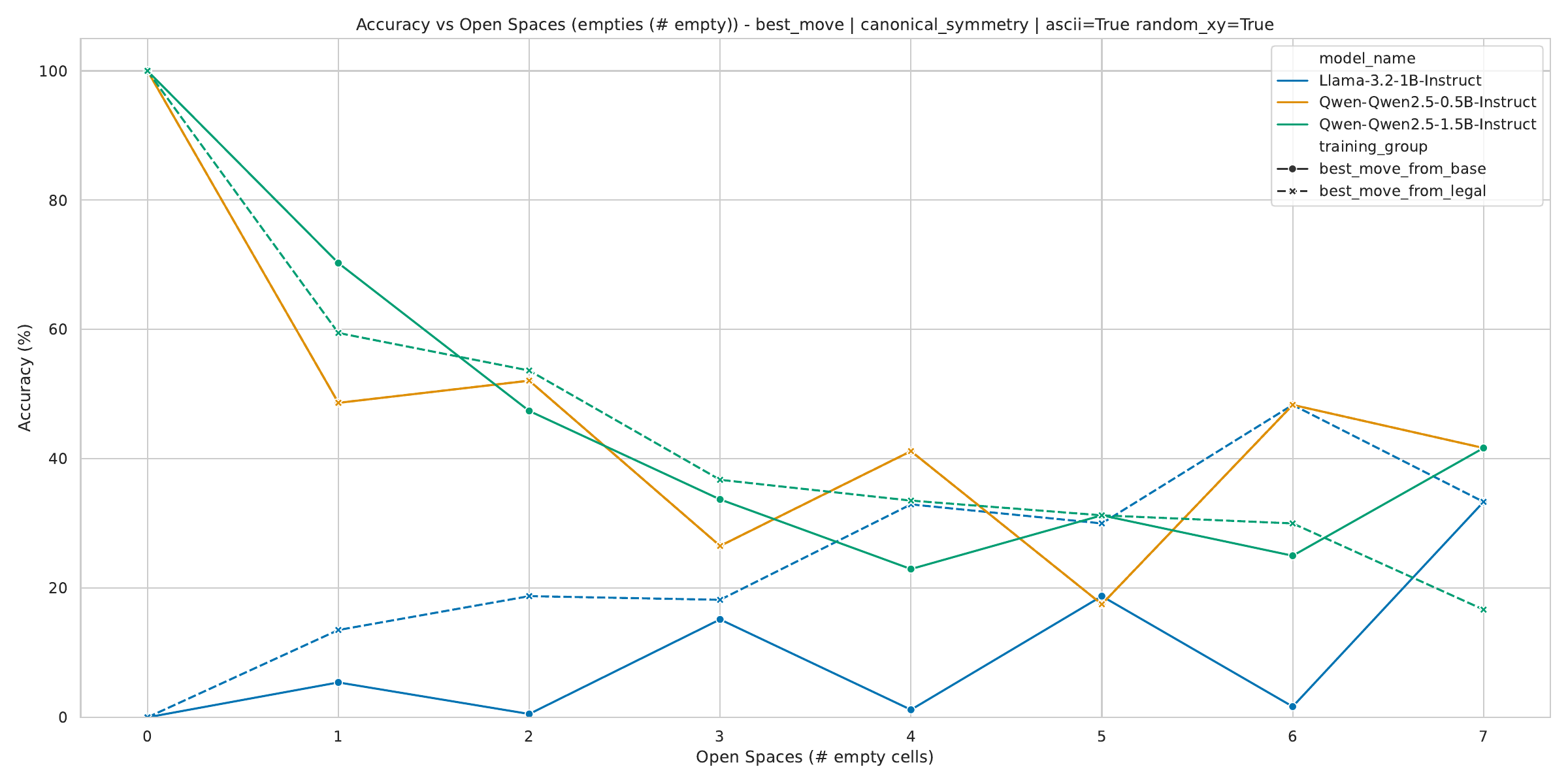}
        \caption{Open spaces, best move, ASCII board representation, random-True}
    \end{subfigure}

    % ---- Legal Move ----
    \begin{subfigure}[t]{0.45\textwidth}
        \includegraphics[width=\linewidth]{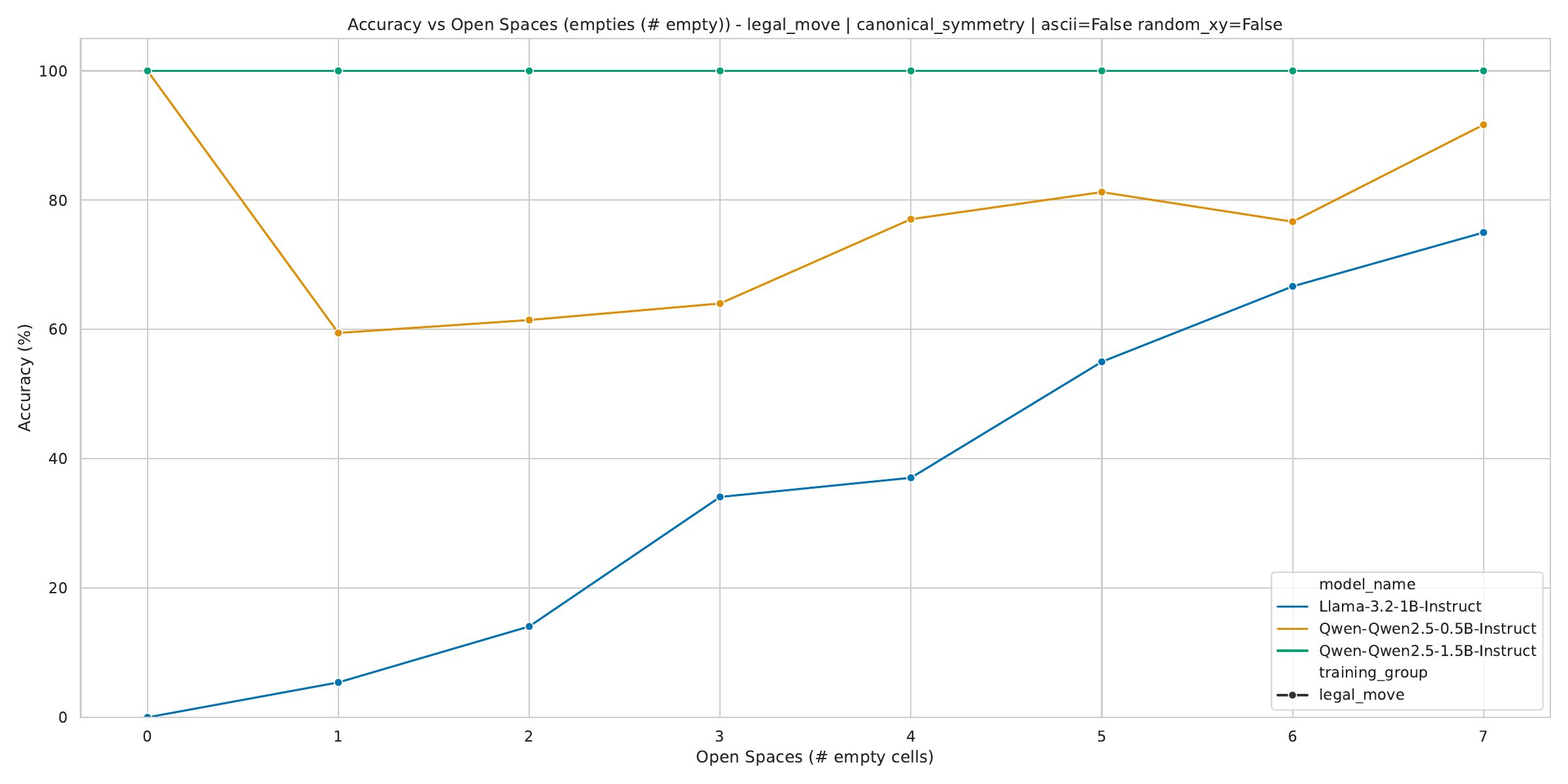}
        \caption{Open spaces, legal move, Natural language board representation, random-False}
    \end{subfigure}%
    \hfill
    \begin{subfigure}[t]{0.45\textwidth}
        \includegraphics[width=\linewidth]{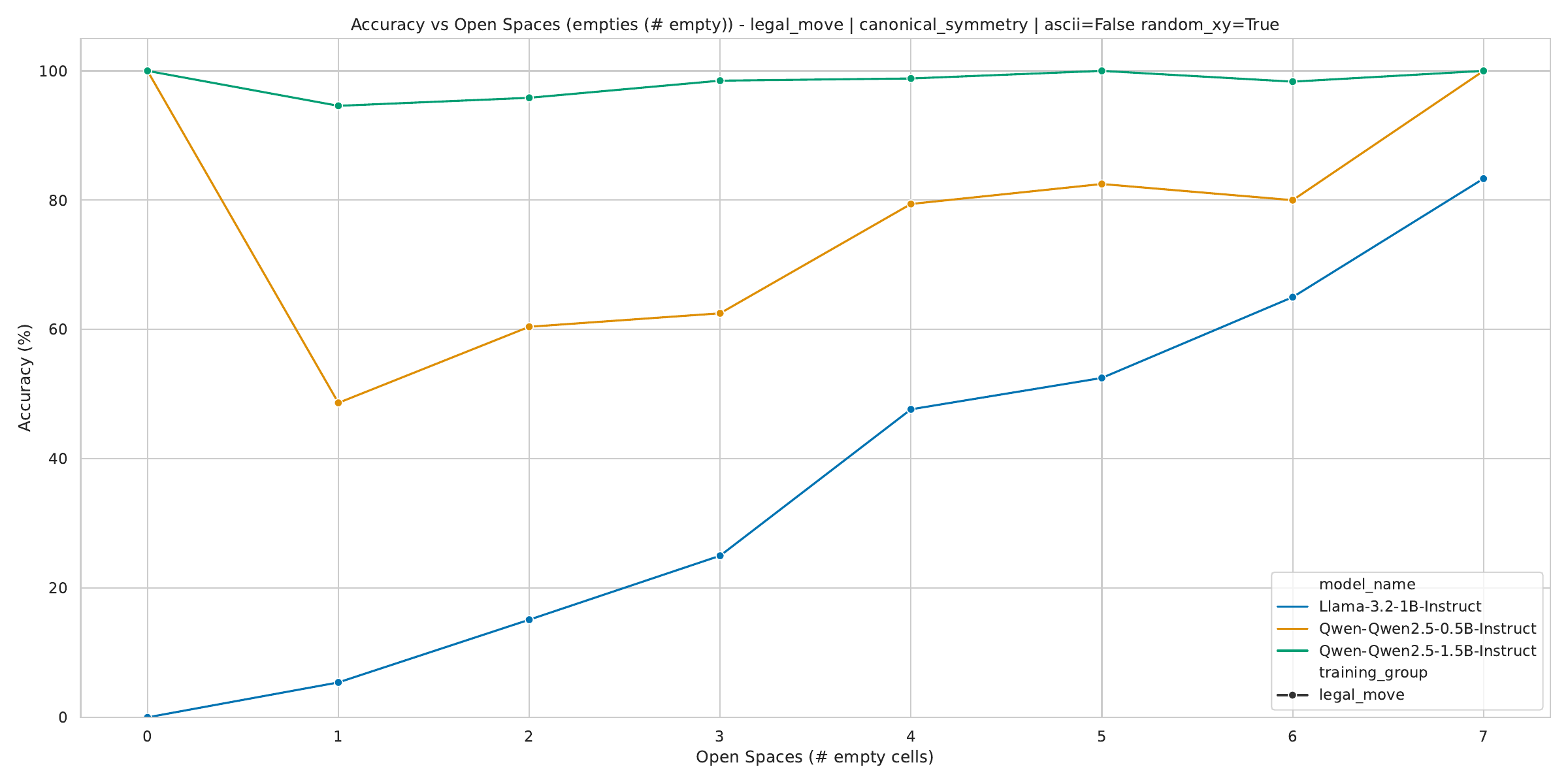}
        \caption{Open spaces, legal move, Natural language board representation, random-True}
    \end{subfigure}

    \begin{subfigure}[t]{0.45\textwidth}
        \includegraphics[width=\linewidth]{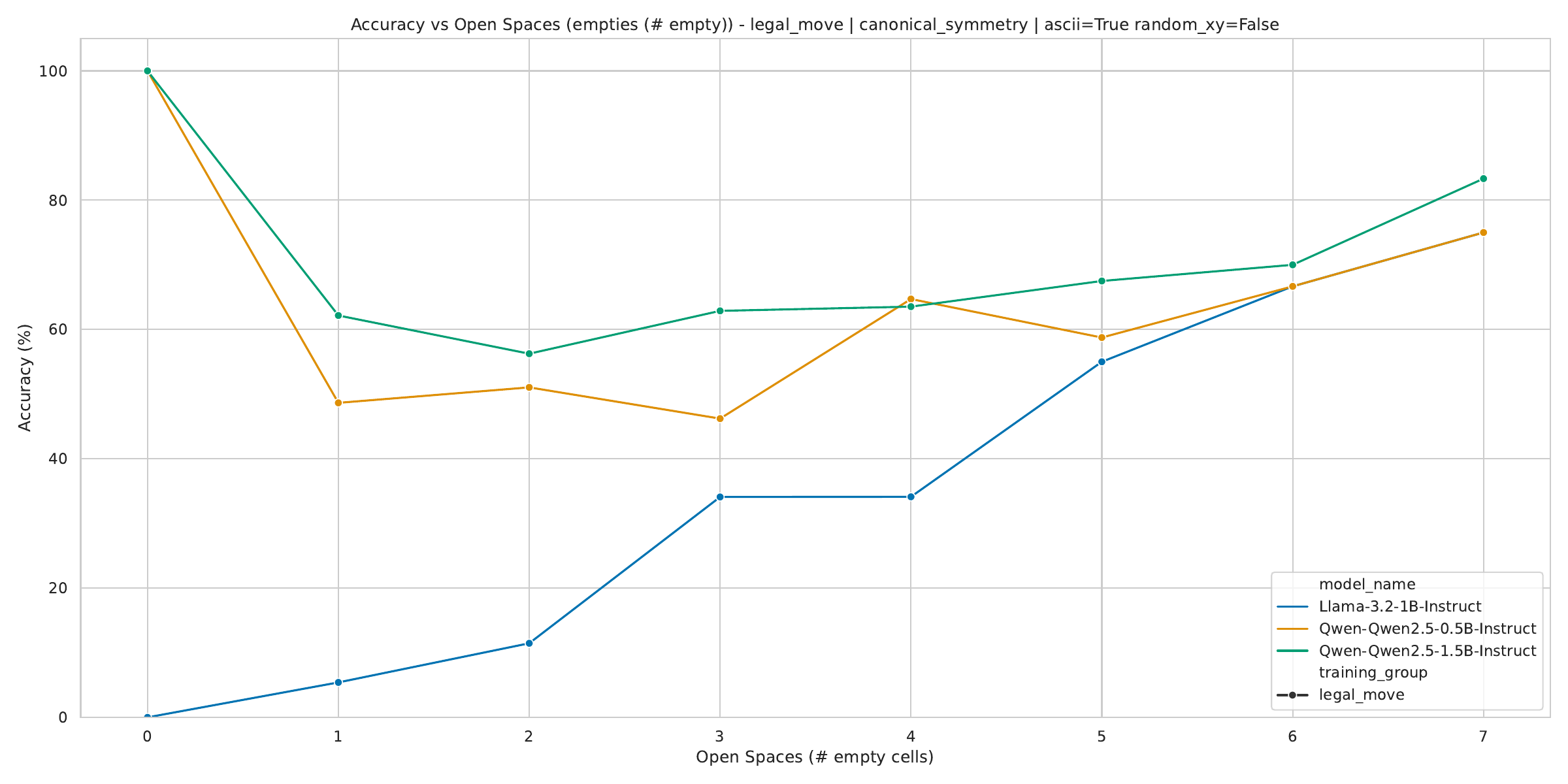}
        \caption{Open spaces, legal move, ASCII board representation, random-False}
    \end{subfigure}%
    \hfill
    \begin{subfigure}[t]{0.45\textwidth}
        \includegraphics[width=\linewidth]{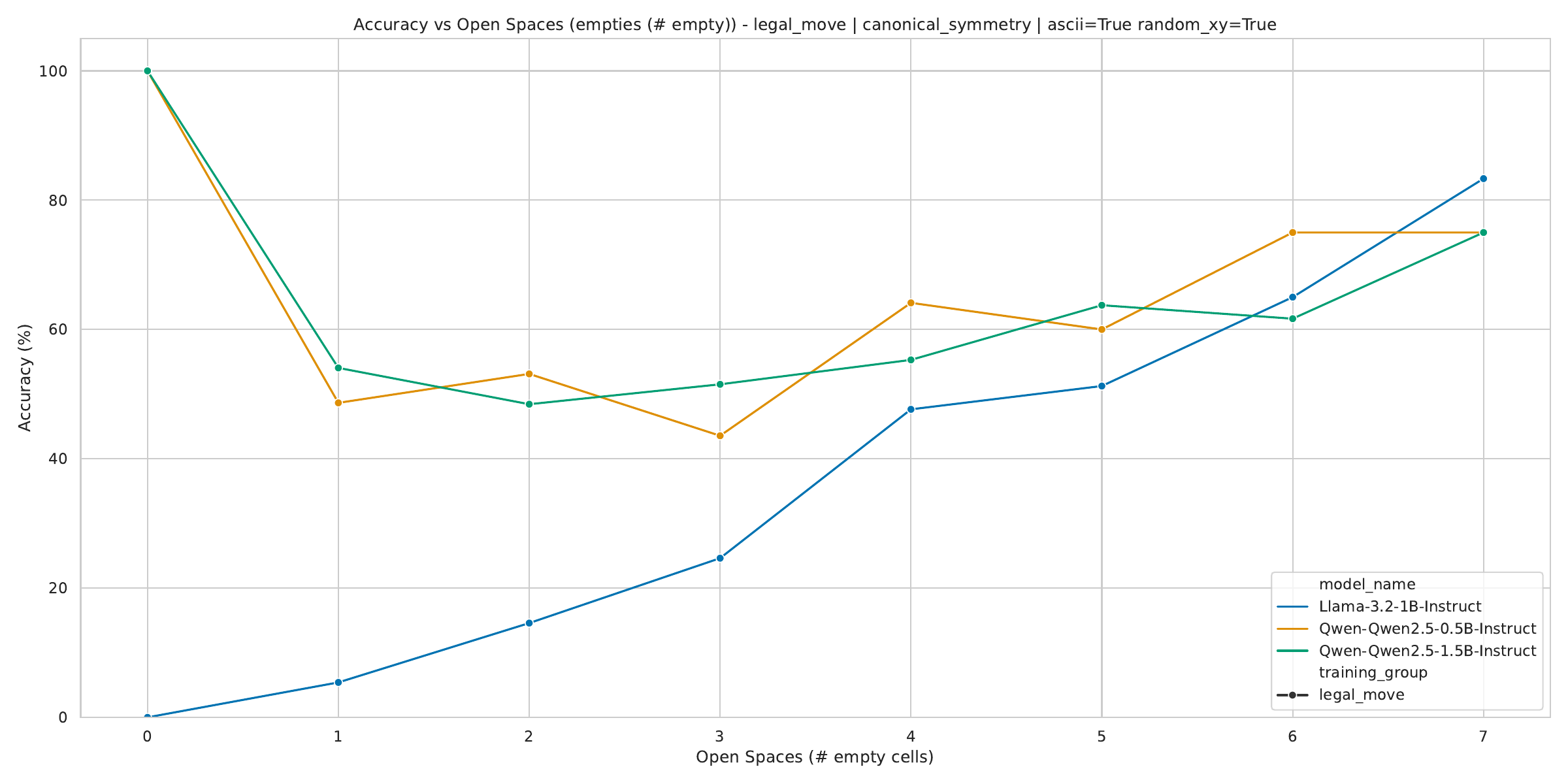}
        \caption{Open spaces, legal move, ASCII board representation, random-True}
    \end{subfigure}

    \caption{Open spaces analysis across board representations and randomization.}
    \label{fig:appendix-openspaces}
\end{figure*}

%%%%%%%%%%%%%%%%%%%%%%%%%%%%%%%%%%%%%%%%%%%%%%%%%%%%%%
% 5. Complexity
%%%%%%%%%%%%%%%%%%%%%%%%%%%%%%%%%%%%%%%%%%%%%%%%%%%%%%
\begin{figure*}[ht]
    \centering
    % ---- Best Move ----
    \begin{subfigure}[t]{0.45\textwidth}
        \includegraphics[width=\linewidth]{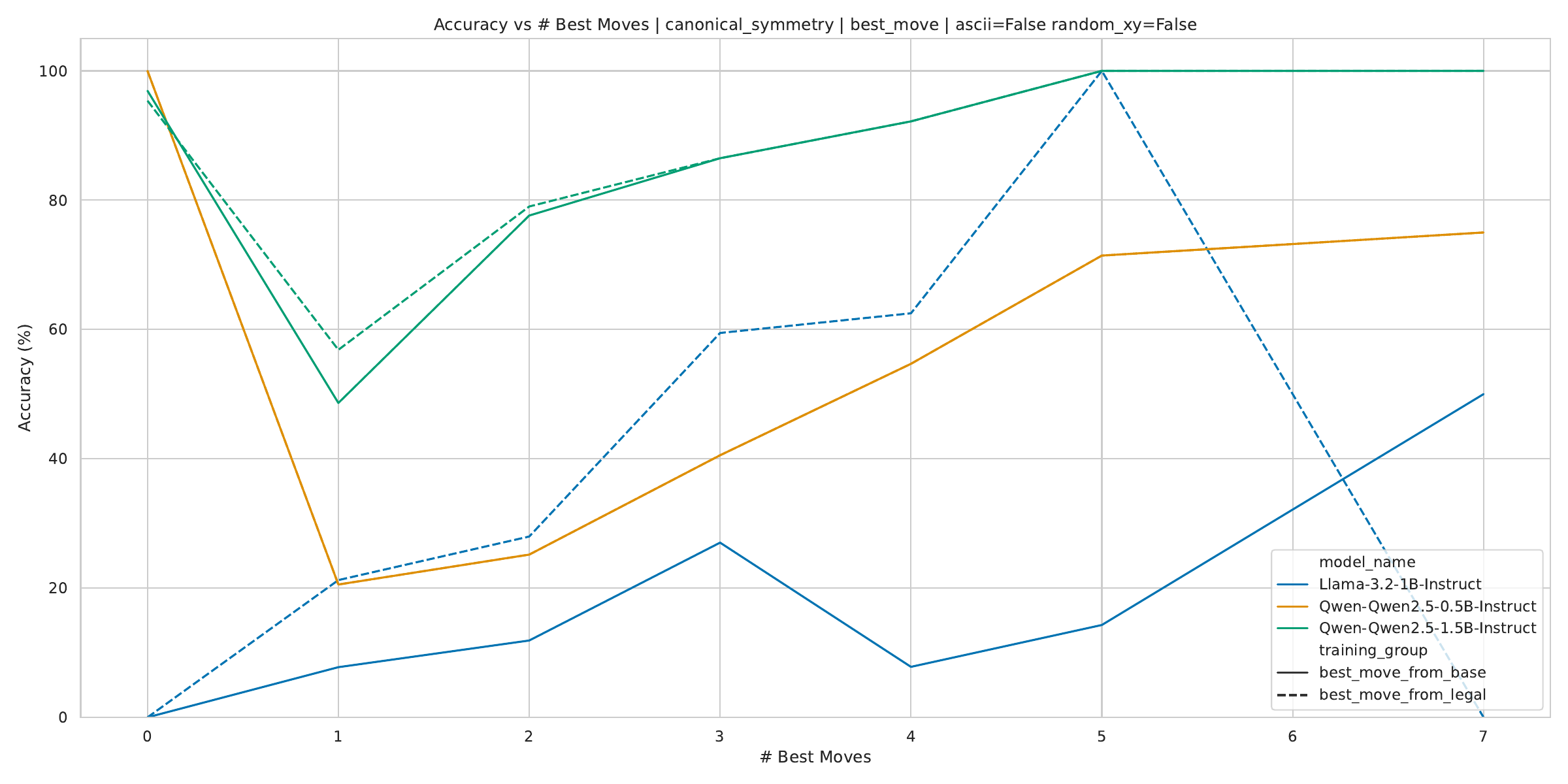}
        \caption{Complexity, best move, Natural language board representation, random-False}
    \end{subfigure}%
    \hfill
    \begin{subfigure}[t]{0.45\textwidth}
        \includegraphics[width=\linewidth]{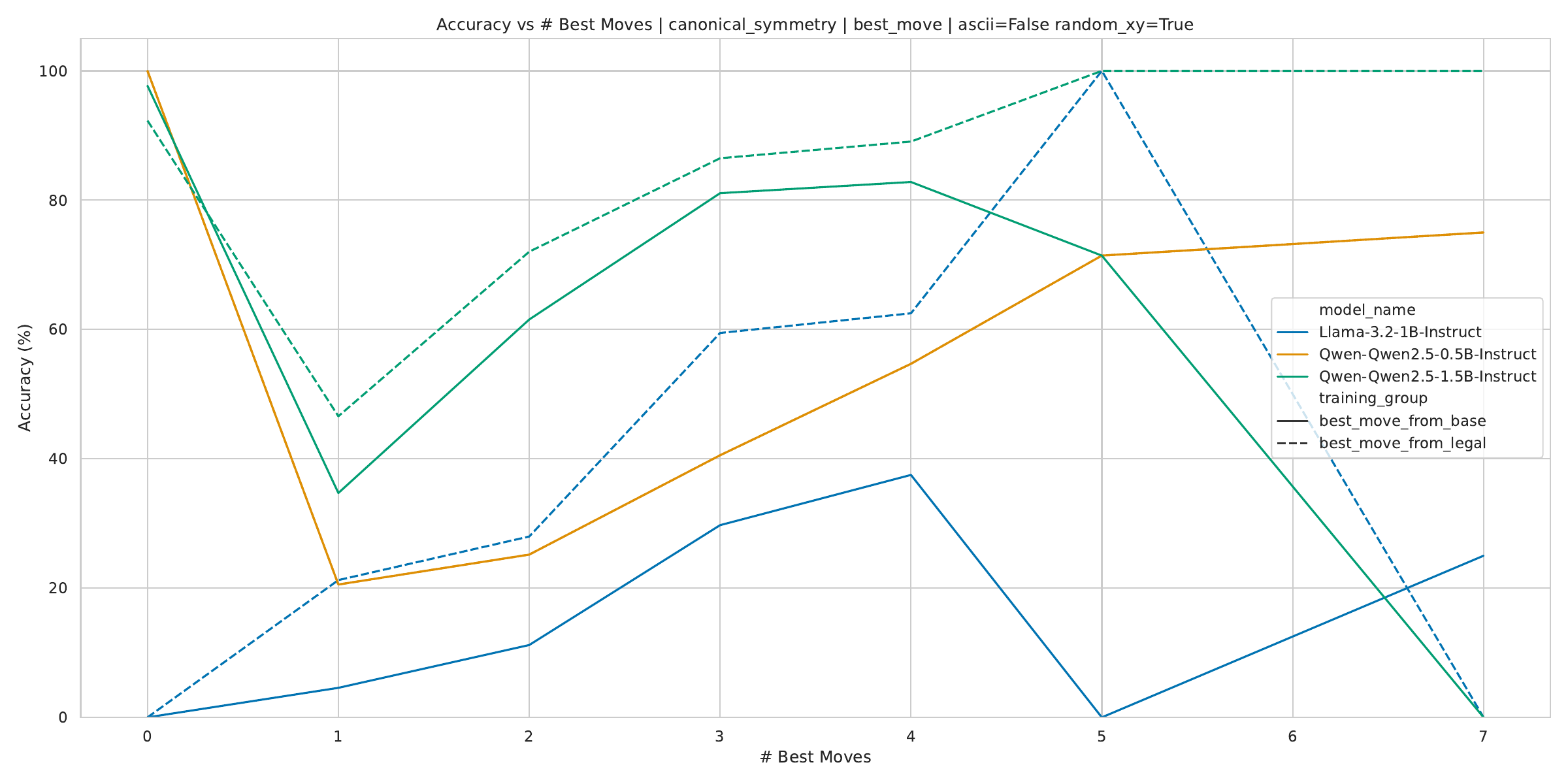}
        \caption{Complexity, best move, Natural language board representation, random-True}
    \end{subfigure}

    \begin{subfigure}[t]{0.45\textwidth}
        \includegraphics[width=\linewidth]{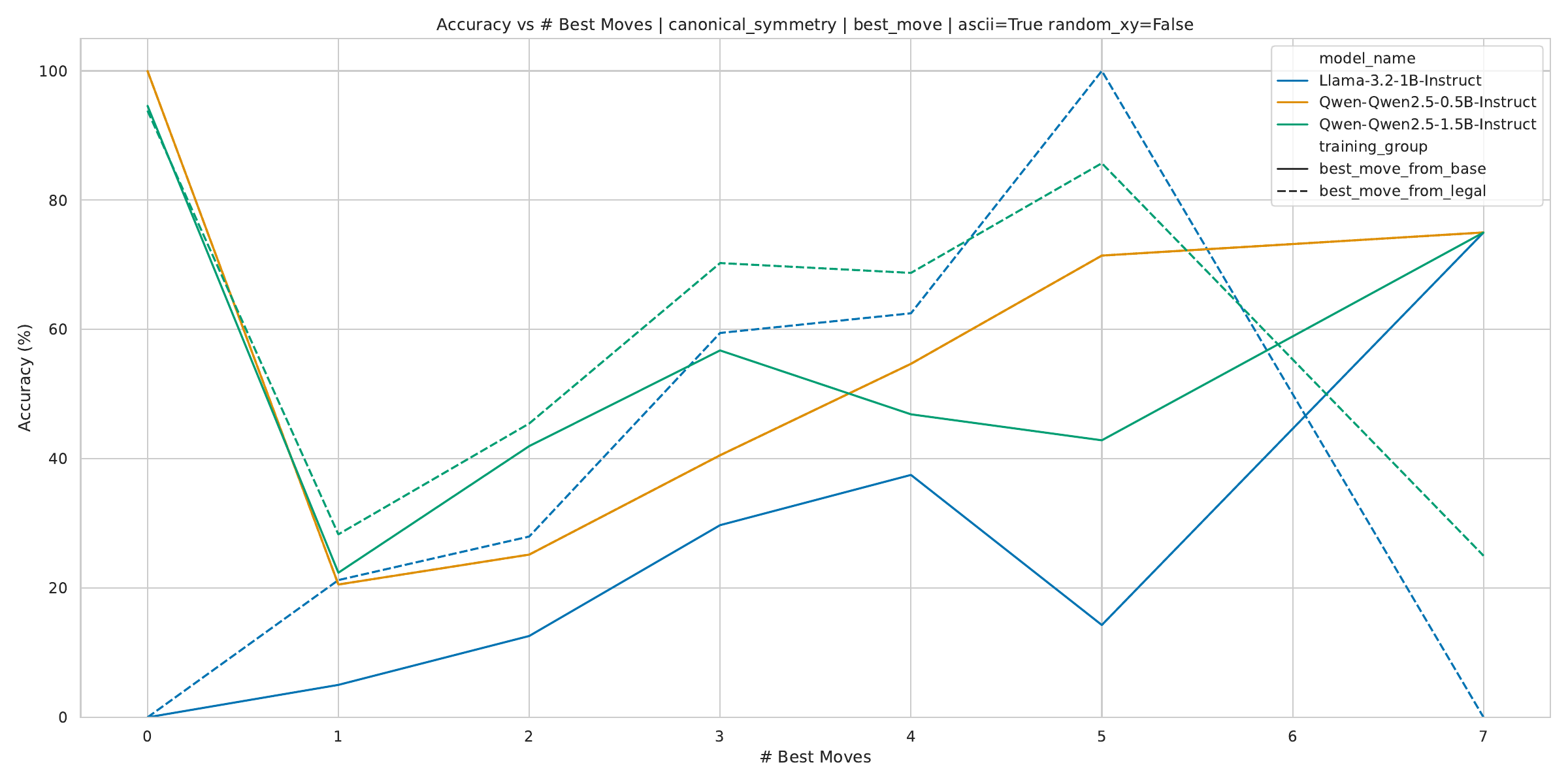}
        \caption{Complexity, best move, ASCII board representation, random-False}
    \end{subfigure}%
    \hfill
    \begin{subfigure}[t]{0.45\textwidth}
        \includegraphics[width=\linewidth]{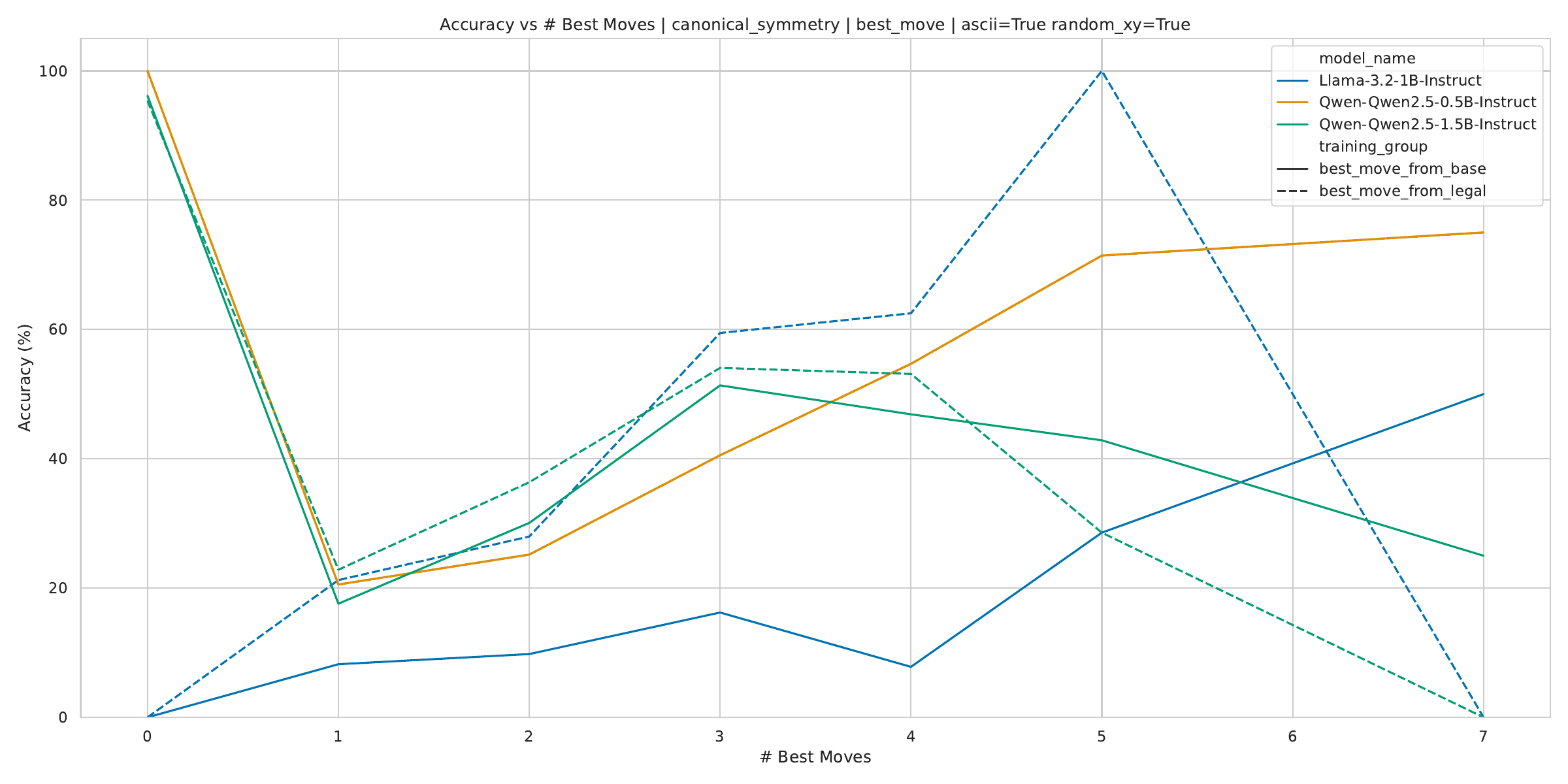}
        \caption{Complexity, best move, ASCII board representation, random-True}
    \end{subfigure}

    % ---- Legal Move ----
    \begin{subfigure}[t]{0.45\textwidth}
        \includegraphics[width=\linewidth]{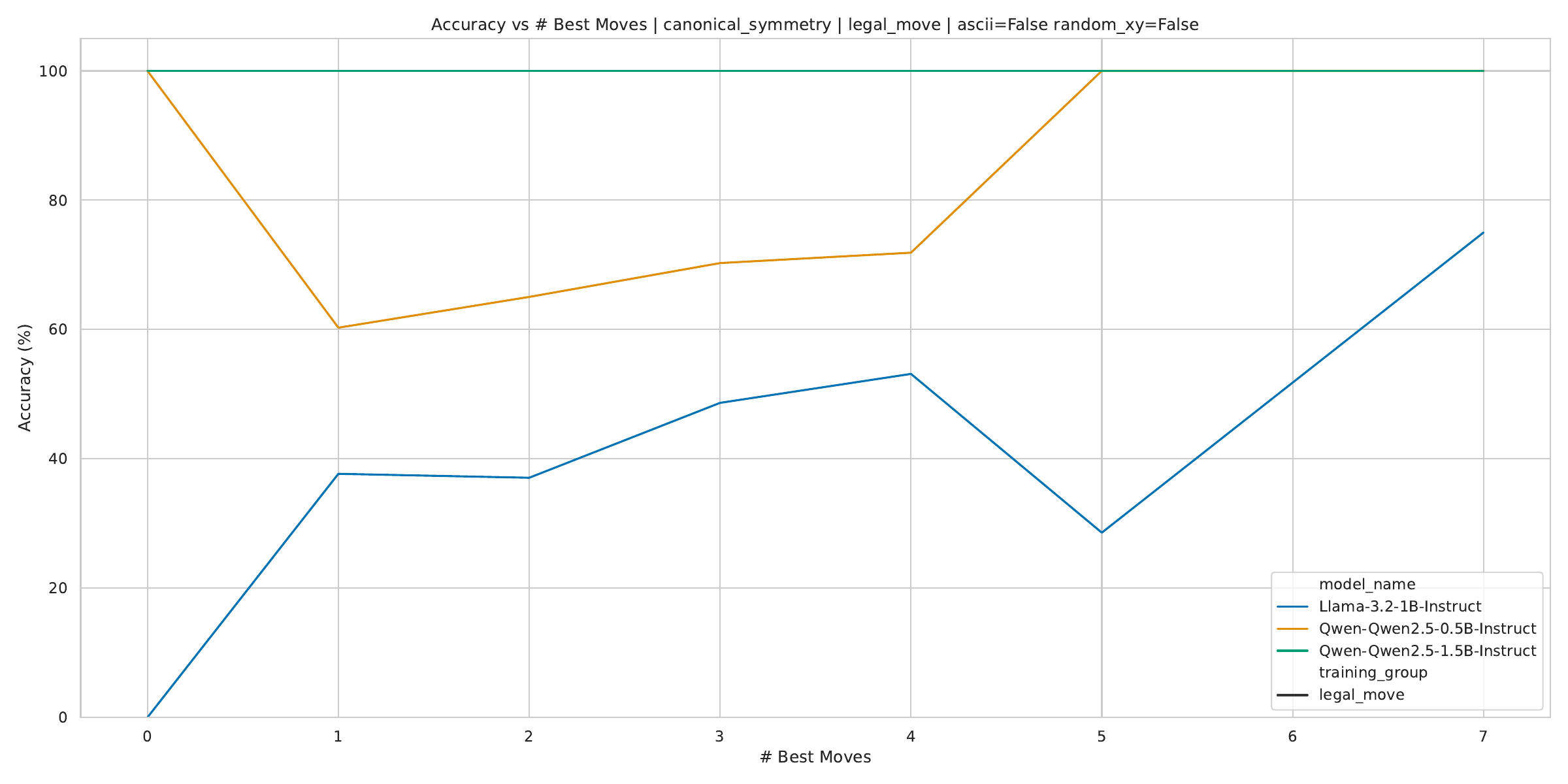}
        \caption{Complexity, legal move, Natural language board representation, random-False}
    \end{subfigure}%
    \hfill
    \begin{subfigure}[t]{0.45\textwidth}
        \includegraphics[width=\linewidth]{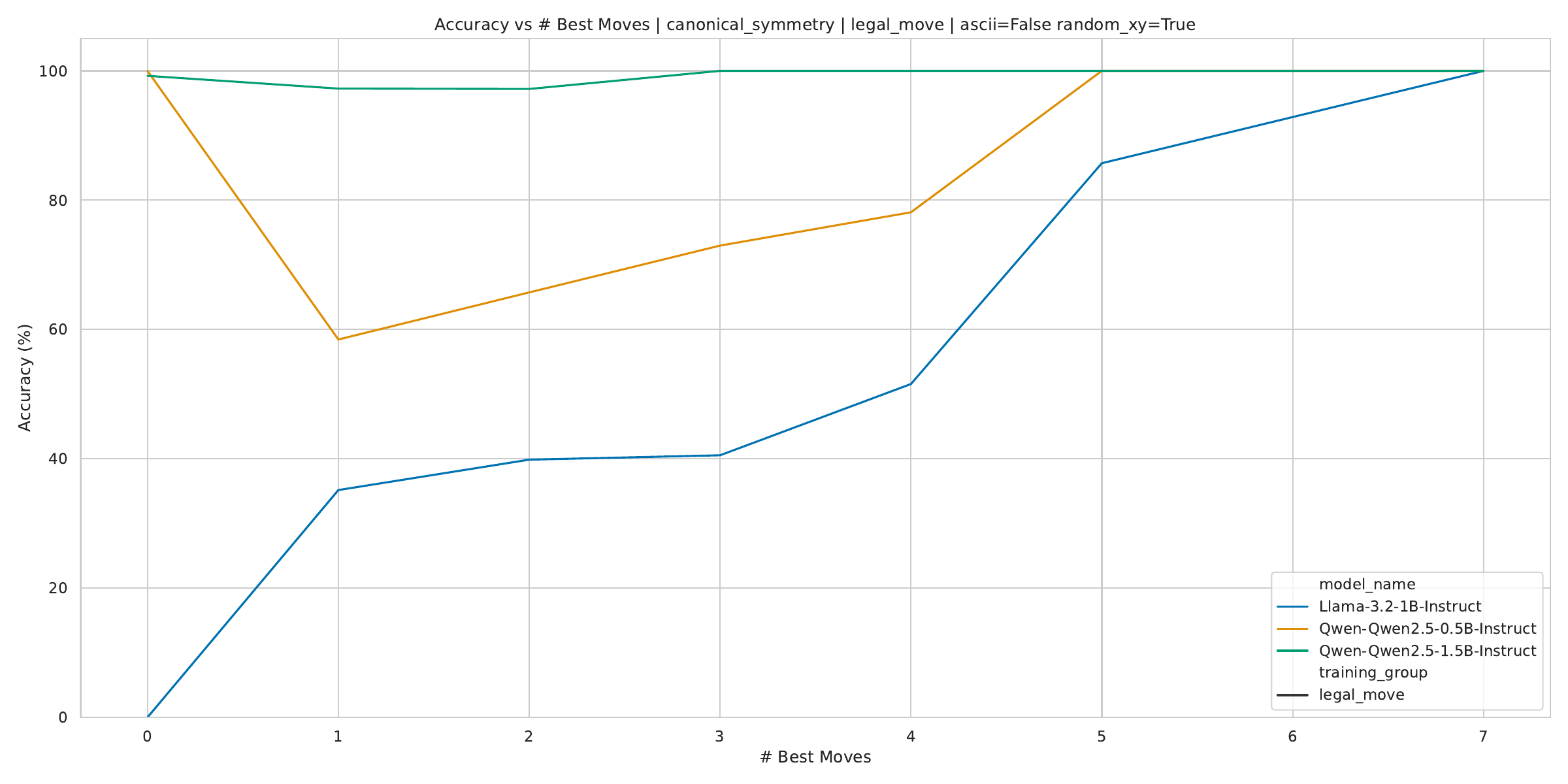}
        \caption{Complexity, legal move, Natural language board representation, random-True}
    \end{subfigure}

    \begin{subfigure}[t]{0.45\textwidth}
        \includegraphics[width=\linewidth]{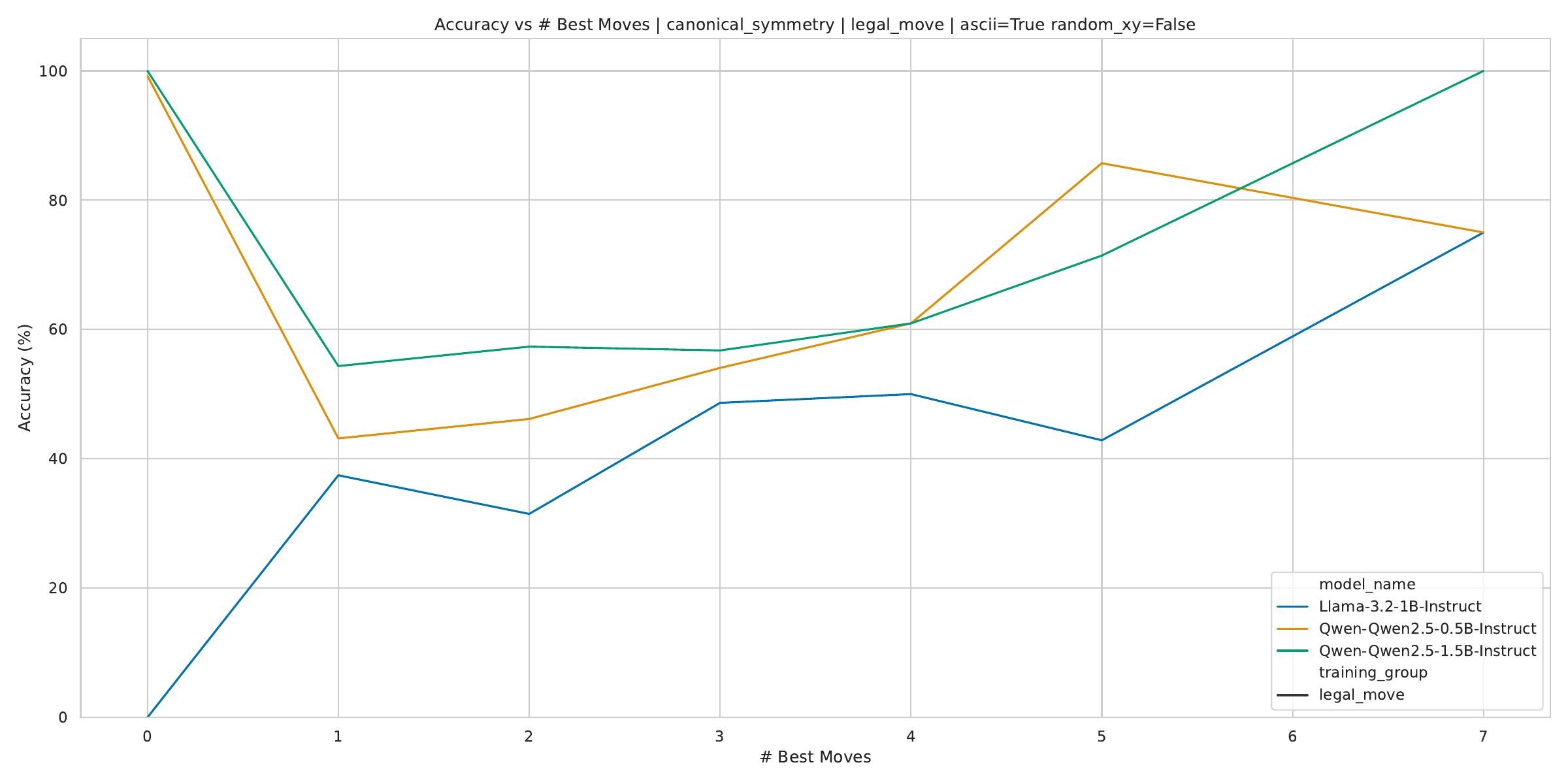}
        \caption{Complexity, legal move, ASCII board representation, random-False}
    \end{subfigure}%
    \hfill
    \begin{subfigure}[t]{0.45\textwidth}
        \includegraphics[width=\linewidth]{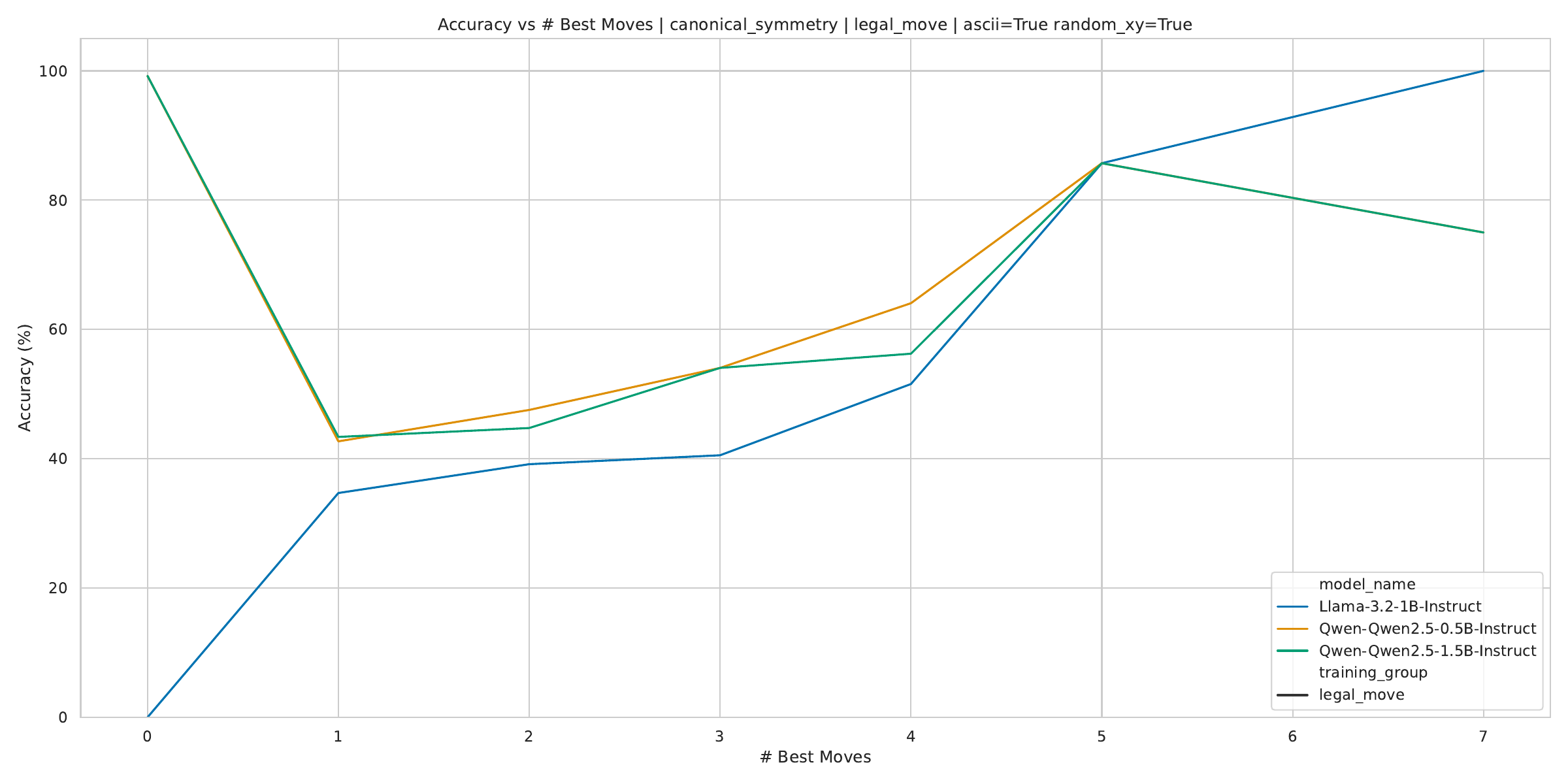}
        \caption{Complexity, legal move, ASCII board representation, random-True}
    \end{subfigure}

    \caption{Complexity analysis across board representations and randomization.}
    \label{fig:appendix-complexity}
\end{figure*}

\clearpage

\section{Additional Mechanistic Interpretability Results}
\label{sec:mech-interp-appendix}
The plots below map the hypotheses from SAE probing across all layers and along training checkpoints for both natural language and ascii representations. Due to size constraints on the main paper, the plots have been compressed to allow the PDF to stay within 50MB. Complete plots have been provided with the supplemental submission for further investigation.

% ------------------------------
% Game Progression
% ------------------------------
\begin{figure*}[ht]
    \centering
    \begin{subfigure}[t]{0.48\textwidth}
        \includegraphics[width=\linewidth]{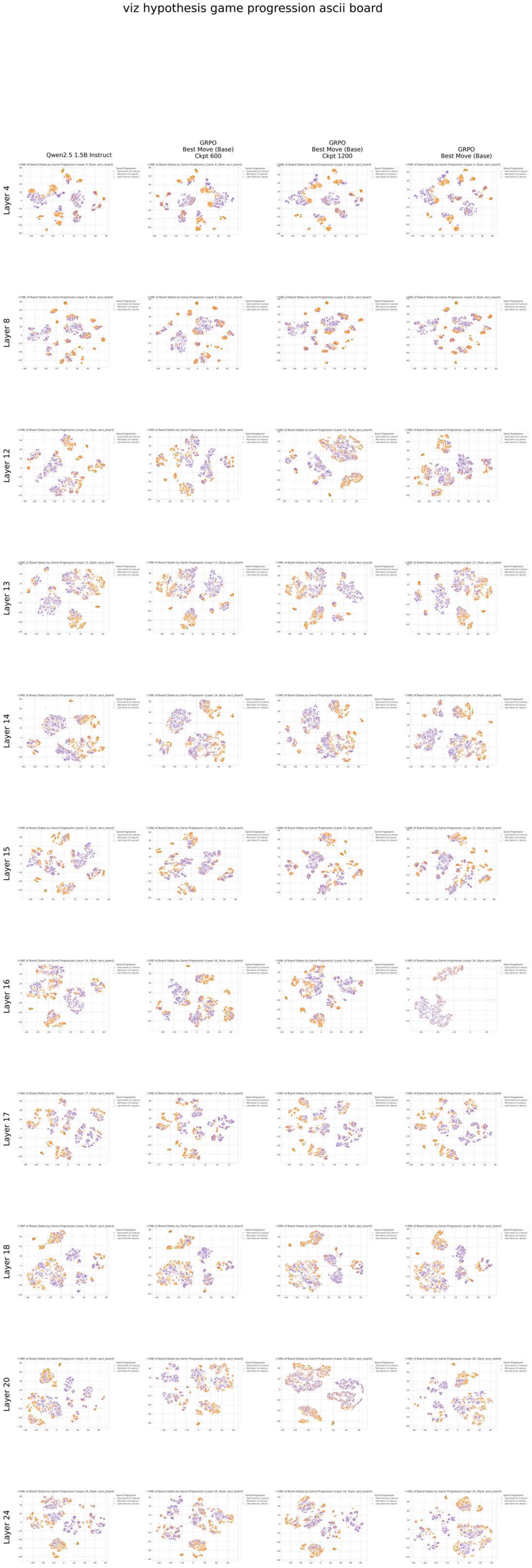}
        \caption{ASCII board representation}
    \end{subfigure}%
    \hfill
    \begin{subfigure}[t]{0.48\textwidth}
        \includegraphics[width=\linewidth]{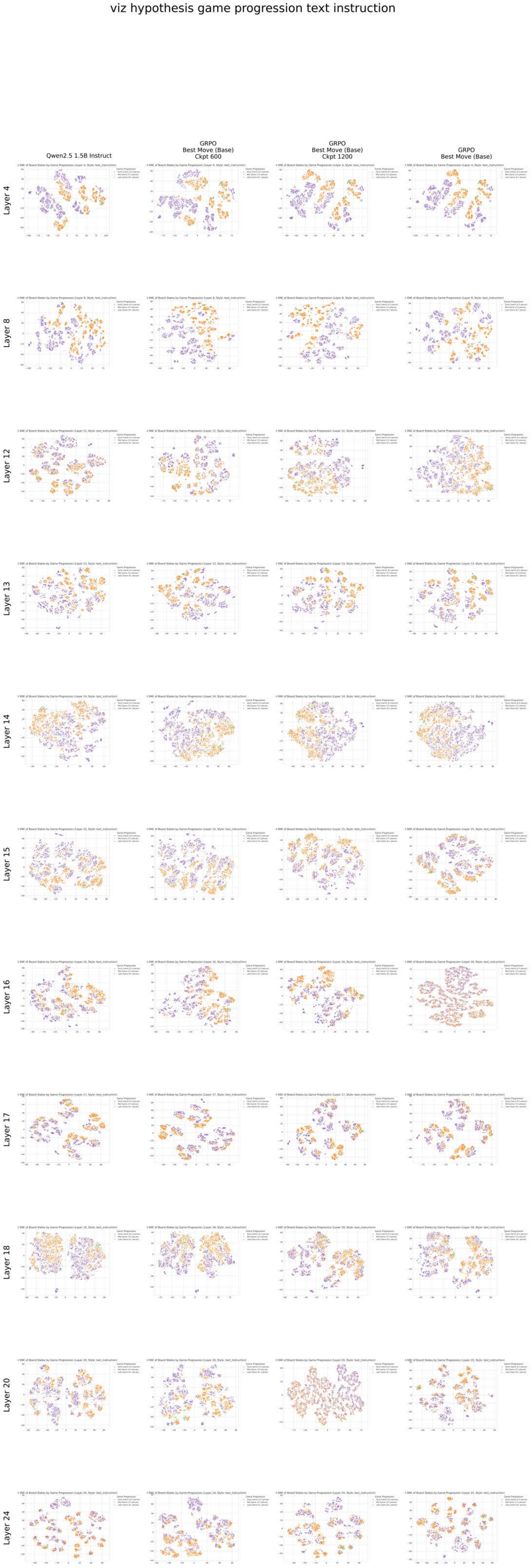}
        \caption{Natural language instruction representation}
    \end{subfigure}
    \caption{Hypothesis testing: Game progression.}
    \label{fig:appendix-interp-game-progression}
\end{figure*}

% ------------------------------
% Game Trajectories
% ------------------------------
\begin{figure*}[ht]
    \centering
    \begin{subfigure}[t]{0.48\textwidth}
        \includegraphics[width=\linewidth]{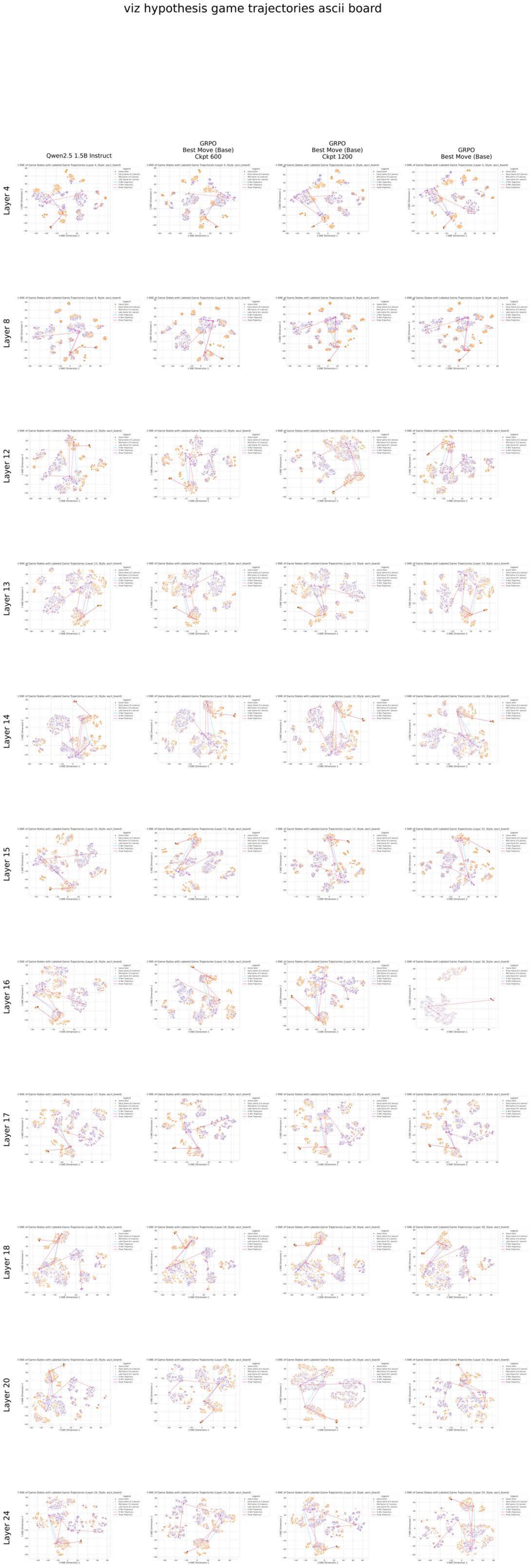}
        \caption{ASCII board representation}
    \end{subfigure}%
    \hfill
    \begin{subfigure}[t]{0.48\textwidth}
        \includegraphics[width=\linewidth]{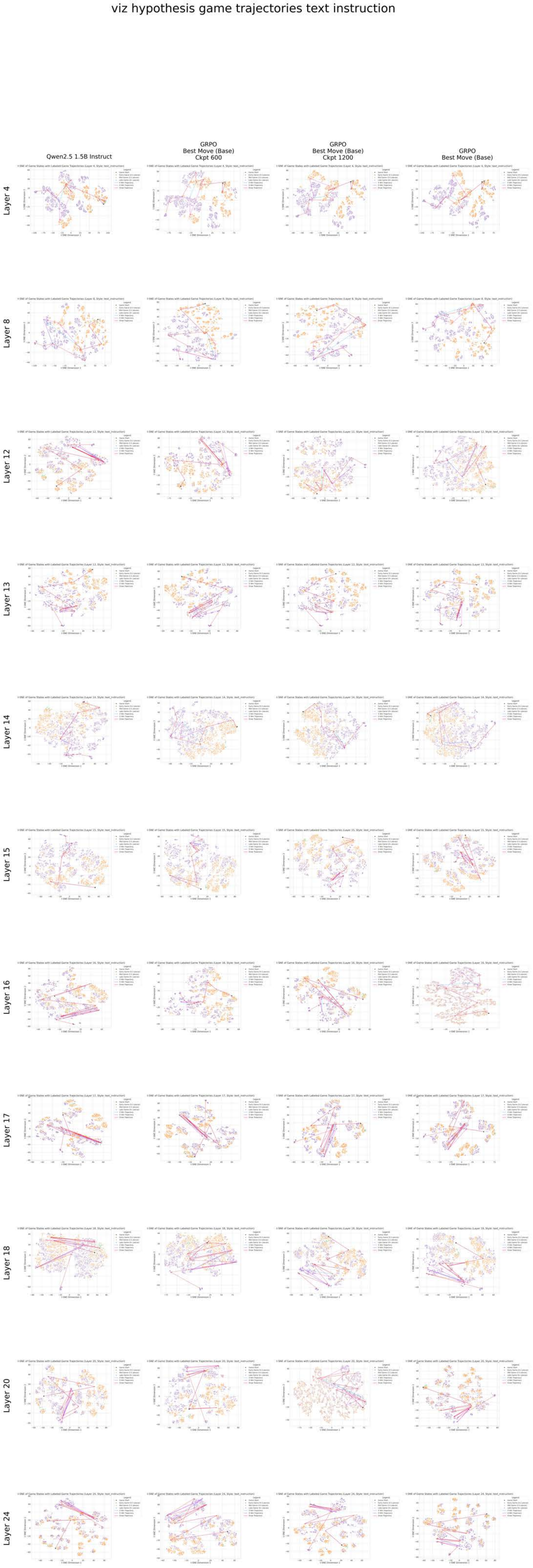}
        \caption{Natural language instruction representation}
    \end{subfigure}
    \caption{Hypothesis testing: Game trajectories.}
    \label{fig:appendix-interp-game-traj}
\end{figure*}

% ------------------------------
% Game Turn
% ------------------------------
\begin{figure*}[ht]
    \centering
    \begin{subfigure}[t]{0.48\textwidth}
        \includegraphics[width=\linewidth]{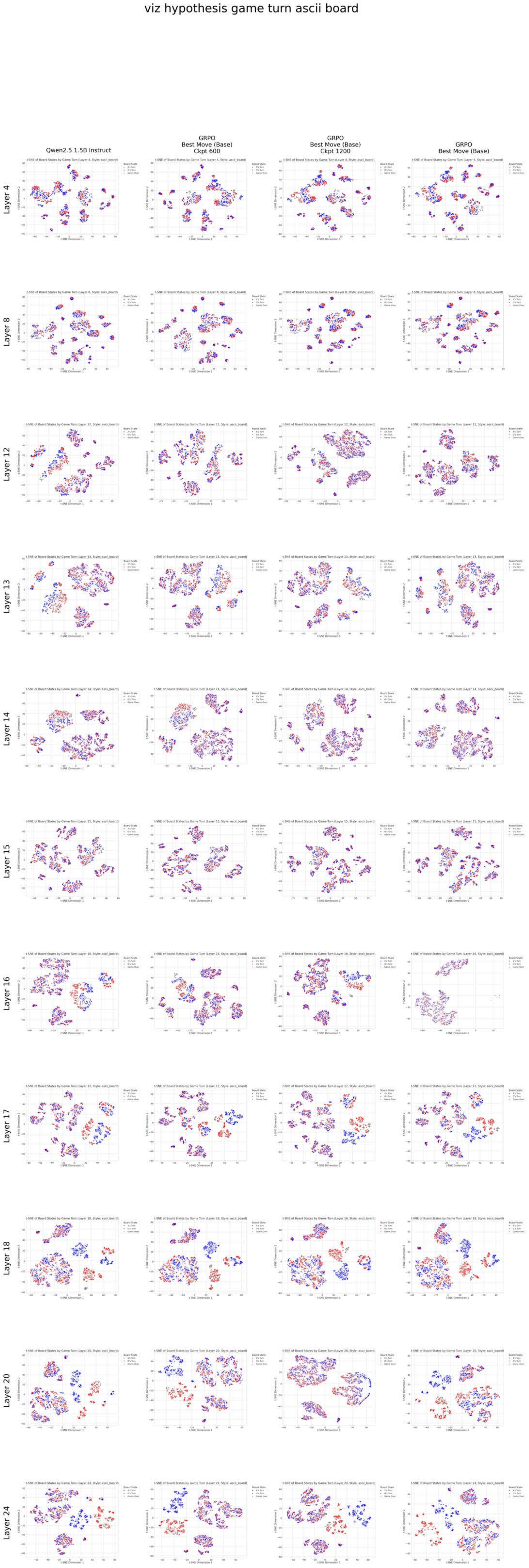}
        \caption{ASCII board representation}
    \end{subfigure}%
    \hfill
    \begin{subfigure}[t]{0.48\textwidth}
        \includegraphics[width=\linewidth]{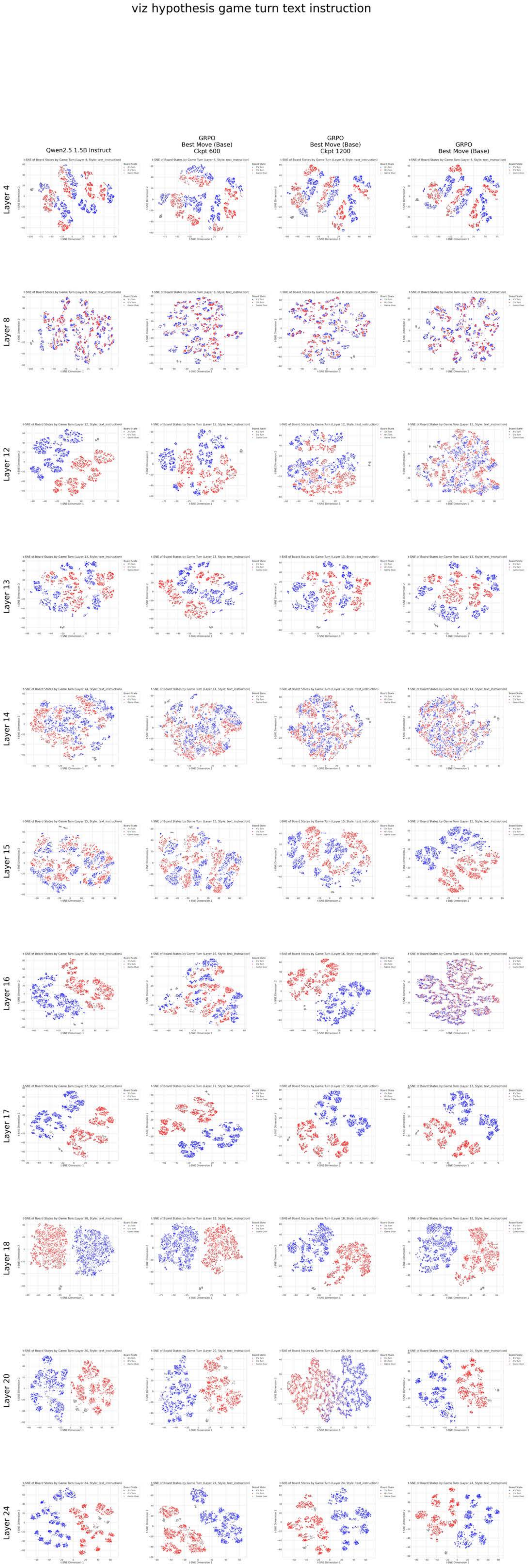}
        \caption{Natural language instruction representation}
    \end{subfigure}
    \caption{Hypothesis testing: Game turn.}
    \label{fig:appendix-interp-game-turn}
\end{figure*}

% ------------------------------
% Prediction Correctness
% ------------------------------
\begin{figure*}[ht]
    \centering
    \begin{subfigure}[t]{0.48\textwidth}
        \includegraphics[width=\linewidth,height=0.9\textheight,keepaspectratio]{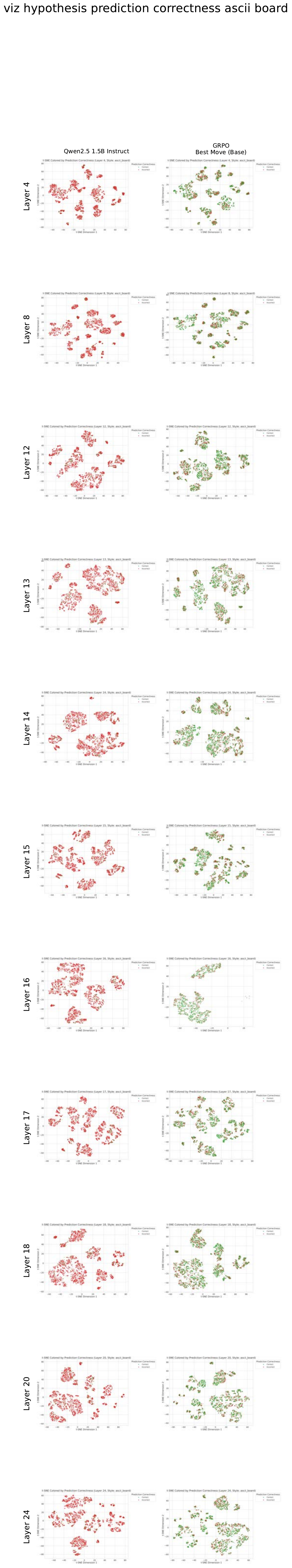}
        \caption{ASCII board representation}
    \end{subfigure}%
    \hfill
    \begin{subfigure}[t]{0.48\textwidth}
        \includegraphics[width=\linewidth,height=0.9\textheight,keepaspectratio]{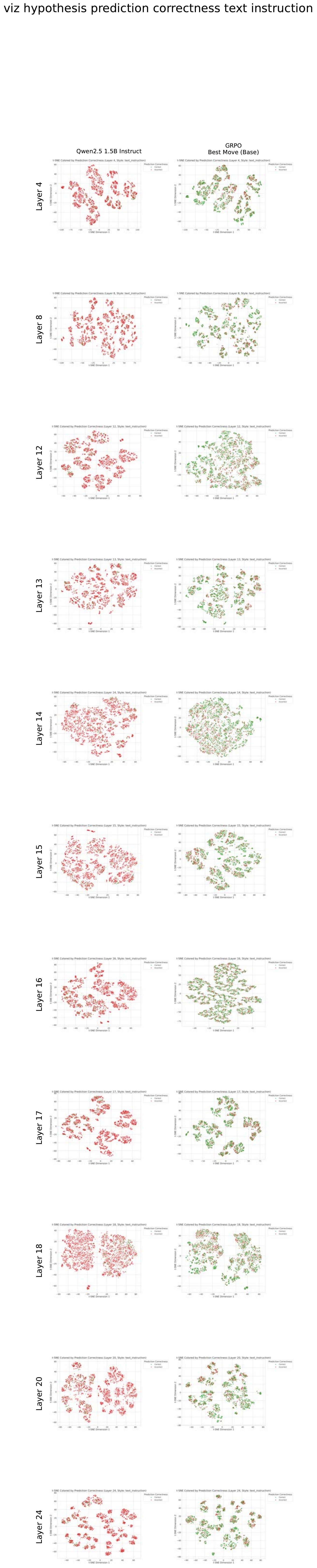}
        \caption{Natural language instruction representation}
    \end{subfigure}
    \caption{Hypothesis testing: Prediction correctness.}
    \label{fig:appendix-interp-pred-correctness}
\end{figure*}

% ------------------------------
% Illegal vs Legal
% ------------------------------
\begin{figure*}[ht]
    \centering
    \begin{subfigure}[t]{0.48\textwidth}
        \includegraphics[width=\linewidth,height=0.9\textheight,keepaspectratio]{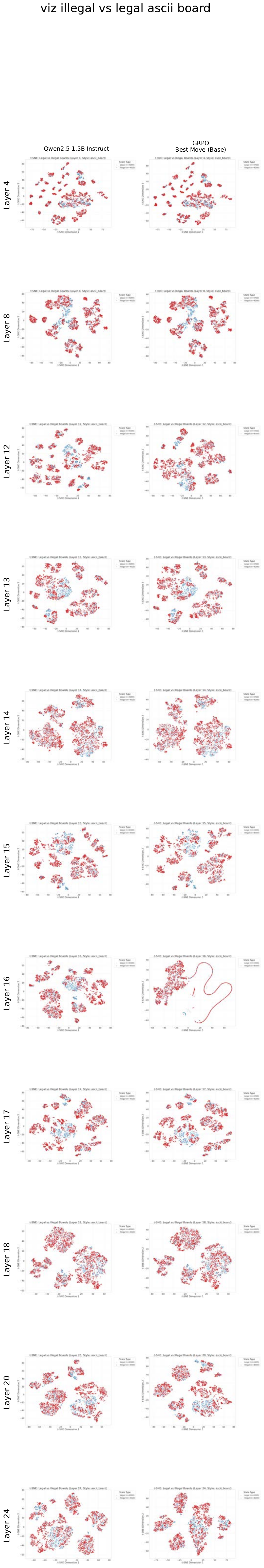}
        \caption{ASCII board representation}
    \end{subfigure}%
    \hfill
    \begin{subfigure}[t]{0.40\textwidth}
        \includegraphics[width=\linewidth,height=0.9\textheight,keepaspectratio]{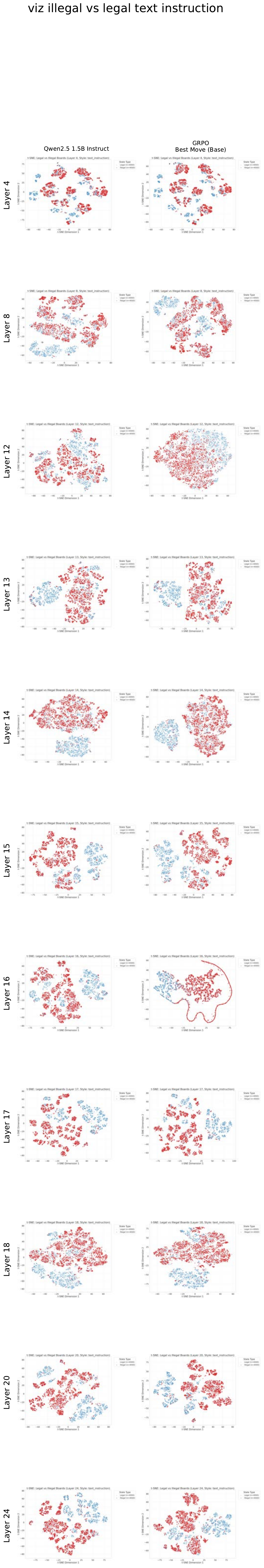}
        \caption{Natural language instruction representation}
    \end{subfigure}
    \caption{Hypothesis testing: Illegal vs Legal boards.}
    \label{fig:appendix-interp-illeval-vs-legal}
\end{figure*}

% ------------------------------
% Strategic Situation (Game-theoretic)
% ------------------------------
\begin{figure*}[ht]
    \centering
    \begin{subfigure}[t]{0.48\textwidth}
        \includegraphics[width=\linewidth]{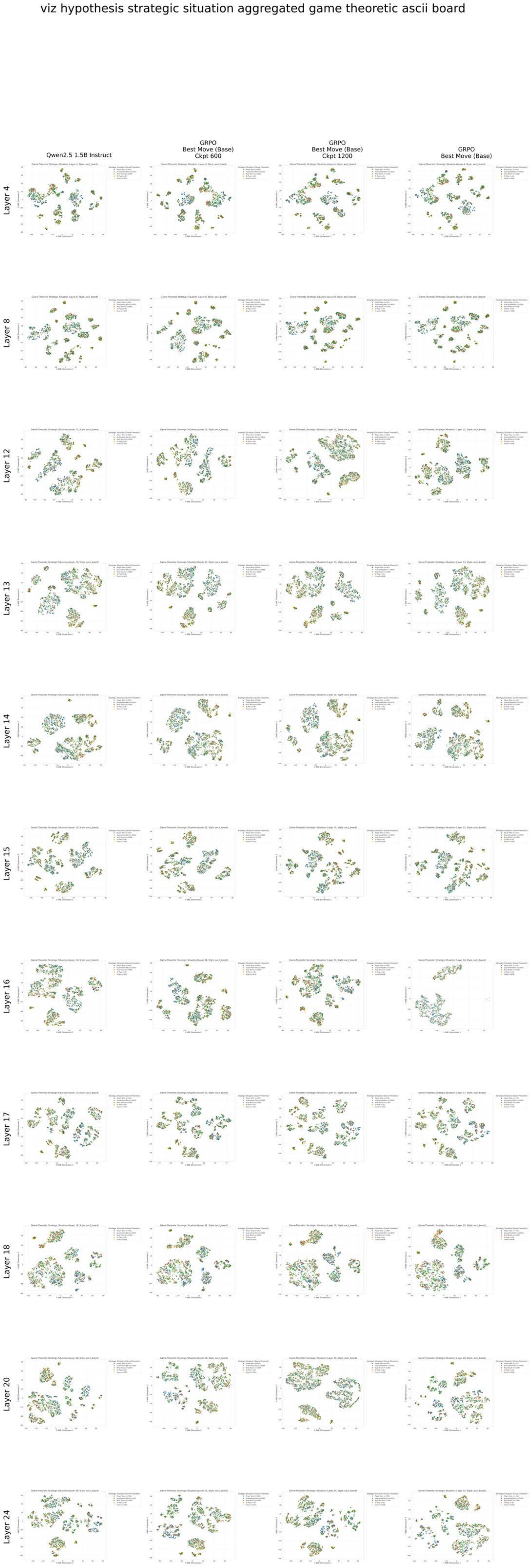}
        \caption{ASCII board representation}
    \end{subfigure}%
    \hfill
    \begin{subfigure}[t]{0.48\textwidth}
        \includegraphics[width=\linewidth]{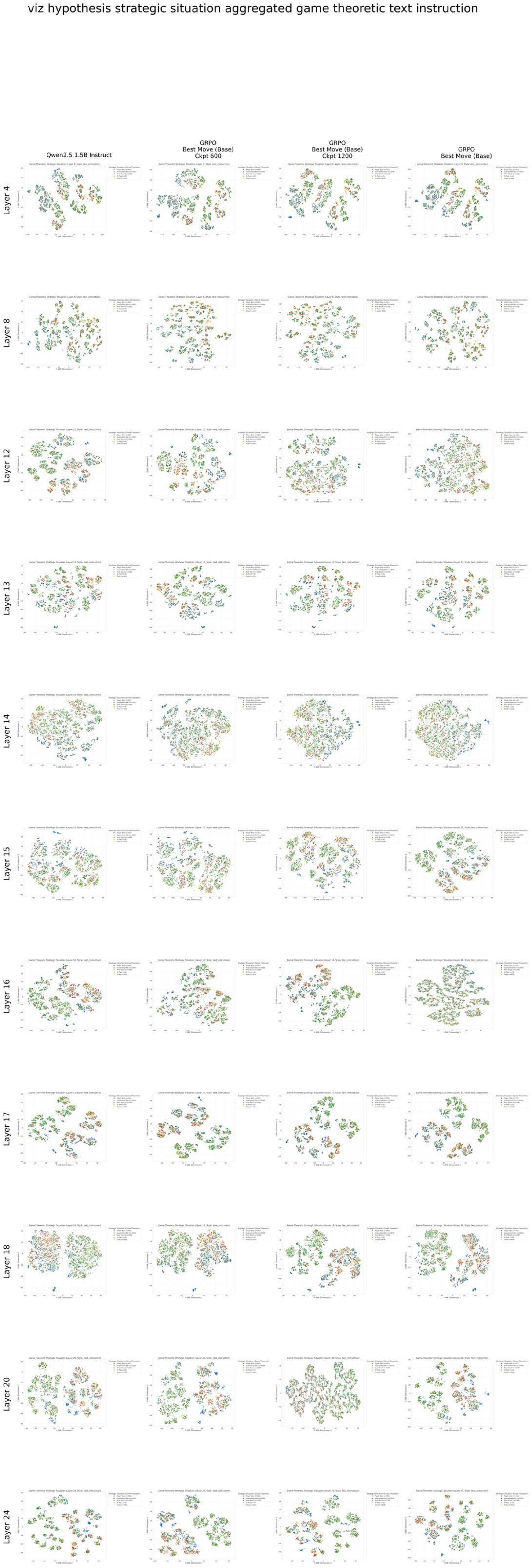}
        \caption{Natural language instruction representation}
    \end{subfigure}
    \caption{Hypothesis testing: Strategic situations.}
    \label{fig:appendix-interp-strat-situation}
\end{figure*}

% ------------------------------
% Symmetry
% ------------------------------
\begin{figure*}[ht]
    \centering
    \begin{subfigure}[t]{0.48\textwidth}
        \includegraphics[width=\linewidth]{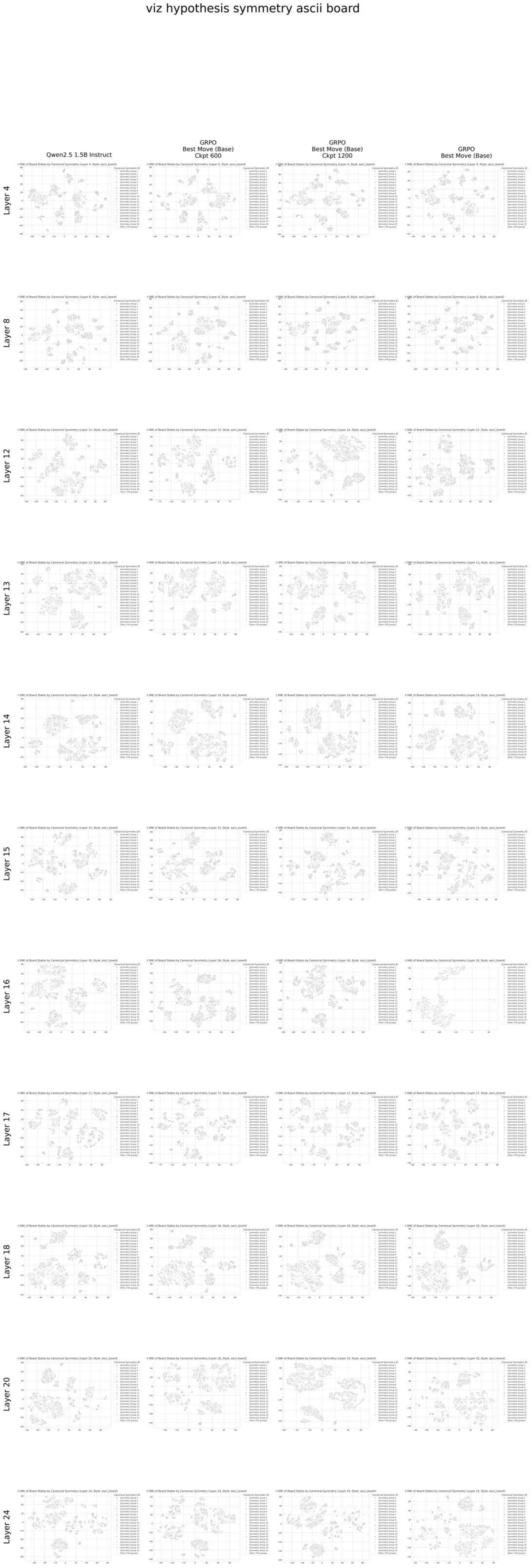}
        \caption{ASCII board representation}
    \end{subfigure}%
    \hfill
    \begin{subfigure}[t]{0.48\textwidth}
        \includegraphics[width=\linewidth]{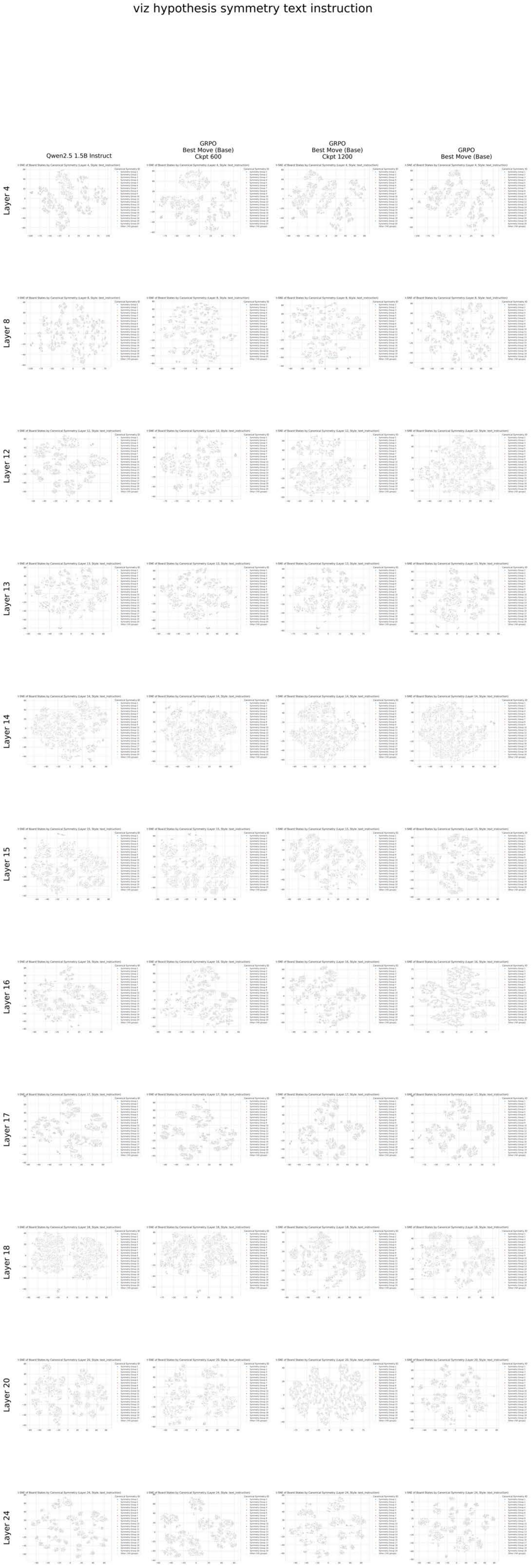}
        \caption{Natural language instruction representation}
    \end{subfigure}
    \caption{Hypothesis testing: Symmetry.}
    \label{fig:appendix-interp-symmetry}
\end{figure*}

% ------------------------------
% Threat Detection
% ------------------------------
\begin{figure*}[ht]
    \centering
    \begin{subfigure}[t]{0.48\textwidth}
        \includegraphics[width=\linewidth]{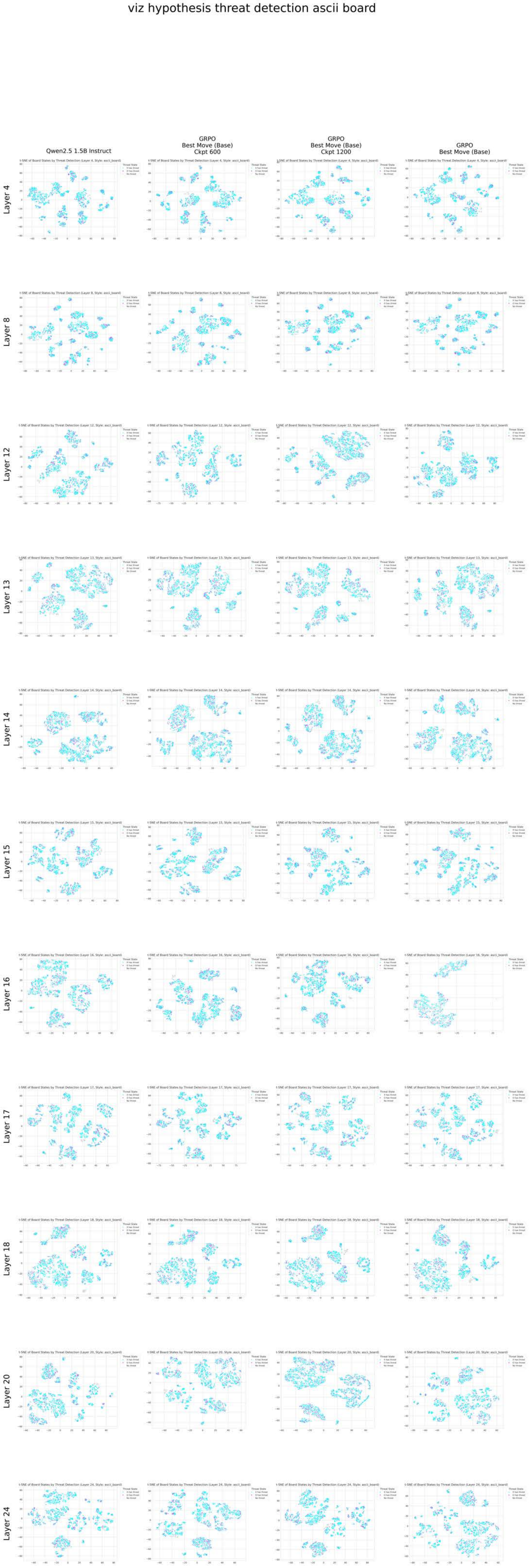}
        \caption{ASCII board representation}
    \end{subfigure}%
    \hfill
    \begin{subfigure}[t]{0.48\textwidth}
        \includegraphics[width=\linewidth]{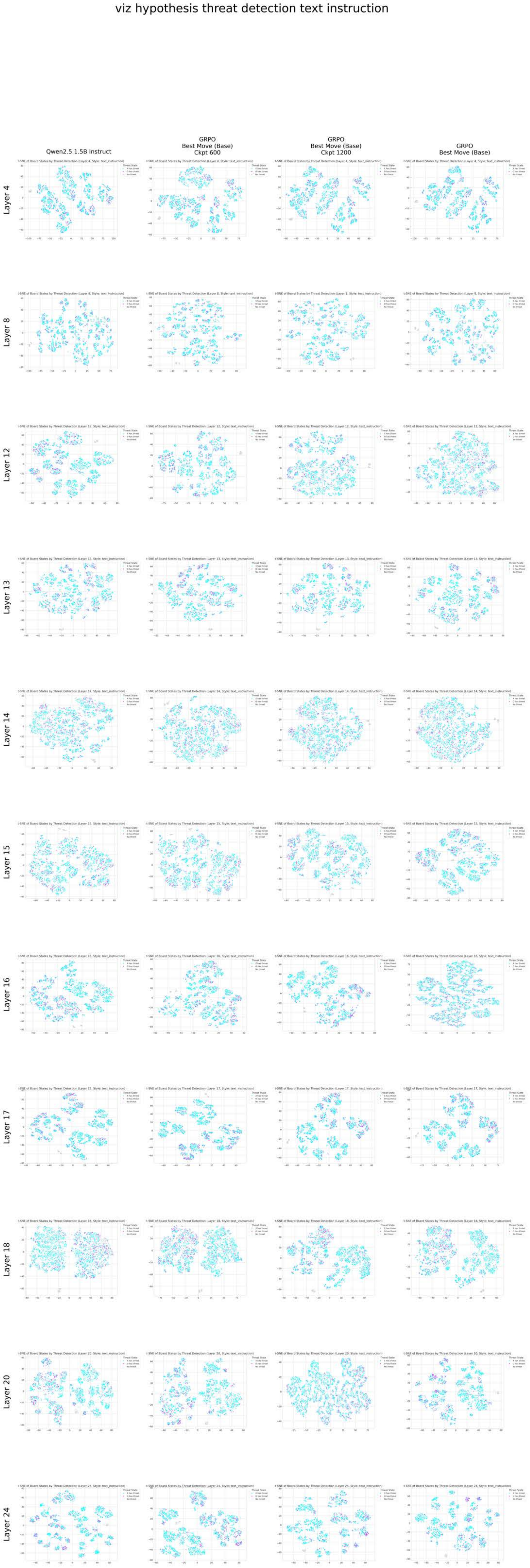}
        \caption{Natural language instruction representation}
    \end{subfigure}
    \caption{Hypothesis testing: Threat detection.}
    \label{fig:appendix-interp-threat}
\end{figure*}

% ------------------------------
% Turn-by-Turn Paired
% ------------------------------
\begin{figure*}[ht]
    \centering
    \begin{subfigure}[t]{0.48\textwidth}
        \includegraphics[width=\linewidth]{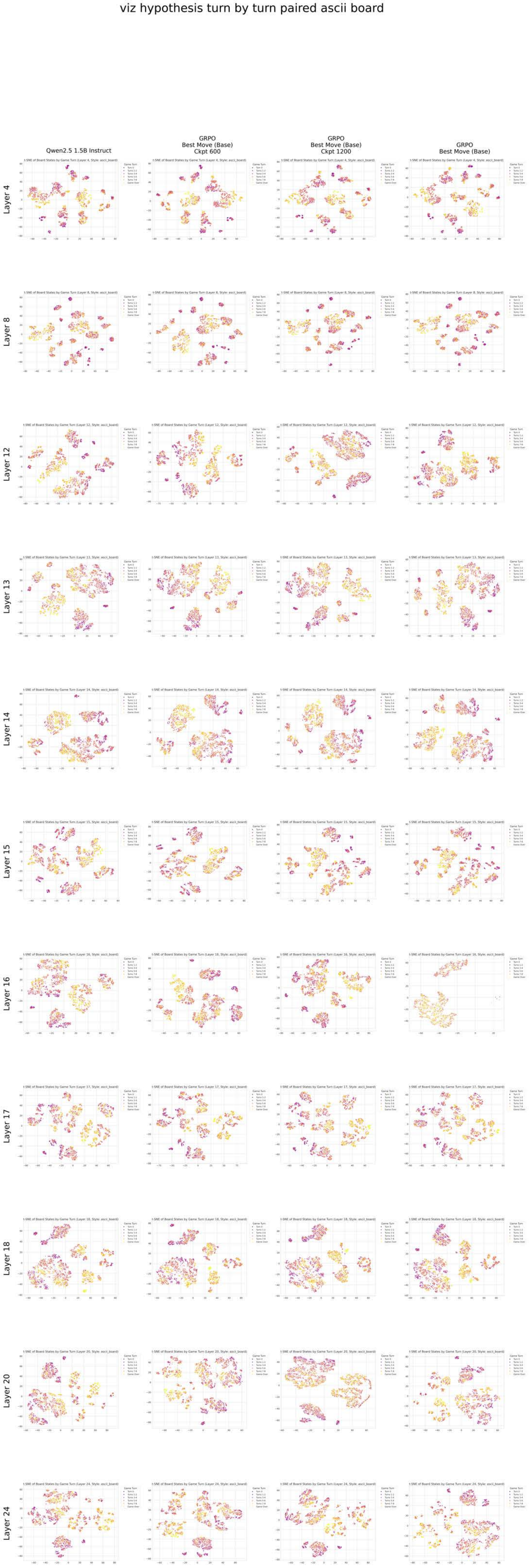}
        \caption{ASCII board representation}
    \end{subfigure}%
    \hfill
    \begin{subfigure}[t]{0.48\textwidth}
        \includegraphics[width=\linewidth]{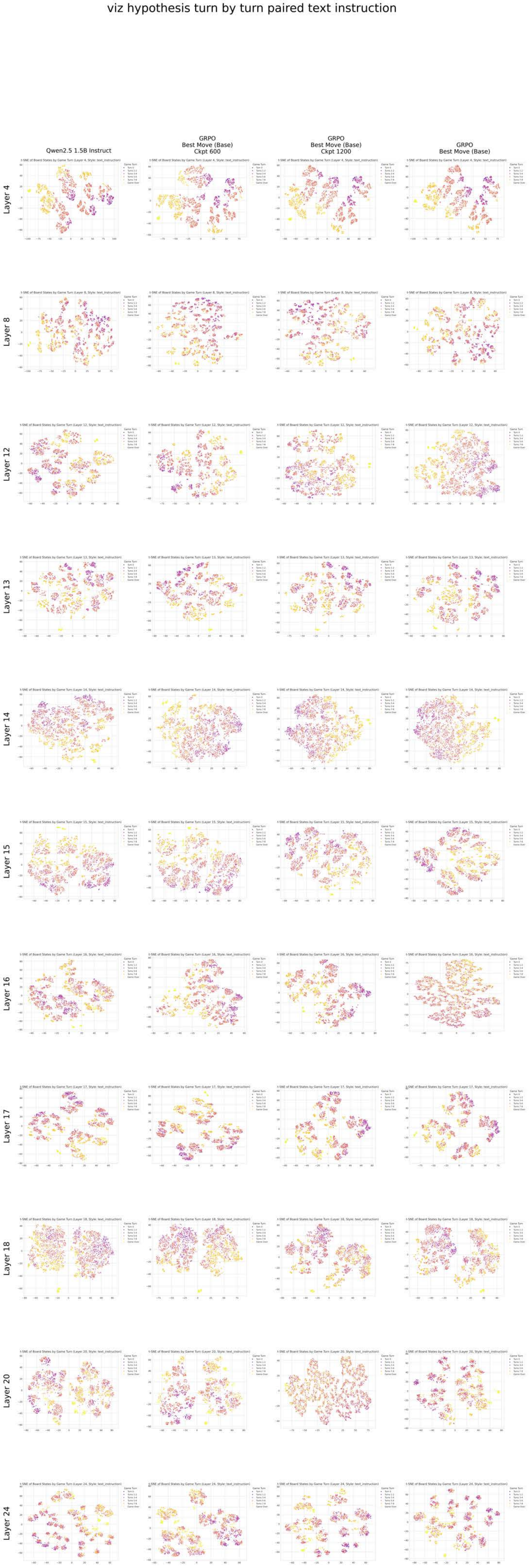}
        \caption{Natural language instruction representation}
    \end{subfigure}
    \caption{Hypothesis testing: Turn-by-turn paired analysis.}
    \label{fig:appendix-interp-game-turn-paired}
\end{figure*}

% ------------------------------
% Winner Analysis
% ------------------------------
\begin{figure*}[ht]
    \centering
    \begin{subfigure}[t]{0.48\textwidth}
        \includegraphics[width=\linewidth]{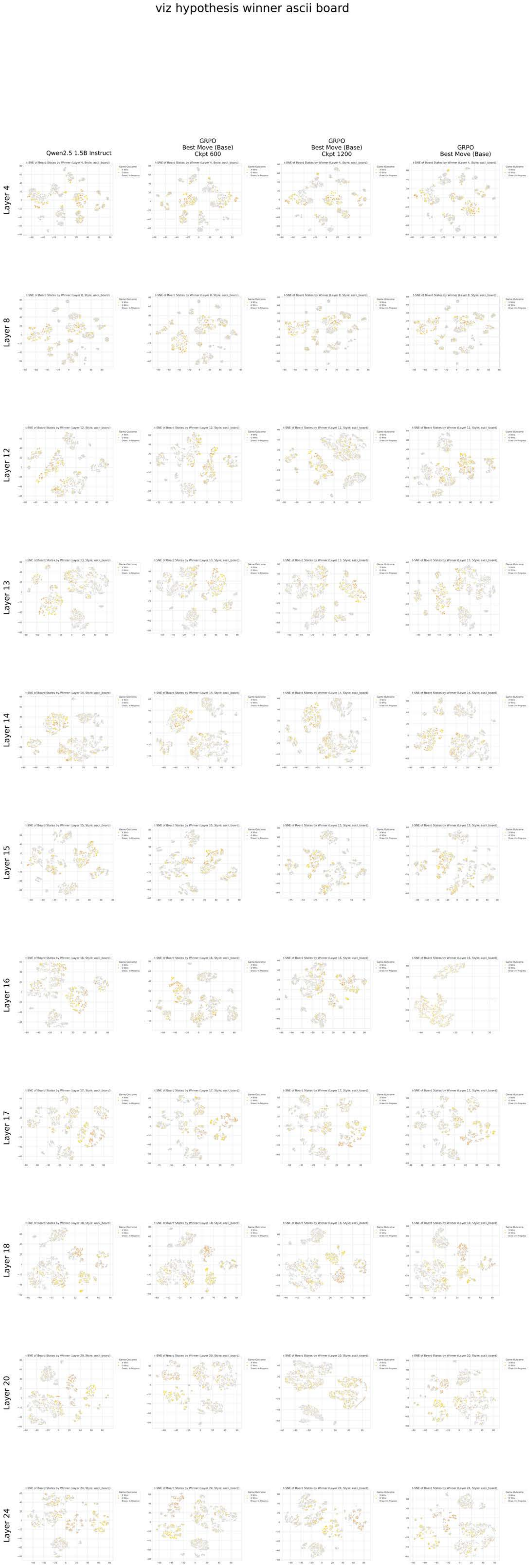}
        \caption{ASCII board representation}
    \end{subfigure}%
    \hfill
    \begin{subfigure}[t]{0.48\textwidth}
        \includegraphics[width=\linewidth]{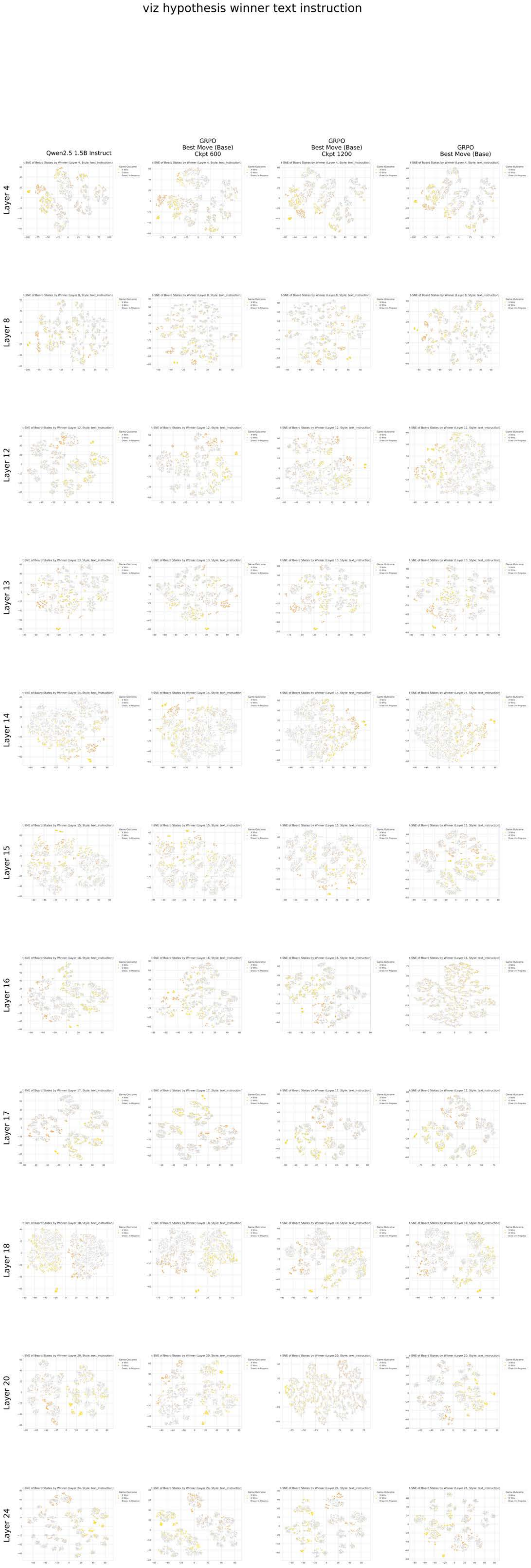}
        \caption{Natural language instruction representation}
    \end{subfigure}
    \caption{Hypothesis testing: Winner identification.}
    \label{fig:appendix-interp-winner}
\end{figure*}

% ------------------------------
% Winner (Best Labels)
% ------------------------------
\begin{figure*}[ht]
    \centering
    \begin{subfigure}[t]{0.48\textwidth}
        \includegraphics[width=\linewidth]{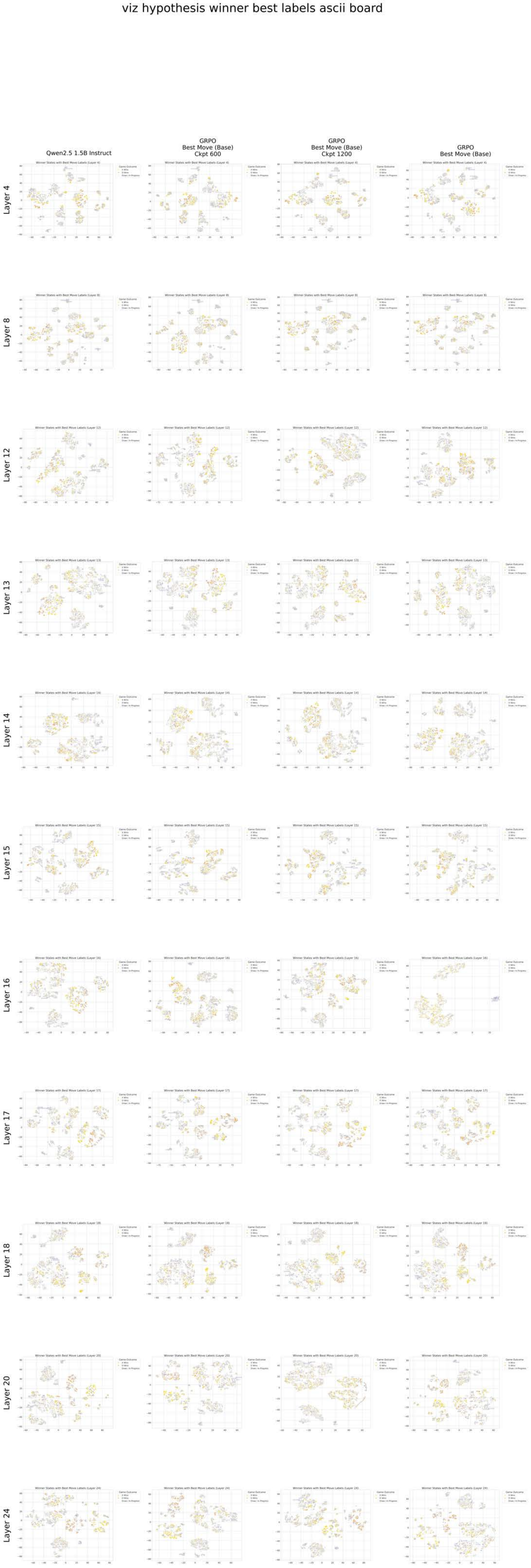}
        \caption{ASCII board representation}
    \end{subfigure}%
    \hfill
    \begin{subfigure}[t]{0.48\textwidth}
        \includegraphics[width=\linewidth]{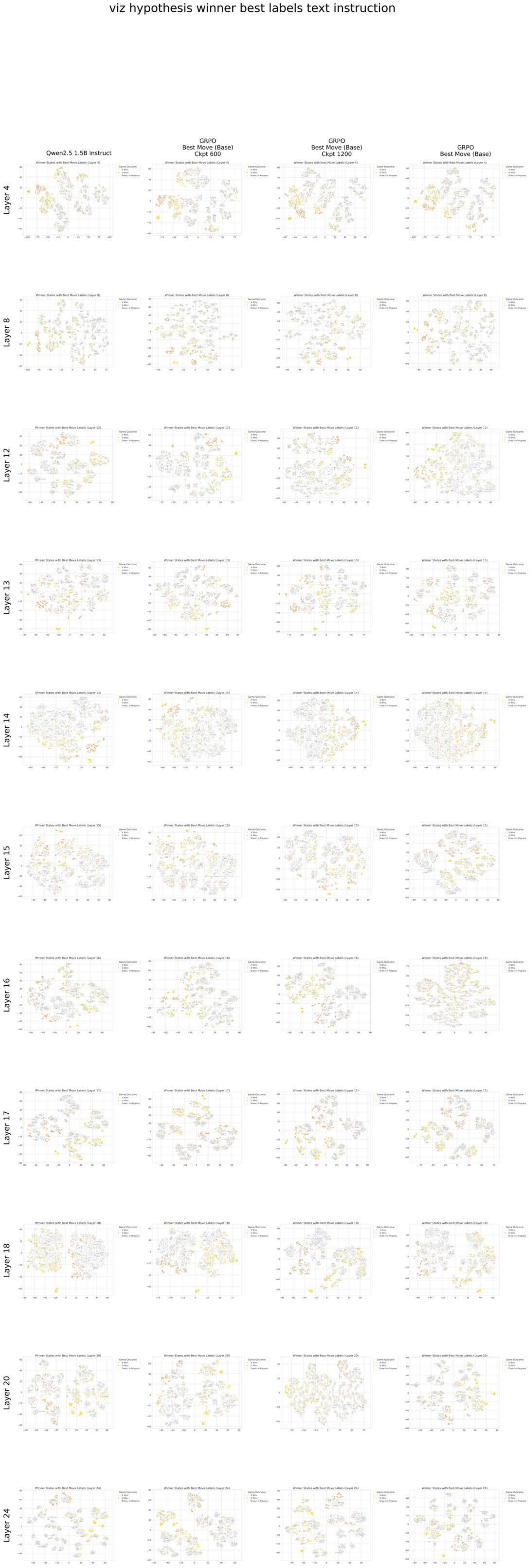}
        \caption{Natural language instruction representation}
    \end{subfigure}
    \caption{Hypothesis testing: Winner identification with best labels.}
    \label{fig:appendix-interp-winner-best-labels}
\end{figure*}

% ------------------------------
% KMeans: Dominant Line Patterns
% ------------------------------
\begin{figure*}[ht]
    \centering
    \begin{subfigure}[t]{0.48\textwidth}
        \includegraphics[width=\linewidth]{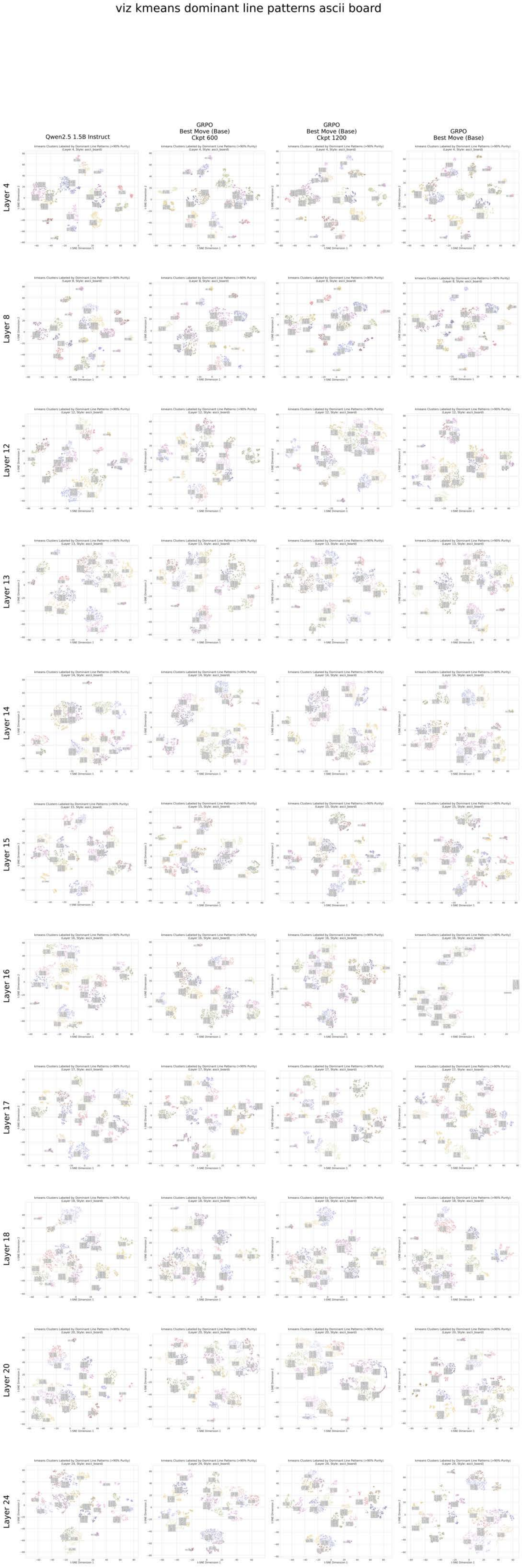}
        \caption{ASCII board representation}
    \end{subfigure}%
    \hfill
    \begin{subfigure}[t]{0.48\textwidth}
        \includegraphics[width=\linewidth]{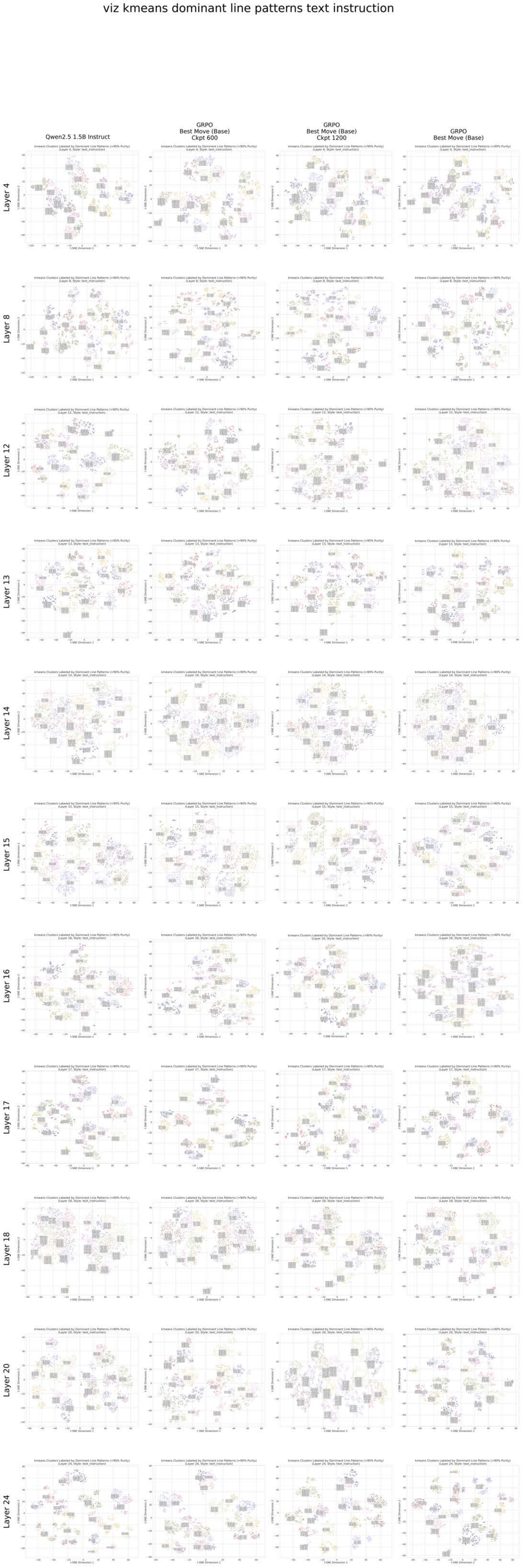}
        \caption{Natural language instruction representation}
    \end{subfigure}
    \caption{KMeans clustering: Dominant line patterns.}
    \label{fig:appendix-interp-dominant-line}
\end{figure*}

% ------------------------------
% KMeans: Global Extremes
% ------------------------------
\begin{figure*}[ht]
    \centering
    \begin{subfigure}[t]{0.48\textwidth}
        \includegraphics[width=\linewidth]{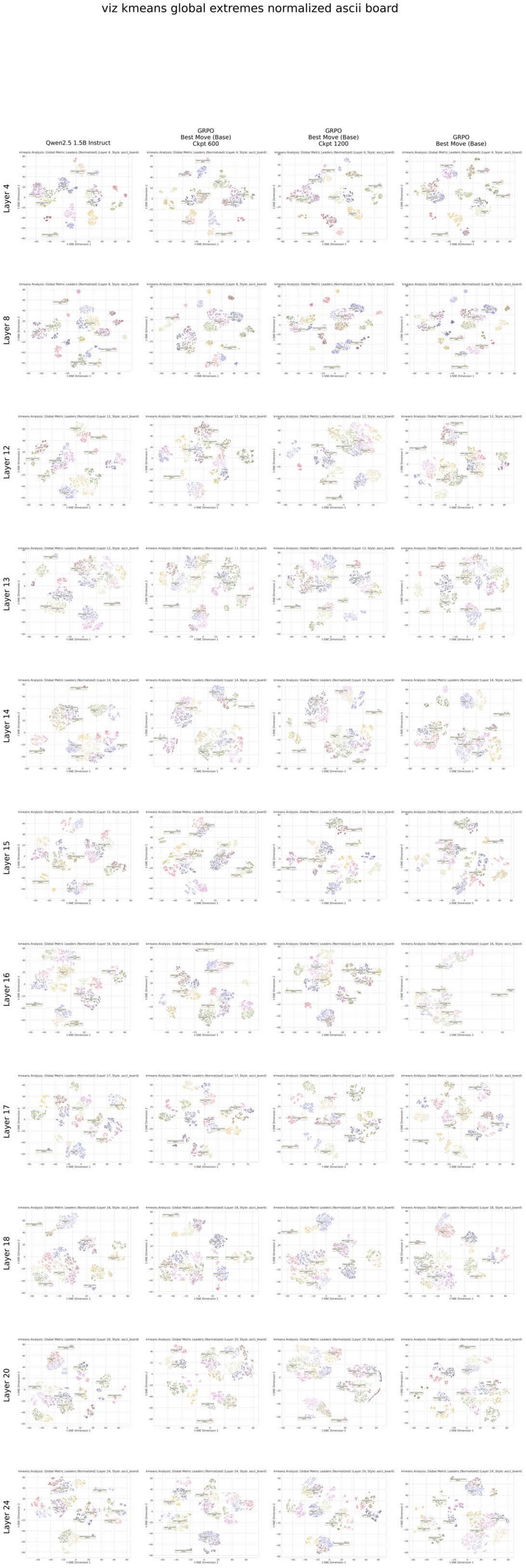}
        \caption{ASCII board representation}
    \end{subfigure}%
    \hfill
    \begin{subfigure}[t]{0.48\textwidth}
        \includegraphics[width=\linewidth]{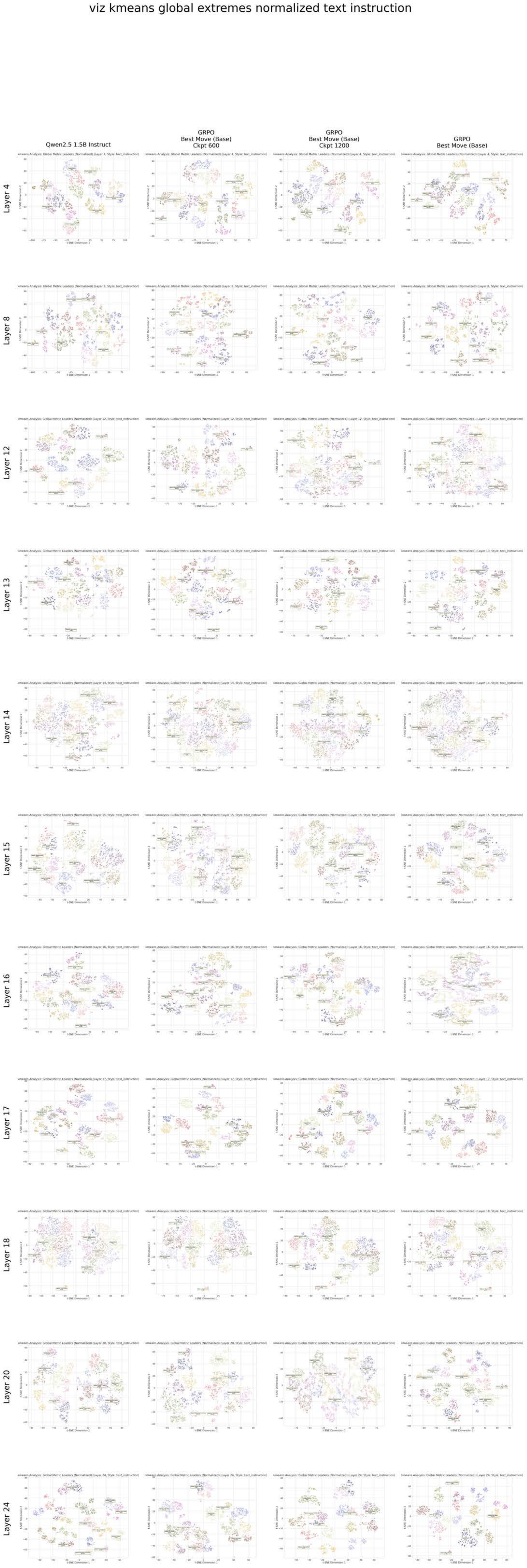}
        \caption{Natural language instruction representation}
    \end{subfigure}
    \caption{KMeans clustering: Global extremes (normalized).}
    \label{fig:appendix-interp-global-extremes}
\end{figure*}

% ------------------------------
% KMeans: Hypothesis Summary
% ------------------------------
\begin{figure*}[ht]
    \centering
    \begin{subfigure}[t]{0.48\textwidth}
        \includegraphics[width=\linewidth]{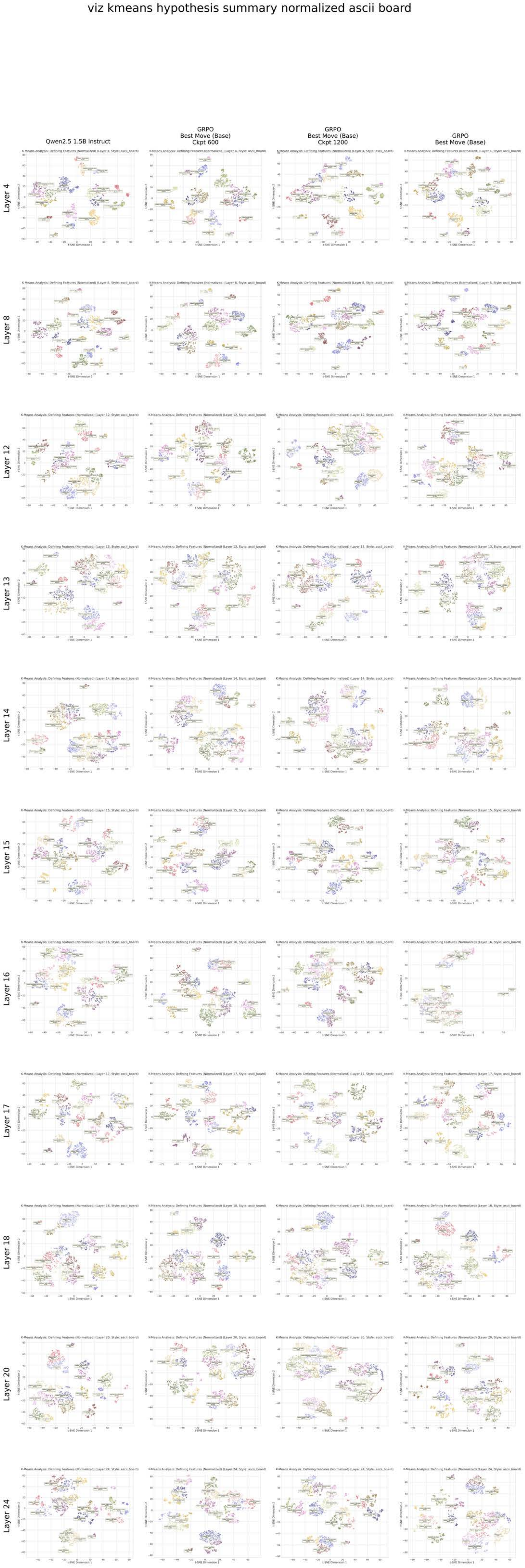}
        \caption{ASCII board representation}
    \end{subfigure}%
    \hfill
    \begin{subfigure}[t]{0.48\textwidth}
        \includegraphics[width=\linewidth]{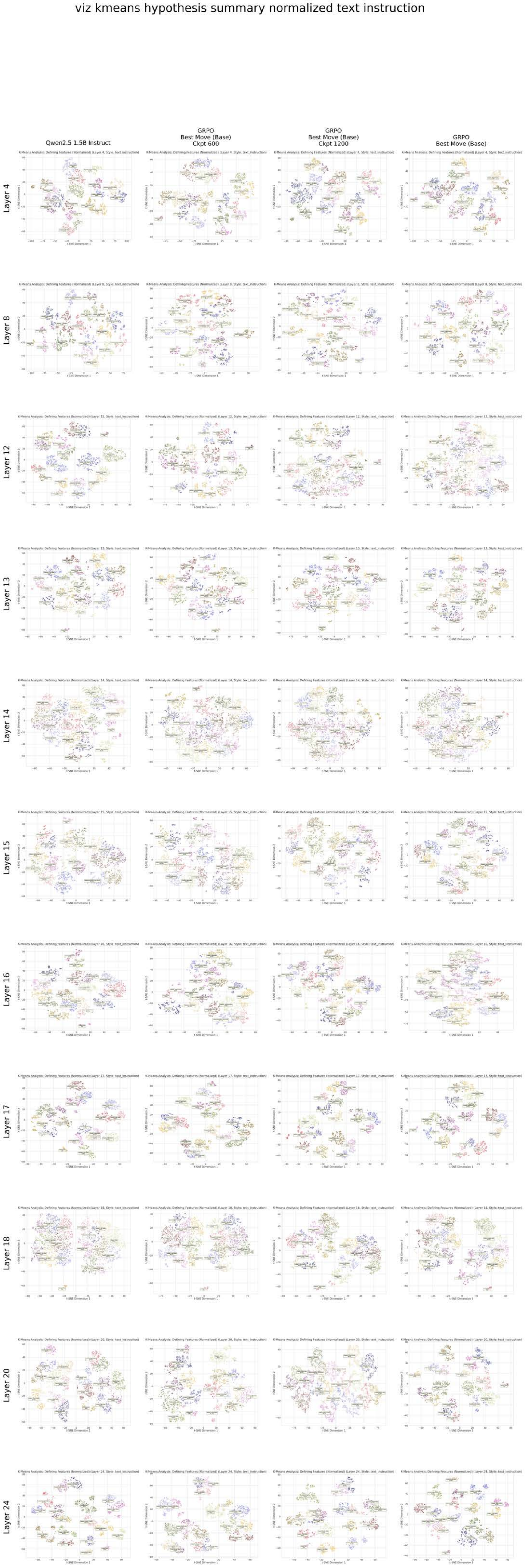}
        \caption{Natural language instruction representation}
    \end{subfigure}
    \caption{KMeans clustering: Hypothesis summary (normalized).}
    \label{fig:appendix-interp-hypothesis-summ}
\end{figure*}

\clearpage

\section{Illegal boards: an out-of-distribution generalization test}
\label{sec:illegal-appendix}

\rev{The unreachable boards test whether the model internalised the rules that generate valid states. Models were never trained to flag illegality, so proposing legal continuations on unreachable boards is expected under the objective, and we present this result as illustrative rather than as evidence about constraint encoding. The informative result is a dissociation between representation and policy: probes separate legal from unreachable boards at AUC $1.00$ at every layer (App.~\ref{sec:probe-appendix}), partly reflecting surface count differences, so the distinguishing information is present in the representations even though the policy never acts on it.} In the SAE geometry, illegal boards sit near legal clusters matched by simple line patterns rather than by legality constraints (details in Appendix Figs.~\ref{fig:appendix-heatmaps}, \ref{fig:appendix-interp-illeval-vs-legal}). \rev{The representation thus generalises by line-pattern similarity}, a distinction also emphasized in recent small-model reasoning studies \citep{dang2025reinforcement,srivastava2025towards,shin2025can}.

\section{Input modality and local cluster structure}
\label{sec:local-structure-appendix}

\paragraph{Input structure matters: ASCII vs.\ natural language.}
The quantitative drop under ASCII boards is reflected mechanistically: SAE features for \emph{tokenized NL boards} contain many crisp detectors aligned to lexical markers (\textit{``Row 1'', ``empty''}) and ordinal positions, cleanly supporting \textbf{H1/H2} as seen in Figures~\ref{fig:interp-clusters} and~\ref{fig:interp-comparison} (middle row). 
% When switched to ASCII, activations rotate; turn/progression axes remain but are noisier, and tactical micro-islands thin out. 
The base-model game-turn macro pattern visible under NL inputs is absent under ASCII (App.~Fig.~\ref{fig:appendix-interp-game-turn}). 
This suggests that reasoning training does not make the models robust to true ``reasoning''.
% The model does not develop true input-agnostic general reasoning for the task of playing Tic-Tac-Toe, but rather updates its internal representations to adapt to the input patterns which can help it achieve the highest reward for the training setting. 
This asymmetry aligns with context-dependence findings \citep{hua2024disentangling} and data-distribution sensitivity \citep{zhao2026chain}, and helps explain why random symbol substitution hurts slightly while ASCII hurts substantially in the overall plots.

\paragraph{Local clusters in model representations.}
\label{subsec:compositionality}
We define line purity of a subcluster as the largest fraction of its boards sharing a row, column, or diagonal pattern, and report subclusters with purity $\geq 0.7$. K-means followed by agglomerative refinement over SAE-PCA space recovers crisp local pockets capturing patterns like ``late-game, exactly one empty in R0'' or ``mid-game R0:\ \texttt{X . .}, center taken'' (Fig.~\ref{fig:interp-clusters}). These structures are present in the base model and retained after training, while the macroscopic arrangement evolves to add game progression. 
Symmetry-class membership is absent at the cluster level: the canonical-symmetry ratio falls below $0.7$ in only $4$ of $792$ clusters, all of them end-game clusters where progression rather than symmetry drives membership (App.~Fig.~\ref{fig:appendix-interp-symmetry}).

\clearpage

\begin{revblock}

\section{Full-dimensional probe analysis}
\label{sec:probe-appendix}

The SAE pipeline in the main paper reads concepts out through a sparse dictionary, a t-SNE projection and clustering. To confirm that our conclusions are not artifacts of that pipeline, we repeat the hypothesis tests directly on the model's raw activations. For each hypothesis H1--H7 we train classifiers to predict the corresponding board property from the raw residual-stream activations at the last prompt token, with no SAE, no t-SNE and no clustering anywhere in the path. We use logistic regression and linear discriminant analysis (LDA) as linear probes, together with a one-hidden-layer MLP as a non-linear reference. All probes use five-fold stratified cross-validation and are scored by macro one-versus-rest AUC, and significance is assessed against label-permutation null distributions. Every number is computed on held-out folds.

\begin{table}[h]
\centering\small\revcolor
\begin{tabular}{lcc}
\toprule
Hypothesis & Base model & Final checkpoint \\
\midrule
H1 turn identity & $1.000 \pm 0.000$ & $1.000 \pm 0.000$ \\
H2 game progression & $1.000 \pm 0.000$ & $1.000 \pm 0.000$ \\
H3 strategic situation & $0.686 \pm 0.011$ & $0.684 \pm 0.012$ \\
H7 line purity & $0.861 \pm 0.014$ & $0.856 \pm 0.011$ \\
H5 legality (legal vs.\ unreachable) & $1.000 \pm 0.000$ & $1.000 \pm 0.000$ \\
H4 same-symmetry-group similarity (pair test) & 0.70 & 0.70 \\
\bottomrule
\end{tabular}
\caption{Linear-probe results at the middle layer of Qwen2.5-1.5B: cross-validated AUC $\pm$ standard deviation across folds, for the base model and the final GRPO checkpoint. For H3, MLP probes reach 0.75 (base) and 0.73 (final checkpoint), so at most 0.05 to 0.06 of additional AUC is recoverable non-linearly. H4 uses the pair-separation test described below.}
\label{tab:probe-main}
\end{table}

Table~\ref{tab:probe-main} compares the base model with the final GRPO checkpoint. Prompt-level properties (H1 turn identity, H2 game progression) and H5 legality are perfectly decodable in both. The higher-order properties are the ones with headroom, and they do not move: H3 strategic situation stays at 0.686 (base) against 0.684 (final checkpoint) with linear probes, and the MLP raises it only to 0.75 and 0.73. H7 line purity is high and likewise stable. H6 best-move correctness is measured in the same form and is reported with the probe grid in the supplementary material.

For H4, the 765 symmetry classes make direct classification impractical. We instead test whether activations of symmetry-related boards are more similar than activations of unrelated boards matched on game stage, summarised as a pair-separation AUC. The resulting 0.70 partly reflects shared piece counts and line patterns, and the informative fact is that it does not grow across training.

\paragraph{Cross-model replication.}
Table~\ref{tab:probe-crossmodel} reports logistic-regression AUC ranges across checkpoints at a mid-network layer for all three models. H1 is at 1.00 at every checkpoint of every model, and H3 stays flat under GRPO in all three, so the decodability findings are not specific to Qwen2.5-1.5B.

\begin{table}[h]
\centering\small\revcolor
\begin{tabular}{lcc}
\toprule
Model & H1 turn identity & H3 strategic situation \\
\midrule
Qwen2.5-1.5B & 1.00 at every checkpoint & 0.68 to 0.70, flat \\
Qwen2.5-0.5B & 1.00 at every checkpoint & 0.61 to 0.65, flat \\
Llama-3.2-1B & 1.00 at every checkpoint & 0.65 to 0.70, flat \\
\bottomrule
\end{tabular}
\caption{Logistic-regression AUC ranges across GRPO checkpoints at a mid-network layer. ``Flat'' means the range shows no upward trend under GRPO training. The dominant-axis reorganisation itself is clearest in Qwen2.5-1.5B and noisier in the smaller models, so we scope that claim to Qwen2.5-1.5B.}
\label{tab:probe-crossmodel}
\end{table}

\paragraph{Permutation nulls.}
Table~\ref{tab:probe-permutation} reports permutation tests with 999 label permutations per comparison, using the LDA probe at the middle layer of Qwen2.5-1.5B on a held-out split. Every observed value exceeds the maximum of its 999 shuffled nulls, which attains the smallest $p$ available at this resolution, and all comparisons remain significant after Benjamini-Hochberg correction at any conventional threshold. The nulls also address probe capacity directly: if this probe width could fit its way to high AUC, the permuted labels would show it, and they do not (null mean 0.50, null maximum 0.61).

\begin{table}[h]
\centering\small\revcolor
\begin{tabular}{lcccc}
\toprule
Hypothesis & Observed (base) & Observed (final) & Largest null & $p$ \\
\midrule
H1 turn identity & 1.000 & 1.000 & 0.57 & 0.001 \\
H2 game progression & 0.996 & 0.994 & 0.58 & 0.001 \\
H3 strategic situation & 0.650 & 0.651 & 0.56 & 0.001 \\
H7 line purity & 0.834 & 0.818 & 0.61 & 0.001 \\
H5 legality & 1.000 & 1.000 & 0.56 & 0.001 \\
\bottomrule
\end{tabular}
\caption{Permutation tests (999 label permutations per comparison, LDA probe, middle layer, Qwen2.5-1.5B, held-out split): observed AUC for the base model and the final checkpoint, against the largest value in the shuffled null distribution. Even H3, the weakest hypothesis, sits far above its null ceiling, so the finding is that strategic information is limited and static under GRPO, not that it is absent.}
\label{tab:probe-permutation}
\end{table}

\paragraph{Dominant-axis statistic without t-SNE.}
As a projection-free counterpart to the t-SNE ANOVA of Table~\ref{tab:cluster-stats-combined}, we compute the variance explained by class means along the first principal component of the raw activations. This statistic makes no separability claim. It measures how prominently a property organises the single largest axis of variation, which is the quantity behind the inversion reported in the main paper, and unlike a t-SNE embedding the variance a principal component captures and discards is exactly measurable. At the middle layer the turn share falls from 0.151 (base) to 0.062 (final checkpoint) while the progression share holds near 0.20, reproducing the direction of the inversion with no t-SNE in the path.

\paragraph{Interpretation.}
Prompt-level properties are perfectly readable before RL starts, so the amplification claim concerns how prominently information is organised and how heavily it is used, not whether it exists. The properties with headroom are the higher-order ones, and RL leaves them unchanged. Turn information never becomes less decodable. It moves off the dominant variance axis mid-network while remaining perfectly present, which refines the inversion in Table~\ref{tab:cluster-stats-combined}: what changes under GRPO is which features dominate the representation, not which concepts are recoverable.

\clearpage

\section{Causal interventions: activation patching and steering}
\label{sec:causal-appendix}

\paragraph{Activation patching.}
We construct 100 minimal board pairs. The two boards in a pair are identical except that one opponent piece is relocated, the same player is to move, and each board has a unique optimal move, with the two optimal moves differing. We call the board whose prompt is run the \emph{receiver} and the board supplying activations the \emph{donor}. The design instantiates Algorithm~\ref{alg:intervention} using standard interchange-intervention and causal-tracing methodology (Vig et al., 2020; Geiger et al., 2021), localising the causal effect by layer as in Meng et al. (2022), in the paired-input form of Wang et al. (2023). At a single layer we overwrite the residual-stream activations at the answer position of the receiver's forward pass with the donor's activations. Everything else the model processes is the receiver, and the question is which board's optimal move it now emits. The controls follow the activation-patching recommendations of Zhang and Nanda (2024).

Unpatched, the model plays the receiver-optimal move on 72 of 100 pairs and the donor-optimal move on 10 of 100, which is its ordinary error rate. Patched at layer 25 of 28, it plays the donor-optimal move on 66 of 100 pairs. Two controls rule out trivial explanations. The same patch at layers 13, 17 and 21 leaves behaviour unchanged, so arbitrary tampering does not redirect the answer. Injecting activations from a random unrelated board instead of the paired donor transfers that board's move in only 12 of 100 cases, so the effect is specific to the donor's board content rather than generic disruption. The difference of 66 against 12 of 100 is significant at $p < 10^{-14}$ by a two-proportion test.

\paragraph{Steering.}
A complementary experiment probes the same question from the other side, using difference-in-means activation addition (Turner et al., 2023; Panickssery et al., 2024). The steering direction connects the average internal state of X-to-move boards to that of O-to-move boards, and multiples of it are added to the activations during generation at each of six layers. The output never flips between the X and O move-token ranges, at any layer or magnitude, while norm-matched random directions at extreme magnitudes do disrupt outputs, which establishes that the intervention reaches the computation. The reorganised mid-layer turn axis therefore does not by itself drive the move, independently confirming what the layer-resolved patching shows.

\paragraph{Interpretation.}
Both interventions place the decision in the same location. Board information becomes causally decisive where it consolidates in late layers, and the mid-layer axes that GRPO reorganises are not individually sufficient to drive behaviour. This is the causal counterpart of the representational result: what training reshapes mid-network is not where the answer is produced, which is exactly the output-decoding interpretation of the late-layer re-emergence of turn structure in Table~\ref{tab:cluster-stats-combined}.

\clearpage

\section{Full-game arena evaluation}
\label{sec:arena-appendix}

To connect per-state accuracy to multi-turn play, models play complete games from the empty board, alternating moves until a win, a draw or a forfeit, where an illegal or unparseable move ends the game in the opponent's favour. The field mixes frontier API models with the GRPO-trained and untrained models, and every move is also scored against the minimax optimum (Algorithm~\ref{alg:minimax}).

\begin{table}[h]
\centering\small\revcolor
\begin{tabular}{lcc}
\toprule
 & GRPO-trained & Untrained \\
\midrule
Elo & 1525 & 1287 \\
Games & 154 & 160 \\
Own forfeits & 0 & 121 \\
Per-move minimax-optimality (completed games) & 77\% & not measurable \\
Draws against the flash-tier frontier model & 16 of 16 & 0 \\
\bottomrule
\end{tabular}
\caption{Arena results for GRPO-trained against untrained Qwen2.5-1.5B. Per-move minimax-optimality is not measurable for the untrained model because it forfeits most of its games before completion. Elo is order-dependent and X always moves first, so the comparison between the two rows rather than the absolute rating is the informative quantity.}
\label{tab:arena}
\end{table}

Table~\ref{tab:arena} compares the two. Training eliminates forfeits (0 against 121), yields 77 percent per-move minimax-optimality in completed games, and turns every game against the flash-tier model into a draw. The trained model's 84 losses were all genuine on-board losses against stronger frontier models, not forfeits. Training on the single-move objective therefore transfers directly to sequential play, as the argument in Sec.~\ref{sec:method} predicts.

\section{Behavioural equivariance and memorisation}
\label{sec:equivariance-appendix}

\paragraph{Behavioural equivariance over the full state space.}
Two boards related by a rotation or reflection are the same position in different orientations, so a policy that uses symmetry should choose corresponding moves on both. For every pair of symmetry-related non-terminal boards over all 5,478 states, we test whether the predicted moves correspond under the transformation relating the two boards. On pairs where each board has a unique optimal move, a symmetry-respecting policy must score 100 percent, so the unique-optimum column of Table~\ref{tab:equivariance} has an exact ceiling. Scores sit far below that ceiling on every subset, and held-out symmetry groups score no lower than training symmetry groups, so the policy does not use symmetry and this finding is not an artifact of the split design.

\begin{table}[h]
\centering\small\revcolor
\begin{tabular}{llcccc}
\toprule
Split & Symmetry groups & Pairs & Equivariance (\%) & Unique-optimum (\%) & $n$ \\
\midrule
Canonical & Training & 1,484 & 56.7 (54.2--59.2) & 60.0 (56.7--63.2) & 857 \\
Canonical & Held-out & 653 & 57.7 (53.9--61.5) & 62.7 (57.6--67.5) & 359 \\
Random & All & 2,119 & 56.1 (54.0--58.2) & 60.5 (57.8--63.3) & 1,214 \\
\bottomrule
\end{tabular}
\caption{Behavioural equivariance of Qwen2.5-1.5B: percentage of symmetry-related board pairs whose predicted moves correspond under the relating transformation, with 95 percent Wilson intervals in parentheses. ``Unique-optimum'' restricts to the $n$ pairs where each board has a unique optimal move, on which a symmetry-respecting policy must score 100 percent.}
\label{tab:equivariance}
\end{table}

\paragraph{Memorisation check.}
Table~\ref{tab:memorisation} compares best-move accuracy on training states with accuracy on held-out states. Held-out accuracy matches training accuracy within the intervals for every model and split, which argues against verbatim state memorisation as the driver of performance.

\begin{table}[h]
\centering\small\revcolor
\begin{tabular}{llcc}
\toprule
Model & Split & Training states & Held-out states \\
\midrule
Qwen2.5-1.5B & Random & 78.7 (77.5--79.9) & 77.1 (74.5--79.5) \\
Qwen2.5-1.5B & Canonical & 72.9 (71.4--74.2) & 71.9 (69.7--74.0) \\
Qwen2.5-0.5B & Random & 43.0 (41.5--44.5) & 43.4 (40.5--46.4) \\
Qwen2.5-0.5B & Canonical & 36.5 (35.0--38.1) & 36.3 (34.0--38.6) \\
\bottomrule
\end{tabular}
\caption{Best-move accuracy (\%) on training against held-out states, with 95 percent Wilson intervals in parentheses.}
\label{tab:memorisation}
\end{table}

\begin{table}[h]
\centering\small
\rev{
\begin{tabular}{lr}
\toprule
Statistic & Count \\
\midrule
Possible board assignments ($3^9$) & 19,683 \\
Reachable legal states & 5,478 \\
\quad Terminal / non-terminal & 958 / 4,520 \\
Symmetry groups & 765 \\
\quad Of size $1/2/4/8$ & $6/6/141/612$ \\
Random split (train/val/test) & 4,382/547/549 \\
Canonical-symmetry split & 3,825/830/823 \\
Illegal boards & 5,000 \\
\bottomrule
\end{tabular}}
\caption{\rev{Dataset statistics. The canonical-symmetry split is assigned at the symmetry-group level, and the illegal boards are balanced across violation categories.}}
\label{tab:dataset-stats}
\end{table}

\paragraph{Why the canonical split is necessary.}
Under the random split, every one of the 549 test states has at least one symmetric sibling in training. For symmetry groups of size eight, the probability that a test state has no sibling in an 80 percent training sample is about 1 in 78,000. Holding out entire symmetry groups is therefore the only way to obtain fully unseen symmetry groups at scale, which is why the two splits are separate designs in Sec.~\ref{sec:method} rather than two arms of one comparison.

\end{revblock}

\clearpage

\section{Extended limitations and discussion}
\label{sec:extended-limitations}

This appendix expands the limitations called out in the main paper and records additional caveats.

\paragraph{Memorisation vs abstraction in a small domain.} With only 5{,}478 reachable states and 80\% of them in training, memorisation and abstraction are hard to disambiguate in principle. Our claim is more specific. Strategic abstractions (forks, threats, winning sequences) do not emerge in the mid-layer feature space, while an easier-to-template abstraction (game progression) does. The asymmetry between H2 and H3 in Table~\ref{tab:cluster-stats-combined} is the most direct evidence against an alternative interpretation in which the model has learned competent pattern-completion across the board. Under that interpretation, both H2 and H3 should strengthen, however only H2 does. Four further pieces of convergent evidence support this: Symmetry-controlled curves track random-split curves rather than separating, which a template-memoriser would not produce. \rev{Chance-corrected skill on held-out non-terminal states is substantial in every branching bin, and accuracy on held-out states matches accuracy on training states within the intervals for every model and split (App.~\ref{sec:equivariance-appendix}), neither of which a memoriser of an 80\% training set would produce.} The ASCII modality drop (73\% to 44\% on Qwen2.5-1.5B) shows the representation is tied to natural-language token boundaries rather than to abstract board state. Illegal boards elicit legal continuations rather than flags. Each piece is weak alone. Together they argue against the strongest interpretation in which memorisation and abstraction reduce to the same thing at this scale.

\paragraph{Why strategic abstractions stay weak.}
Table~\ref{tab:cluster-stats-combined} shows H2 growing by $\sim$$8.5\times$, H3 by only $\sim$$2\times$, and H4 staying at noise. Two complementary explanations cover this asymmetry.

\textbf{(i) Reward shape.} GRPO rewards the correctness of the emitted move. It does not reward faithfulness of any internal computation. When prompt-level cues already determine the answer, such as forced wins, low-branching positions, and late game, there is no gradient pressure to internalise a search-based abstraction. \rev{Near-ceiling terminal-state detection from the first checkpoint, alongside only gradual chance-corrected gains on non-terminal states, is the signature of a policy that has learned to exploit cues without learning to plan.}

\textbf{(ii) Architectural constraints.} Autoregressive transformers commit a fixed amount of compute per generated token. This biases the optimisation toward extractive features rather than search-heavy ones~\citep{venhoff2025understanding,cosentino2024reasoning}. They also lack any built-in equivariance to the dihedral $D_4$ group acting on the board, so symmetry-aware structure must be learned end-to-end from data. The flat H4 row of Table~\ref{tab:cluster-stats-combined} is consistent with this: there is neither an architectural prior pushing the model toward symmetry-aware features nor a training signal rewarding them, and at no probed layer does symmetry-class membership exceed noise. Architectures that allocate variable compute have, in principle, more room to develop reusable strategic circuits. Examples include energy-based transformers~\citep{gladstone2025energy}, mixture-of-recursions~\citep{bae2025mor}, soft-reasoning embedding exploration~\citep{zhu2025soft}, and diffusion language models~\citep{nie2025diffusion}. We see this as a productive target for follow-up mechanistic studies.

% We note one caveat. SAEs may fail to expose strategic concepts that are encoded non-linearly or distributed across layers. We acknowledge this in the limitations. Causal interventions on SAE features (Algorithm~\ref{alg:intervention}) are the natural next step. They would distinguish ``the feature is missing'' from ``our probe missed it''.

\paragraph{Domain simplicity is a feature, not a bug.}
We chose Tic-Tac-Toe for its closed structure and exhaustive solvability. The trade-off is that the domain does not capture open-ended or probabilistic reasoning. For the present study this is a feature, not a bug. Representational dynamics can be measured against ground truth. Negative findings about strategic abstraction are unambiguous. Extending the framework to incrementally larger logical domains (Connect Four, simple algebra, grid-world planning) is an obvious next step.

\paragraph{Reward composition.}
The GRPO reward we use combines legality, format compliance, and correctness with equal weight. The structured-output baseline (Table~\ref{tab:reward-full}) rules out the format component as a primary driver. A fuller decomposition (Table~\ref{tab:reward-full}) shows that GRPO improvements hold across NL/ASCII inputs and random symbol substitution. It also shows that warm-starting best-move training from a legal-move checkpoint adds a consistent uplift over training from base. A complete reward-component ablation, for example dropping the format term, would clarify which aspects of the optimisation drive each effect. We leave this to a follow-up.

\paragraph{Evaluation scope.}
Our metrics are per-state best-move accuracy under controlled prompt perturbations (NL/ASCII, random symbol substitution, illegal boards). Tic-Tac-Toe is solved, so per-state optimality with respect to the minimax oracle (Algorithm~\ref{alg:minimax}) is equivalent to perfect play against an arbitrary opponent. The ``best move'' label encodes the entire continuation. We therefore do not separately report vs-MCTS curves in the main paper. \rev{Full multi-turn game evaluation, in an arena mixing the trained, untrained and frontier models, is reported in App.~\ref{sec:arena-appendix}.}

\paragraph{Generalisation and abstraction.}
The framework isolates structural reasoning from prompt-level adaptation. It does not exhaust all possible forms of abstraction. The absence of high-level reasoning signals here does not imply that such capabilities cannot emerge under different training, reward, or architecture choices. It indicates that under the GRPO and small-transformer setting we study, the optimisation favours strengthening existing prompt-level features over conceptual reorganisation.

\section{Behavioural rollouts}
\label{sec:rollouts}

We include one success and one failure rollout from Qwen2.5-1.5B (best-move checkpoint, NL prompt). Rollouts make concrete the pattern in Table~\ref{tab:per-position}. The model succeeds when the answer is extractable from the prompt. It fails confidently when mid-game lookahead is required.

\paragraph{Success: low-branching forced position.}
\textit{Board:} Row 0: empty, O, X. Row 1: X, empty, X. Row 2: O, empty, O. Player 1 to move.
The position has three legal moves and a unique optimal answer. The model emits a brief rationale that mentions cell $(1,1)$ as the relevant square and outputs the correct token~\texttt{5}. The verbalised reasoning is partly wrong. It claims a win is achievable along $(0,0)$, $(1,1)$, $(2,2)$, which is not the case from this state. The answer token is still optimal. This is consistent with prior reports that explanations and decisions can decouple~\citep{stechly2025self}.

\paragraph{Failure: terminal state misread as open.}
\textit{Board:} Row 0: O, empty, empty. Row 1: X, X, O. Row 2: empty, X, empty. Player 2 to move.
The position has three legal moves. The model verbally asserts that ``the board is currently full'' and that any of the $18$ tokens is acceptable. It eventually emits~\texttt{16}. It misreads occupancy despite the explicit board-state listing in the prompt. This failure mode appears most often on mid-game positions. It explains a substantial share of the unique-best-move accuracy gap in Table~\ref{tab:per-position}.

These rollouts illustrate the central thesis. The model has internalised which prompt features predict the answer. It has not internalised how to verify a board state and search for a good move. Mid-game positions, where prompt features under-determine the answer, are exactly where this strategy breaks down. A curriculum that front-loads such positions would force the gradient off pure extraction and into verification.

\begin{revblock}
\subsection{Reasoning traces show verbalised look-ahead}
\label{sec:reasoning-traces}

The chain-of-thought traces also contain explicit multi-step plans, which we exhibit qualitatively here, since a common concern is that a single-move objective gives the model no room to plan.

\textit{Board:} Row 0: O, O, X. Row 1: X, X, empty. Row 2: empty, O, X. Player 2 to move.
On this board the strongest checkpoint writes an explicit conditional plan: ``If Player 2 plays in (1,2), Player 1 will have to play in (1,1) \dots\ to block the win \dots\ Therefore, Player 2 should play in (1,2) to force a win''. The chosen move is minimax-optimal. In the same trace, however, the verbalised board description misstates the piece counts and the row contents.

\textit{Board:} Row 0: empty, X, empty. Row 1: empty, X, empty. Row 2: empty, O, empty. A second example shows the same dissociation. The trace claims two O pieces and a single empty square, both false, and the chosen move is still optimal.

Multi-step planning is thus explicitly attempted in the language substrate, with both the incentive and the room to develop, while the verbalised state is often ungrounded. This is consistent with prompt-level feature use rather than grounded internal search, and it is why the mechanistic readout, rather than the trace, is what settles the question.
\end{revblock}

\clearpage

\section{Cluster-vs-hypothesis statistics: full sweep}
\label{sec:cluster-stats}

The main paper reports as its primary statistic the $F$-statistic on the principal t-SNE axis (Table~\ref{tab:cluster-stats-combined}). Here we add a snapshot of three complementary cluster-vs-hypothesis metrics at layer 12 (Table~\ref{tab:cluster-main-snapshot}).

We use t-SNE coordinates from the cached pipeline (Algorithm~\ref{alg:tsne}) and run $K$-means with $k\!=\!18$ on the 2D projections. We then compute three statistics for each hypothesis: the one-way ANOVA $F$ on the principal t-SNE axis (does the dominant axis separate the hypothesis classes?), Cram\'er's $V$ on the cluster-vs-hypothesis contingency table (overall strength of association), and adjusted mutual information (AMI, chance-corrected). All three agree on the same H1$\rightarrow$H2 inversion seen in the main paper. \rev{These three statistics are computed from the same clustering, so they are consistency checks rather than independent evidence, and the independent measurement is the probe analysis of App.~\ref{sec:probe-appendix}, which uses no SAE, projection or clustering. Because the ANOVA runs across hypothesis classes over all 5,478 states with degrees of freedom fixed by the class counts, every $F$ in these tables is nominally significant at any conventional threshold, so the informative comparisons are the relative magnitudes together with the effect sizes ($V$, AMI) reported here.}

\begin{table}[h]
\centering\small
\begin{tabular}{lccc}
\toprule
Statistic & $F$ & df & Nominal $p$ \\
\midrule
H1 turn, base model & 5285 & (1, 5476) & $<10^{-300}$ \\
H1 turn, final checkpoint & 316 & (1, 5476) & $<10^{-68}$ \\
H2 progression, base model & 274 & (2, 5475) & $<10^{-100}$ \\
H2 progression, final checkpoint & 2330 & (2, 5475) & $<10^{-300}$ \\
\bottomrule
\end{tabular}
\caption{Degrees of freedom and nominal $p$ for the central layer-12 $F$ values of Table~\ref{tab:cluster-stats-combined}. Both are fixed by the design: the numerator is the number of hypothesis classes minus one and the denominator is 5,478 minus the number of classes. Every value is significant at any conventional threshold, which is why the paper compares relative magnitudes rather than resting on significance.}
\label{tab:dof-pvalues}
\end{table}

\begin{table}[h]
\centering\small
\input{revisions/tables/sae/cluster_stats_main.tex}
\caption{Cluster-vs-hypothesis statistics at layer 12 of Qwen2.5-1.5B (NL prompt), comparing the base model and the converged GRPO-from-base run. $F$ is the ANOVA on the principal t-SNE axis. $V$ is Cram\'er's V on the $K$-means cluster by hypothesis contingency table. AMI is adjusted mutual information. The H1 to H2 inversion is consistent across all three measures. H1 weakens (e.g.\ $V$ $0.70\!\to\!0.36$), H2 strengthens ($V$ $0.48\!\to\!0.62$), H3 and H4 stay weak, and H7 is stable.}
\label{tab:cluster-main-snapshot}
\end{table}

\clearpage

\section{Reward decomposition: full table}
\label{sec:reward-decomp}

\begin{table*}[h]
\centering\small
\input{revisions/tables/format_vs_grpo_full.tex}
\caption{Best-checkpoint accuracy (\%) for each model, training group, and input perturbation. ``NL+XY'' is a natural-language prompt with random symbol substitution applied. ``ASCII'' is an ASCII board input. ``--'' marks settings where the off-the-shelf evaluations were not run under the corresponding random symbol substitution. The structured-output \textit{Off-the-shelf} rows isolate the format component of the GRPO reward. The gap to the GRPO rows in the same column quantifies the contribution of the correctness signal. The persistent ASCII degradation across both legal-move and best-move objectives indicates that GRPO does not produce input-format-invariant board understanding.}
\label{tab:reward-full}
\end{table*}

\section{Per-position behavioural breakdown}
\label{sec:per-position-app}

\begin{table*}[h]
\centering\small
\input{revisions/tables/per_position_main.tex}
\caption{Per-position best-move accuracy (\%) for the strongest GRPO best-move checkpoint per model on the random split (NL prompt, no symbol substitution). Branching is the number of legal moves. Multiplicity is the number of equally-optimal moves. Phase counts ply (move number). \rev{Values are raw accuracies, not corrected for the per-state chance level, which itself varies with branching, and the chance-corrected analysis in the main text shows the branching gradient is largely a chance artifact.} Llama3.2-1B is an outlier: its late-game accuracy ($17\%$) is below its mid-game accuracy ($31\%$), indicating that even basic terminal-state recognition was not internalised. The \rev{raw gradient recurs} at lower absolute accuracy for Qwen2.5-0.5B (low-branch $63\%$ vs.\ high-branch $33\%$).}
\label{tab:per-position}
\end{table*}

\end{document}

%% file: revisions/tables/sae/cluster_stats_combined.tex
\begin{tabular}{l rrr @{\hskip 0.6em}|@{\hskip 0.6em} rrrrrrrrrrr}
\toprule
 & \multicolumn{3}{c}{Trajectory at L12} & \multicolumn{11}{c}{Per-layer at converged ckpt} \\
\cmidrule(lr){2-4}\cmidrule(lr){5-15}
Hypothesis & Base & Mid & Final & L4 & L8 & \textbf{L12} & L13 & L14 & L15 & L16 & L17 & L18 & L20 & L24\\
\midrule
H1 turn identity        & 5285 &  366 &  316 &  270 &  140 & \textbf{316}  &  114 &  269 &  188 &   95 & 2208 & 5379 & 1508 & 5484\\
H2 game progression     &  274 & 1686 & 2330 & 1151 &  594 & \textbf{2330} &  412 & 1278 &  287 &  358 &  747 &  284 &  114 &  400\\
H3 strategic situation  &   82 &  181 &  194 &  123 &   68 & \textbf{194}  &   63 &  136 &   53 &   69 &  187 &   86 &   32 &  126\\
H4 symmetry class       &   26 &    9 &   12 &    4 &    3 & \textbf{12}   &    3 &    5 &    2 &    2 &   13 &   42 &    7 &   23\\
H7 line-purity template &   11 &   38 &   39 &   28 &   24 & \textbf{39}   &   27 &   33 &   36 &   23 &   22 &   13 &   17 &   12\\
\bottomrule
\end{tabular}

%% file: revisions/tables/sae/cluster_stats_main.tex
\begin{tabular}{lcccccc}
\toprule
 & \multicolumn{3}{c}{Base} & \multicolumn{3}{c}{GRPO best ckpt} \\
\cmidrule(lr){2-4}\cmidrule(lr){5-7}
Hypothesis & $F$ & V & AMI & $F$ & V & AMI \\
\midrule
H1 turn identity & 5285 & 0.70 & 0.33 & 316 & 0.36 & 0.07\\
H2 game progression & 274 & 0.48 & 0.13 & 2330 & 0.62 & 0.22\\
H3 strategic situation & 82 & 0.24 & 0.05 & 194 & 0.22 & 0.05\\
H4 symmetry class & 26 & 0.56 & 0.13 & 12 & 0.47 & 0.07\\
H7 line-purity template & 11 & 0.40 & 0.25 & 39 & 0.40 & 0.22\\
\bottomrule
\end{tabular}

%% file: revisions/tables/format_vs_grpo_full.tex
\begin{tabular}{llccccc}
\toprule
Model & Setting & NL & NL+XY & ASCII & ASCII+XY\\
\midrule
Qwen2.5-1.5B & Off-the-shelf legal & 11.7 & --  & 11.1 & --\\
 & GRPO legal & 100.0 & 96.5  & 65.6 & 54.5\\
 & Off-the-shelf best & 8.0 & --  & 12.2 & --\\
 & GRPO best (from base) & 71.6 & 61.9  & 47.4 & 41.9\\
 & GRPO best (from legal) & 75.6 & 70.1  & 50.3 & 50.5\\
\midrule
Qwen2.5-0.5B & Off-the-shelf legal & 26.8 & --  & 26.6 & --\\
 & GRPO legal & 56.3 & 53.4  & 49.2 & 50.8\\
 & Off-the-shelf best & 0.4 & --  & 1.7 & --\\
 & GRPO best (from base) & 34.6 & 34.1  & 34.1 & 34.1\\
 & GRPO best (from legal) & 42.1 & 37.7  & 37.3 & 33.5\\
\midrule
Llama3.2-1B & Off-the-shelf legal & 16.2 & --  & 17.5 & --\\
 & GRPO legal & 36.4 & 32.4  & 33.9 & 29.9\\
 & Off-the-shelf best & 8.0 & --  & 7.7 & --\\
 & GRPO best (from base) & 10.4 & 10.4  & 9.3 & 10.4\\
 & GRPO best (from legal) & 27.3 & 26.2  & 22.6 & 22.2\\
\bottomrule
\end{tabular}

%% file: revisions/tables/per_position_main.tex
\begin{tabular}{lcccccc}
\toprule
 & \multicolumn{2}{c}{Branching ($\#$legal moves)} & \multicolumn{2}{c}{Best-move multiplicity} & \multicolumn{2}{c}{Game phase}\\
\cmidrule(lr){2-3}\cmidrule(lr){4-5}\cmidrule(lr){6-7}
Model & low ($\le2$) & high ($\ge5$) & unique & $\ge2$ & mid (5--6) & late ($\ge7$)\\
\midrule
Qwen2.5-1.5B & 89.4 & 63.6 & 62.1 & 86.9 & 73.3 & 88.4\\
Qwen2.5-0.5B & 62.9 & 33.3 & 21.8 & 44.3 & 37.3 & 57.8\\
Llama3.2-1B & 14.7 & 32.3 & 18.6 & 55.7 & 30.7 & 17.0\\
\bottomrule
\end{tabular}